\documentclass{article}

\usepackage{microtype}
\usepackage{graphicx}
\usepackage{booktabs} 
\usepackage[colorlinks,
            linkcolor=red,
            anchorcolor=blue,
            citecolor=purple,
            ]{hyperref}

\usepackage{pifont}
\usepackage{xcolor}
\usepackage{multirow}
\usepackage{pifont}
\usepackage{float}
\definecolor{sgreen}{RGB}{30, 150, 30} 
\definecolor{green}{RGB}{0,210,0}
\definecolor{myblue}{RGB}{218,232,252}
\definecolor{mygray}{RGB}{220,220,220}
\definecolor{mypink}{RGB}{251,49,153}
\usepackage{graphicx} 
\usepackage{pifont}
\usepackage{stfloats}
\usepackage{colortbl}
\usepackage{multicol} 
\usepackage{wrapfig}
\usepackage{subfig}
\usepackage{lmodern}

\usepackage[accepted]{icml2024}

\usepackage{amsmath}
\usepackage{amssymb}
\usepackage{mathtools}
\usepackage{amsthm}

\usepackage[capitalize,noabbrev]{cleveref}

\theoremstyle{plain}

\theoremstyle{definition}

\theoremstyle{remark}

\usepackage[textsize=tiny]{todonotes}

\begin{document}

\twocolumn[
\icmltitle{MoE-LLaVA: Mixture of Experts for Large Vision-Language Models}



\icmlsetsymbol{equal}{*}

\begin{icmlauthorlist}

\icmlauthor{Bin Lin}{pku}
\icmlauthor{Zhenyu Tang}{pku}
\icmlauthor{Yang Ye}{sysu}
\icmlauthor{Jinfa Huang}{rochester}
\icmlauthor{Junwu Zhang}{pku}
\icmlauthor{Yatian Pang}{nus,pku}
\icmlauthor{Peng Jin}{pku}
\icmlauthor{Munan Ning}{pku,pclab}
\icmlauthor{Jiebo Luo}{rochester}
\icmlauthor{Li Yuan}{pku,pclab}
\end{icmlauthorlist}

\icmlaffiliation{pku}{Peking University}
\icmlaffiliation{pclab}{Peng Cheng Laboratory}
\icmlaffiliation{sysu}{Sun Yat-sen University}
\icmlaffiliation{rochester}{University of Rochester}
\icmlaffiliation{nus}{National University of Singapore}
\icmlcorrespondingauthor{Li Yuan}{yuanli-ece@pku.edu.cn}

\icmlkeywords{Machine Learning, ICML}

\vskip 0.3in
]



\printAffiliationsAndNotice{}  

\begin{abstract}


Recent advances demonstrate that scaling Large Vision-Language Models (LVLMs) effectively improves downstream task performances. However, existing scaling methods enable all model parameters to be active for each token in the calculation, which brings massive training and inferring costs.
In this work, we propose a simple yet effective training strategy \textbf{MoE-Tuning} for LVLMs. This strategy innovatively addresses the common issue of performance degradation in multi-modal sparsity learning, consequently constructing a sparse model with an outrageous number of parameters but a constant computational cost.
Furthermore, we present the \textbf{MoE-LLaVA}, a MoE-based sparse LVLM architecture, which uniquely activates only the top-$k$ experts through routers during deployment, keeping the remaining experts inactive.
Extensive experiments show the significant performance of MoE-LLaVA in a variety of visual understanding and object hallucination benchmarks.
Remarkably, with only approximately \textbf{3B} sparsely activated parameters, MoE-LLaVA demonstrates performance comparable to the LLaVA-1.5-7B on various visual understanding datasets and even surpasses the LLaVA-1.5-13B in object hallucination benchmark.
Through MoE-LLaVA, we aim to establish a baseline for sparse LVLMs and provide valuable insights for future research in developing more efficient and effective multi-modal learning systems.
Code is released at \href{https://github.com/PKU-YuanGroup/MoE-LLaVA}{\textcolor{mypink}{https://github.com/PKU-YuanGroup/MoE-LLaVA}}.

\end{abstract}    
\section{Introduction}

\label{sec:intro}

\begin{figure}[h]
\centering
    \includegraphics[width=1.0\linewidth]{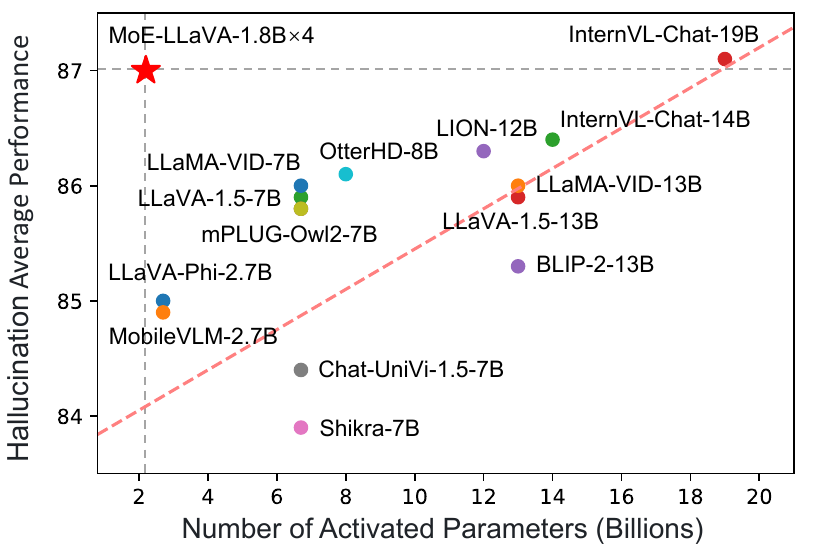}       
\vskip -0.1in 
\caption{\textbf{Comparison between MoE-LLaVA-1.8B×4 and open-source LVLMs on object hallucination benchmark.} We report the average performance on the POPE~\cite{li2023evaluating} benchmark, which includes three subsets of Adversarial, Random, and Popular. The red dashed line represents the linear fit to the data points of all models except MoE-LLaVA.}
\label{fig:intro0}
\vskip -0.2in
\end{figure}

\begin{figure}[h]
\centering
    \includegraphics[width=1.0\linewidth]{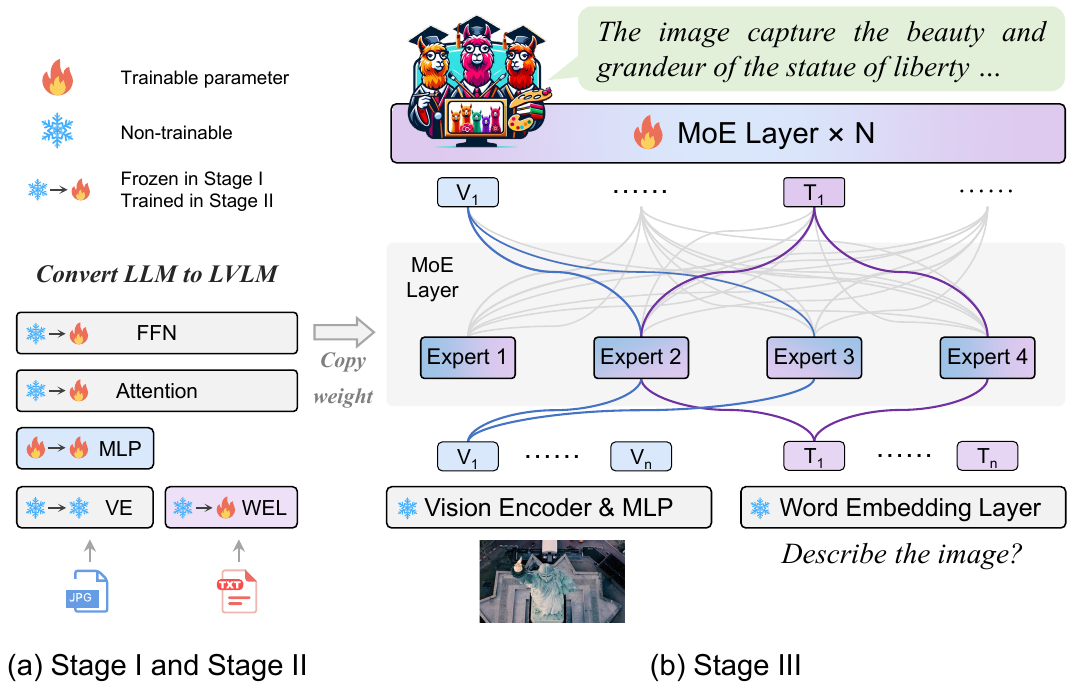}       
\vskip -0.1in 
\caption{\textbf{Illustration of MoE-Tuning.} The MoE-Tuning consists of three stages. In stage \uppercase\expandafter{\romannumeral1}, only the MLP is trained. In stage \uppercase\expandafter{\romannumeral2}, all parameters are trained except for the Vision Encoder (VE). In stage \uppercase\expandafter{\romannumeral3}, FFNs are used to initialize the experts in MoE, and only the MoE layers are trained. For each MoE layer, only two experts are activated for each token, while the other experts remain silent.}
\label{fig:intro}
\vskip -0.2in
\end{figure}

Large Vision-Language Models (LVLMs), such as LLaVA~\cite{liu2023visual} and MiniGPT-4~\cite{zhu2023minigpt}, have shown promising results by leveraging an image encoder and several visual projection layers to enhance the visual perception capabilities of the Large Language Models (LLMs). Typically, increasing the model size~\cite{zhang2023internlm, bai2023qwenvl} and dataset scale~\cite{zhang2023llavar, zhao2023svit, chen2023sharegpt4v} can improve model performance. For instance, InternVL~\cite{chen2023internvl} has extended the image encoder to 6B parameters.
A series of works~\cite{li2022blip, dai2023instructblip, liu2023improved} have expanded the backend of LVLM to 13B parameters and achieved state-of-the-art performance on downstream tasks. IDEFICS~\cite{laurencon2023obelics} even trained an LVLM with 80B parameters. These methods have demonstrated superior performance even in LLMs, which are typically pretrained on 34B parameters~\cite{SUS-Chat, yi, Aquila2} or 70B parameters~\cite{touvron2023llama, touvron2023llama2, bai2023qwen, deepseek-llm, zhang2023xuanyuan}, and models surpassing 100B parameters are common~\cite{brown2020language, zeng2022glm, zhang2022opt, scao2022bloom, li2023flm, falcon} .

In practical applications, scaling model with high-quality training data is crucial for improving model performance~\cite{lepikhin2020gshard}. However, training and deploying such large models demand significant computational costs and efficient implementation on parallel devices, which can be extremely expensive. This is because each token requires computations with all model parameters, called the dense model. In contrast, sparse Mixtures of Experts (MoE)~\cite{jacobs1991adaptive, eigen2013learning} effectively scale model capacity by using fixed activated parameters to process data, which has thrived in the field of NLP~\cite{fedus2022switch, zoph2022st, komatsuzaki2022sparse}. Recently, Mistral LLM~\cite{jiang2023mistral} equipped with the MoE layers has gained popularity in LLMs. Mixtral-MoE-8×7B~\cite{jiang2024mixtral} achieves performance comparable to LLaMA 2-70B with fewer computational resources.


However, directly applying MoE to train sparse LVLMs is challenging. We observe that simultaneously converting LLM to LVLM and sparsifying the model leads to significant performance degradation. After multiple attempts, we find that proper initialization is crucial for sparsifying the LVLM, Therefore, we introduce a simple yet effective three-stage training strategy \textbf{MoE-Tuning}. Specifically, as shown in~\cref{fig:intro}, we first train an MLP that adapts visual tokens to the LLM in stage \uppercase\expandafter{\romannumeral1}. Then, we pre-empower the LVLM with a general multi-modal understanding capability by training the whole LLM's parameters in stage \uppercase\expandafter{\romannumeral2}. Furthermore, in stage \uppercase\expandafter{\romannumeral3} we replicate the FFN as the initialization weights for the experts and only train the MoE layers. Finally, the sparse model gradually transitions from a general LVLM initialization to sparse mixture of experts.

In this work, we explore a baseline for the LVLM with mixture of experts called \textbf{MoE-LLaVA}, which incorporates mixture of experts and learnable routers. MoE-LLaVA consists of multiple sparse paths where each token is dispatched to different experts through the router. The activated experts collectively process the tokens, while the inactive paths remain silent. By iteratively stacking MoE encoder layers, MoE-LLaVA provides a sparse path toward a larger and more powerful LVLM.

As a result, in~\cref{fig:intro0}, our MoE-LLaVA with only 2.2B sparse activated parameters outperforms models with similar activated parameters and LLaVA-1.5-13B, surpassing it by a large margin on the POPE object hallucination benchmark. Additionally, MoE-LLaVA achieves comparable performance to InternVL-Chat-19B, which has approximately 8 times the activated parameters. We further scale MoE-LLaVA to 3.6B sparse activated parameters, which outperform LLaVA-1.5-7B by 1.9\%, 0.4\%, 0.9\%, 30.7\%, and 3.8\% in ScienceQA, POPE, MMBench, LLaVA$^\text{W}$, and MM-Vet, respectively. Extensive experiments validate the rationality of our MoE-LLaVA architecture and MoE-Tuning strategy.

We summarize our primary contributions as follows:
\begin{itemize}
\item We explore the \textbf{MoE-Tuning}, a novel three-stage training strategy for adapting MoE to LVLMs and preventing the model degradation caused by sparsity.
\item We propose \textbf{MoE-LLaVA}, a MoE-based sparse LVLM framework, which significantly expands the number of parameters while maintaining computational costs.
\item Extensive experiments demonstrate that our \textbf{MoE-LLaVA} has excellent multi-modal understanding and hallucination mitigation abilities. 
With only approximately \textbf{3B} sparse activated parameters, our method achieves comparable performance with SOTA 7B models on the visual understanding benchmarks. It is worth noting that MoE-LLaVA outperforms LLaVA-1.5-13B by 1.1\% on the POPE hallucination benchmark with \textbf{2.2B} activated parameters.


\end{itemize}

\section{Related Work}
\label{sec:related}

\subsection{Large Vision-Language Models}
Powerful LLMs~\cite{openai2023gpt4, touvron2023llama, wei2022chain, touvron2023llama2, zheng2023judging, team2023internlm, sun2023moss, du2021glm, bai2023qwen, yang2023baichuan, penedo2023refinedweb, taori2023alpaca} with strong instruction-following and generalization capabilities have been applied to LVLMs. Early works such as BLIP-2~\cite{li2023blip} and FROMAGe~\cite{koh2023grounding} encoded visual signals into a sequence of visual tokens, successfully adapting vision to LLMs through several projection layers. Subsequently, recent works have focused on improving performance through methods such as expanding the instruction-tuning dataset~\cite{liu2023aligning, liu2023visual, zhang2023llavar, zhao2023svit, chen2023sharegpt4v}, optimizing training strategies~\cite{bai2023qwenvl, chen2023minigpt}, increasing resolution of image~\cite{liu2023improved, bai2023qwenvl, wang2023cogvlm} enhancing image encoders~\cite{chen2023internvl, zhang2023internlm, bai2023qwenvl}, aligning the input~\cite{lin2023video} and projection layers~\cite{cha2023honeybee, alayrac2022flamingo, bai2023qwenvl, dai2023instructblip, ye2023mplug, zhao2023svit}. These works empowered LVLMs with powerful visual understanding capabilities by expanding the visual instruction fine-tuning datasets and model scales. 

Currently, some works have endowed LVLMs with fine-grained image understanding capabilities, such as region understanding~\cite{chen2023shikra, zhao2023bubogpt, liu2023interngpt}, multi-region understanding~\cite{wang2023visionllm, pi2023detgpt, peng2023kosmos}, and pixel-wise grounding~\cite{rasheed2023glamm, lai2023lisa}. However, the cost of scaling up dense visual data and models is challenging to bear~\cite{liu2022swin, yin2023survey}. In this work, we aim to make state-of-the-art LVLMs research more accessible by leveraging mixture of experts.

\subsection{Mixture of Experts in Multi-modal Learning}
Mixture of Experts (MoE)~\cite{jacobs1991adaptive, eigen2013learning} is a hybrid model consisting of multiple sub-models, known as experts, which are integrated together. The key concept of MoE is the use of a router to determine the token set that each expert handles, thereby reducing interference between different types of samples.

\begin{figure*}[t]
\centering
    \includegraphics[width=0.97\linewidth]{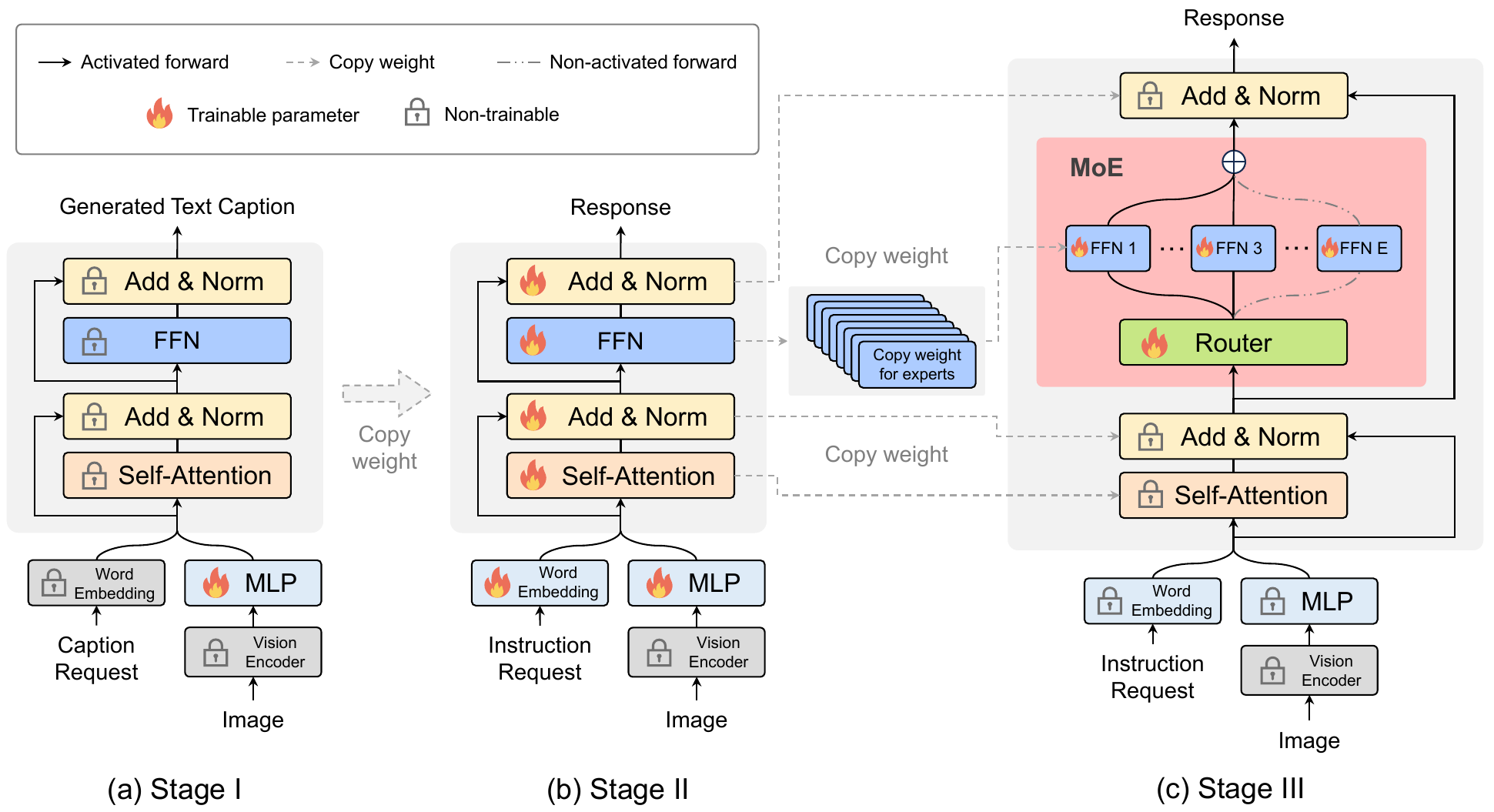}     
\vskip -0.1in
\caption{\textbf{Training framework and strategy.} MoE-LLaVA adopts a three-stage training strategy. (a) We solely train the MLP to adapt the LLM to visual inputs. (b) Training the LLM backend empowers multi-modal understanding capability and MoE layers are not involved. (c) In this stage, we replicate the weights of the FFN to initialize each expert.}
\label{fig:framework}
\vskip -0.1in
\end{figure*}

\textbf{Hard Routers.} In the hard router mode, each expert is typically pre-defined as a specific pattern. This is because multi-modal data naturally exhibit gaps~\cite{liang2022mind}, making it difficult for soft routers to learn the optimal patterns for assigning tokens to different experts. A series of works~\cite{bao2022vlmo, long2023multiway, satar2022rome, wang2022image, shen2023scaling} naturally decouple experts based on modal categories and pre-define each expert to handle a specific modality. An important feature of these hard-based routers is that they do not require learning the router. This mode is also widely applied in the task-specific MoE~\cite{li2023pace, zhu2022uni, ma2023cot, kudugunta2021beyond}.

\textbf{Soft Routers.} Some works~\cite{shazeer2017outrageously, lepikhin2020gshard, fedus2022switch, zoph2022st, komatsuzaki2022sparse} in natural language process have explored the MoE based on soft routers. Soft routers enable dynamic allocation of data among different experts, allowing each expert to focus on its expertise and achieve model sparsity. Therefore, our main focus is on leveraging soft routers in the MoE. Small-scale (million-level) models based on soft routers have also been explored in the context of multi-modal learning, such as EVE~\cite{chen2023eve} and LIMoE~\cite{mustafa2022multimodal}, which attempt a fusion of data by using soft routers. The work most relevant to ours is MoCLE~\cite{gou2023mixture}. However, MoCLE clusters different instruction sets and distributes them to different experts, which compromises the flexibility and autonomy of the experts. Differently, MoE-LLaVA relies on knowledge-rich routers to distribute tokens to different paths.


\section{Method}
\label{sec:moellava}

\subsection{Overview}
As shown in~\cref{fig:framework}, MoE-LLaVA consists of a vision encoder, a visual projection layer (MLP), a word embedding layer, multiple stacked LLM blocks, and MoE blocks. We first introduce the model architecture of MoE-LLaVA in three stages in~\cref{sec:model}. Furthermore, in~\cref{sec:moetuning}, we explain how to train MoE-LLaVA. Finally, in~\cref{sec:objective}, we elaborate on the training objectives of MoE-LLaVA.

\begin{table*}[t]
  \small
  \setlength\tabcolsep{1.0mm}
  \caption{\textbf{Architecture details of the MoE-LLaVA model.} ``FFN Factor'' represents the number of linear layers in the FFN. ``1.6B×4-Top2'' represents a dense foundation model with 1.6B parameters, which is equipped with a total of four experts, two of them being activated.}
  \vskip 0.1in
  \label{tab:model}
  \centering
  \begin{tabular}{l|ccc|cccccc|cc}
    \toprule
    \multirow{2}{*}{\textbf{Name}} & \multirow{2}{*}{\textbf{Experts}} & \multirow{2}{*}{\textbf{Top-k}} & \textbf{MoE} & \multirow{2}{*}{\textbf{Embedding}} & \multirow{2}{*}{\textbf{Width}} & \multirow{2}{*}{\textbf{Layers}} & \multirow{2}{*}{\textbf{FFN}} & \textbf{FFN} & \multirow{2}{*}{\textbf{Heads}} & \textbf{Activated} & \textbf{Total} \\
     &  &  & \textbf{Layers} & & &  & & \textbf{Factor} &  & \textbf{Param} & \textbf{Param} \\
    \midrule
    StableLM-1.6B~\cite{StableLM-2-1.6B} & - & - & - & 100352 & 2560 & 32 & 10240 & 2 & 32 & 1.6B & 1.6B \\
    \rowcolor{myblue} MoE-LLaVA-1.6B×4-Top2 & 4 & 2 & 16 & 100352 & 2560 & 32 & 10240 & 2 & 32 & 2.0B & 2.9B \\
    \midrule
    Qwen-1.8B~\cite{bai2023qwen} & - & - & - & 151936 & 2048 & 24 & 5504 & 3 & 16 & 1.8B & 1.8B \\
    \rowcolor{myblue} MoE-LLaVA-1.8B×4-Top2 & 4 & 2 & 12 & 151936 & 2048 & 24 & 5504 & 3 & 16 & 2.2B & 3.1B \\
    \midrule
    Phi2-2.7B~\cite{phi2} & - & - & - & 51200 & 2560 & 32 & 10240 & 2 & 32 & 2.7B & 2.7B \\
    \rowcolor{myblue} MoE-LLaVA-2.7B×4-Top2 & 4 & 2 & 16 & 51200 & 2560 & 32 & 10240 & 2 & 32 & 3.6B & 5.3B \\
    \bottomrule
  \end{tabular}
\vskip -0.1in
\end{table*}

\subsection{Architecture of MoE-LLaVA}
\label{sec:model}

As shown in~\cref{tab:model}, we present the detailed configuration of MoE-LLaVA and more details can be found in~\cref{appendix:more_model}. Given a RGB image $\mathbf{v} \in \mathbb{R}^{H \times W \times 3}$, where $H$ and $W$ are the origin resolution. The vision encoder processes input images to obtain a visual token sequence $\mathcal{Z}=[z_{1},z_{2},\cdots,z_{P}]\in\mathbb{R}^{P\times C}$, where $P=\frac{H \times W}{14^2}$ represents the sequence length of visual tokens. A visual projection layer $f$ is used to map $\mathcal{Z}\in\mathbb{R}^{P\times C}$ to $\mathcal{V}\in\mathbb{R}^{P\times D}$, where $D$ represents the hidden size of LLM. Similarly, the text undergoes a word embedding layer $g$ and is projected to obtain the sequence tokens $\mathcal{T}=[t_{1},t_{2},\cdots,t_{N}]\in\mathbb{R}^{N\times D}$, where $N$ represents the sequence length of text tokens.

Subsequently, we concatenate the visual tokens and text tokens together and feed them into a large language model. Instead, we solely train the visual projection layer. The large language model consists of stacked multi-head self-attention (MSA) and feed-forward neural networks (FFN). Layer normalization (LN) and residual connections are applied within each block~\cite{wang2019learning, baevski2018adaptive}. Therefore, we formulate as:

\vskip -0.15in
\begin{equation}
    \mathbf{x}_0 =[v_{1},v_{2},\cdots,v_{P},\cdots,t_{1},t_{2},\cdots,t_{N}],
\end{equation}
\vspace{-0.5cm}
\begin{equation}
    \mathbf{x}_{\ell}^{\prime} =\mathrm{MSA}(\mathrm{LN}(\mathbf{x}_{\ell-1}))+\mathbf{x}_{\ell-1},  \ell=1\ldots L,
\label{ffn}
\end{equation}
\vspace{-0.5cm}
\begin{equation}
    \mathbf{x}_{\ell} =\mathrm{MoE}(\mathrm{LN}(\mathbf{x^{\prime}}_{\ell}))+\mathbf{x^{\prime}}_{\ell}, \ell=1\ldots L,
\label{ffn2}
\end{equation}
\vspace{-0.5cm}
\begin{equation}
    \mathcal{Y}=\mathrm{LN}(\mathbf{x}_L).
\end{equation}

\textbf{MoE Forward.} Typically, a MoE layer consists of multiple FFNs. As an initialization step, we replicate the FFNs from stage \uppercase\expandafter{\romannumeral2} to form an ensemble of experts $\mathcal{E}=[e_{1},e_{2},\cdots,e_{E}]$. The router is a linear layer that predicts the probability of each token being assigned to each expert. We formulate as:

\vskip -0.15in
\begin{equation}
    \mathcal{P}(\mathbf{x})_i=\frac{e^{f(\mathbf{x})_i}}{\sum_j^E e^{f(\mathbf{x})_j}},
\end{equation}

where the router produces weight logits $f(\mathbf{x}) = \mathbf{W}\cdot\mathbf{x}$, which are normalized by the softmax function. The $\mathbf{W}\in\mathbb{R}^{D\times E}$ represents the lightweight training parameters and $E$ represents the number of experts.  Therefore, each token is processed by the top-$k$ experts with the highest probabilities, and the weighted sum is calculated based on
the softmax results of the probabilities:

\vskip -0.15in
\begin{equation}
    \mathrm{MoE}(\mathbf{x})=\sum_{i=1}^k\mathcal{P}(\mathbf{x})_i\cdot\mathcal{E}(\mathbf{x})_i.
\end{equation}


\subsection{MoE-Tuning}
\label{sec:moetuning}


\textbf{Stage \uppercase\expandafter{\romannumeral1}:} In this stage, our objective is to adapt the image tokens to LLM, allowing the LLM to comprehend the instances in the images. To achieve this, we employ an MLP to project the image tokens into the input domain of the LLM, treating the image patches as pseudo-text tokens. During this stage, the LLM is trained to describe the images. MoE layers are not applied to the LLM during this stage.

\textbf{Stage \uppercase\expandafter{\romannumeral2}:} Tuning with multi-modal instruction data is a key technique to enhance the capabilities and controllability of large models~\cite{zhang2023instruction}. In this stage, LLM is adjusted to become an LVLM with multi-modal understanding. We use more complex instructions, including tasks such as image logical reasoning and text recognition, which require the model to have a stronger multi-modal understanding. Typically, for dense models, the LVLM training is considered complete at this stage. However, we encounter challenges in simultaneously transforming the LLM into an LVLM and sparsifying the LVLM. Therefore, MoE-LLaVA utilizes the weights from the second stage as initialization for the third stage to alleviate the learning difficulty of the sparse model.

\textbf{Stage \uppercase\expandafter{\romannumeral3}:} As an initialization, we replicate the FFN multiple times to initialize the experts. When image tokens and text tokens are fed into the MoE layers, the router calculates the matching weights between each token and the experts. Each token is then processed by the top-$k$ experts, and the outputs are aggregated by weighted summation based on the router's weights. When the top-$k$ experts are activated, the remaining experts remain silent. This modeling approach forms the MoE-LLaVA with infinitely possible sparse pathways, offering a wide range of capabilities.

\subsection{Training Objectives}
\label{sec:objective}

The $\mathcal{L}_{\text{total}}$ consists of auto-regressive loss $\mathcal{L}_{\text{regressive}}$ and auxiliary loss $\mathcal{L}_{\text{aux}}$, and auxiliary loss are scaled by the balancing coefficient $\alpha$:

\vskip -0.15in
\begin{equation}
  \mathcal{L}_{\text{total}}=\mathcal{L}_{\text{regressive}} + \alpha \cdot \mathcal{L}_{\text{aux}}.
  \label{eq:gen}
\end{equation}

\textbf{Auto-Regressive Loss.} We optimize the output of LLM through a generative loss in an auto-regressive manner. Given an image and text, MoE-LLaVA generates the output sequence $\mathcal{Y}=[y_{1},y_{2},\cdots,y_{K}]\in\mathbb{R}^{K\times D}$ by progressively generating each element, where $K = P + N$ represents the length of the output sequence. The formula is:

\vskip -0.15in
\begin{equation}
  \mathcal{L}_{\text{regressive}}=-\sum_{i=1}^N \text{log} \ p_\theta\left(\mathcal{Y}^{[P+i]} \mid \mathcal{V}, \mathcal{T}^{[:i-1]}\right),
  \label{eq:gen}
\end{equation}
where $\theta$ is a trainable parameter and we only calculate the loss for the generated text.

\textbf{Auxiliary Loss.} Due to the presence of multiple experts, it is necessary to impose load balancing constraints on the MoE layer. We incorporate differentiable load balancing loss~\cite{fedus2022switch} into each MoE layer to encourage experts to handle tokens in a balanced manner as follows:

\vskip -0.15in
\begin{equation}
    \mathcal{L}_{\text{aux}}=E\cdot\sum_{i=1}^{E}\mathcal{F}_i\cdot \mathcal{G}_i,
\end{equation} 

where $\mathcal{F}$ represents the fraction of tokens processed by each expert $\mathcal{E}_i$, and $\mathcal{G}$ represents the average routing probability of $\mathcal{E}_i$, which can be expressed by the following formulas:
    
\vskip -0.15in
\begin{equation}
    \mathcal{F}=\frac{1}{K}\sum_{i=1}^E\mathrm{1}\{\operatorname{argmax}\mathcal{P}(\mathbf{x})=i\},
\end{equation} 
\vskip -0.15in
\begin{equation}
    \mathcal{G}=\frac{1}{K}\sum_{i=1}^{K}\mathcal{P}(\mathbf{x})_i.
\end{equation}

\section{Experiments}
\label{sec:exp}

\subsection{Experimental Setup}

\textbf{Model Settings.} Following LLaVA 1.5~\cite{liu2023improved}, we utilize CLIP-Large~\cite{radford2021learning} as the vision encoder, and the MLP consists of two linear layers with GELU activation function~\cite{hendrycks2016gaussian} between them. Unless otherwise specified, MoE-LLaVA employs an alternating replacement of FFN with MoE layers, meaning that the number of MoE layers is half of the total number of layers. The value of balancing coefficient $\alpha$ is 0.01. We provide additional training details in~\cref{appendix:train_detail}.

\textbf{Data Details.} As shown in~\cref{tab:data}, we reorganize the currently available data for the three-stage training. For the first stage of pretraining, we use the pretrained data of LLaVA 1.5-558k~\cite{liu2023improved}. For the second stage, we collect datasets from MIMIC-IT~\cite{li2023mimic}, LRV~\cite{liu2023aligning}, SViT~\cite{zhao2023svit} and LVIS~\cite{ wang2023see} to provide a robust initialization for MoE-LLaVA. For the third stage, we utilize the same data pipeline as LLaVA-mix-665k~\cite{liu2023improved}.

\begin{table}[t]
  \small
  \setlength\tabcolsep{1.3mm}
  \caption{\textbf{Composition of the data groups.} For MIMIC-IT, and SViT datasets, we only use the LA split, and core split, respectively.}
  \vskip 0.1in
  \label{tab:data}
  \centering
  \begin{tabular}{l|llc}
    \toprule
    \textbf{Data group} & \textbf{Usage} & \textbf{Source} & \textbf{\#Sample} \\
    \midrule
    LLaVA-PT & Stage \uppercase\expandafter{\romannumeral1}  & LLaVA 1.5-558k & 558k \\
    \midrule
    \multirow{2}{*}{Hybird-FT} & \multirow{2}{*}{Stage \uppercase\expandafter{\romannumeral2}} & SViT-157k, LVIS-220k & \multirow{2}{*}{964k} \\
     & &  LRV-331k, MIMIC-IT-256k &  \\
    \midrule
    LLaVA-FT & Stage \uppercase\expandafter{\romannumeral3} & LLaVA 1.5-mix-665k & 665k \\
    \bottomrule
  \end{tabular}
\vskip -0.1in
\end{table}

\begin{table*}[!t]
\small
  \setlength\tabcolsep{0.85mm}
\caption{\textbf{Comparison among different LVLMs on image understanding benchmarks.} ``Res.'', ``Act.'', ``L'', ``V'', ``S'', ``Q'', ``P'', ``M'' and ``I'' respectively represent the input image resolution, activated parameters, LLaMA~\cite{touvron2023llama}, Vicuna~\cite{chiang2023vicuna}, StableLM~\cite{StableLM-2-1.6B}, Qwen~\cite{bai2023qwen}, Phi-2~\cite{phi2} MobileLLaMA~\cite{chu2023mobilevlm} and IDEFICS~\cite{laurencon2023obelics}.
Evaluation Benchmarks include
VQA-v2~\cite{goyal2017making}; GQA~\cite{hudson2019gqa}; VisWiz~\cite{gurari2018vizwiz}; SQA$^\text{I}$: ScienceQA-IMG~\cite{lu2022learn}; VQA$^\text{T}$: TextVQA~\cite{singh2019towards}; POPE~\cite{li2023evaluating}; MME~\cite{fu2023mme}; MMB: MMBench~\cite{liu2023mmbench}; LLaVA$^\text{W}$: LLaVA-Bench (in-the-Wild)~\cite{liu2023visual}; MM-Vet~\cite{yu2023mm}. $^*$ donates that there is some overlap in the training data. $^\dag$ donates that the model is trained with an image resolution of 384. The best results and second best results are indicated by \textbf{boldface} and \underline{underline}, respectively.}
\vskip 0.1in
  \label{tab:main_res}
  \centering
  \begin{tabular}{l|ccc|ccccc|ccccc}
    \toprule
     \multirow{2}{*}{\textbf{Methods}} & \multirow{2}{*}{\textbf{LLM}} & \multirow{2}{*}{\textbf{Act.}} & \multirow{2}{*}{\textbf{Res.}}  & \multicolumn{5}{c|}{\textbf{Image Question Answering}} & \multicolumn{4}{c}{\textbf{Benchmark Toolkit}} \\
      &  & \textbf{} &  & VQA$^\text{v2}$ & GQA & VisWiz & SQA$^\text{I}$ & VQA$^\text{T}$ & POPE & MME & MMB & LLaVA$^\text{W}$ & MM-Vet \\
    \midrule
    \multicolumn{13}{l}{\textit{Dense Model}} \\
    \rowcolor{mygray}  \textcolor{gray}{I-80B~\cite{laurencon2023obelics}} &  \textcolor{gray}{L-65B}  & \textcolor{gray}{65B}&  \textcolor{gray}{224} &\textcolor{gray}{60.0} &\textcolor{gray}{45.2} & \textcolor{gray}{36.0} &  \textcolor{gray}{-} &  \textcolor{gray}{30.9} &  \textcolor{gray}{-} & \textcolor{gray}{-} &  \textcolor{gray}{54.5} &  \textcolor{gray}{-} &  \textcolor{gray}{-} \\
    \rowcolor{mygray} \textcolor{gray}{LLaVA-1.5~\cite{liu2023improved}} &  \textcolor{gray}{V-13B} &  \textcolor{gray}{13B}  & \textcolor{gray}{336} &  \textcolor{gray}{80.0$^*$} & \textcolor{gray}{63.3$^*$} &  \textcolor{gray}{53.6} &  \textcolor{gray}{71.6} & \textcolor{gray}{61.3} &  \textcolor{gray}{85.9} &  \textcolor{gray}{1531.3}  & \textcolor{gray}{67.7}  & \textcolor{gray}{70.7}  &  \textcolor{gray}{35.4} \\
    \rowcolor{mygray}  \textcolor{gray}{Qwen-VL~\cite{bai2023qwenvl}} & \textcolor{gray}{Q-7B} & \textcolor{gray}{6.7B} & \textcolor{gray}{448} & \textcolor{gray}{78.8$^*$} & \textcolor{gray}{59.3$^*$} & \textcolor{gray}{35.2} & \textcolor{gray}{67.1} & \textcolor{gray}{63.8} & \textcolor{gray}{-} & \textcolor{gray}{-} & \textcolor{gray}{38.2} & \textcolor{gray}{-} & \textcolor{gray}{-} \\
    \rowcolor{mygray} \textcolor{gray}{LLaVA-1.5~\cite{liu2023improved}} &  \textcolor{gray}{V-7B} &  \textcolor{gray}{6.7B}  &\textcolor{gray}{336} &  \textcolor{gray}{78.5$^*$} & \textcolor{gray}{62.0$^*$} &  \textcolor{gray}{50.0} &  \textcolor{gray}{66.8} & \textcolor{gray}{58.2} &  \textcolor{gray}{85.9} &  \textcolor{gray}{1510.7} &  \textcolor{gray}{64.3} &  \textcolor{gray}{63.4} &  \textcolor{gray}{30.5} \\
    TinyGPT-V~\cite{yuan2023tinygpt} & P-2.7B & 2.7B & 448 & - & 33.6$^*$ & 33.4 & - & - & - & - & - & - & - \\
    MobileVLM~\cite{chu2023mobilevlm} & M-2.7B & 2.7B & 336 & - & 59.0$^*$ & - & 61.0 & 47.5 & 84.9 & 1288.9 & 59.6 & - & - \\
    LLaVA-Phi~\cite{zhu2024llavaphi} & P-2.7B & 2.7B & 336 & 71.4$^*$ & - & 35.9 & {68.4} & {48.6} & {85.0} & 1335.1 & {59.8} & - & {28.9} \\
    \midrule
     \multicolumn{13}{l}{\textit{Sparse Model}} \\
    \rowcolor{myblue} MoE-LLaVA-1.6B×4-Top2 & S-1.6B & 2.0B & 336 & {76.7$^*$} & {60.3$^*$} & {36.2} & {62.6} & {50.1} & {85.7} & 1318.2 & {60.2} & {86.8} & {26.9} \\
    \rowcolor{myblue} MoE-LLaVA-1.8B×4-Top2 & Q-1.8B & 2.2B & 336 & {76.2$^*$} & \underline{61.5$^*$} & {32.6} & {63.1} & {48.0} & \textbf{87.0} & 1291.6 & 59.7 & {88.7} & {25.3} \\
    \rowcolor{myblue} MoE-LLaVA-2.7B×4-Top2 & P-2.7B & 3.6B & 336 & {77.6$^*$} & {61.4$^*$} & \textbf{43.9} & \underline{68.5} & {51.4} & \underline{86.3} & \underline{1423.0} & \underline{65.2} & \underline{94.1} & \underline{34.3} \\
    \midrule
    \rowcolor{myblue} MoE-LLaVA-1.6B×4-Top2$^\dag$ & S-1.6B & 2.0B & 384 & \underline{78.6$^*$} & \underline{61.5$^*$} & {40.5} & {63.9} & \underline{54.3} & {85.9} & 1335.7 & {63.3} & {90.3} & {32.3} \\
    \rowcolor{myblue} MoE-LLaVA-2.7B×4-Top2$^\dag$ & P-2.7B & 3.6B & 384 & \textbf{79.9$^*$} & \textbf{62.6$^*$} & \underline{43.7} & \textbf{70.3} & \textbf{57.0} & {85.7} & \textbf{1431.3} & \textbf{68.0} & \textbf{97.3} & \textbf{35.9} \\
    \bottomrule
  \end{tabular}
\vskip -0.1in
\end{table*}


\begin{table*}[!t]
\small
  \setlength\tabcolsep{1.5mm}
  \caption{\textbf{Zero-shot object hallucination evaluation results.} ``Yes'' indicates the proportion of positive responses to the given question.}
    \vskip 0.1in
  \label{tab:pope}
  \centering
  \begin{tabular}{l|cc|ccc|ccc|ccc}
    \toprule
    \multirow{2}{*}{\textbf{Methods}} & \multirow{2}{*}{\textbf{LLM}}& \multirow{2}{*}{\textbf{Activated}} & \multicolumn{3}{c|}{\textbf{Adersarial}} & \multicolumn{3}{c|}{\textbf{Popular}}  & \multicolumn{3}{c}{\textbf{Random}}  \\
     &  & & Acc & F1-Score & Yes & Acc & F1-Score & Yes & Acc & F1-Score & Yes \\
    \midrule
    \multicolumn{12}{l}{\textit{Dense Model}} \\
    mPLUG-Owl~\cite{ye2023mplug} & L-7B & 6.7B & 82.4 & 81.6 & 45.2 & 85.5 & 84.3 & 42.1 & 86.3 & 85.3 & 42.3 \\
    MM-GPT~\cite{gong2023multimodal} & L-7B & 6.7B & 50.0 & 66.7 & 100.0 & 50.0 & 66.7 & 100.0 & 50.0 & 66.7 & 100.0 \\
    \text{LLaVA-1.5}~\cite{liu2023improved} & V-13B & 13B & 85.5 & {84.4} & 43.3 & {87.4} & {86.2} & 41.3 & {88.0} & {87.1} & 41.7 \\
    \midrule
    \multicolumn{12}{l}{\textit{Sparse Model}} \\
    \rowcolor{myblue} MoE-LLaVA-1.6B×4-Top2 & S-1.6B & 2.0B & \textbf{86.9} & \textbf{85.7} & 41.7 & {85.3} & {84.2} & 43.5 & {88.0} & {87.1} & 41.6 \\
    \rowcolor{myblue} MoE-LLaVA-1.8B×4-Top2 & Q-1.8B & 2.2B & \underline{86.1}& {85.4} & 44.9  & \textbf{88.6}& \textbf{87.7} & 42.5 & \textbf{88.7} & \textbf{88.0} & 43.0 \\
    \rowcolor{myblue} MoE-LLaVA-2.7B×4-Top2 & P-2.7B & 3.6B & {85.9} & {84.9} & 43.2 & \underline{87.5} & \underline{86.4} & 41.8 & \underline{88.5} & \underline{87.7} &  41.8\\
    \midrule
    \rowcolor{myblue} MoE-LLaVA-1.6B×4-Top2$^\dag$ & S-1.6B & 2.0B & \textbf{86.9} & \underline{85.6} & 41.5 & {85.7} & {84.6} & 43.0 & {88.4} & {87.5} & 41.5 \\
    \rowcolor{myblue} MoE-LLaVA-2.7B×4-Top2$^\dag$ & P-2.7B & 3.6B & {85.5} & {84.2} & 41.9 & {86.7} & {84.4} & 41.7 & {87.9} & {86.9} & 40.6 \\
    \bottomrule
  \end{tabular}
\vskip -0.1in
\end{table*}



\begin{figure*}[t]
\vskip 0.2in
\centering
    \includegraphics[width=1.0\linewidth]{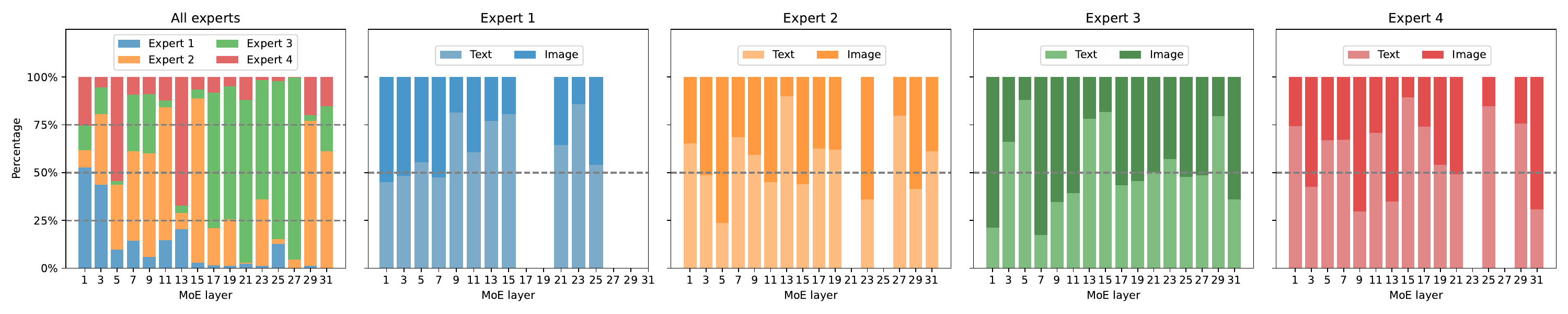}      
\vskip -0.1in  
\caption{\textbf{Distribution of expert loadings.} The discontinuous lines represent a perfectly balanced distribution of tokens among different experts or modalities. The first figure on the left illustrates the workload among experts, while the remaining four figures depict the preferences of experts towards different modalities.}
\label{fig:expert_load_prefer}
\vskip -0.2in
\end{figure*}

\subsection{Image Understanding Evaluation}

\textbf{Zero-shot Image Question Answering.} As shown in~\cref{tab:main_res}, since MoE-LLaVA is a sparse model equipped with a soft router based on LVLM, we categorize the previous models as dense models. We evaluate the performance of MoE-LLaVA on five image question-answering benchmarks and report the number of activated parameters. Compared to the state-of-the-art method LLaVA 1.5, MoE-LLaVA demonstrates powerful image understanding capabilities and performs very close to LLaVA-1.5 on five benchmarks. Specifically, MoE-LLaVA-Phi-2.7B×4 surpasses LLaVA-1.5-7B by 2.7\% on SQA$^\text{I}$ using 3.6B sparse activated parameters. Notably, MoE-LLaVA-StableLM-1.6B×4 achieves comprehensive superiority over IDEFICS-80B with only 2.0B activated parameters. Furthermore, we observe the recent small-scale vision-language model, LLaVA-Phi. MoE-LLaVA-Phi-2.7B×4 outperforms LLaVA-Phi by more than 6.2\% on VQA$^\text{v2}$, highlighting the strong comprehension abilities of MoE-LLaVA in natural vision.

\textbf{Evaluation under Benchmark Toolkits.} To comprehensively evaluate the multi-modal understanding capabilities of MoE-LLaVA, we evaluate its performance on four benchmark toolkits. These benchmark toolkits typically involve open-ended answers, serving as tools to verify a model's ability to engage in natural language questioning. In~\cref{tab:main_res}, MoE-LLaVA-Qwen-1.8B×4 surpasses Qwen-VL-7B by 21.5\%, on MMBench, despite the latter utilizing higher image resolutions. These results collectively demonstrate that the sparse model MoE-LLaVA achieves comparable or even superior performance to dense models with fewer activated parameters.

\subsection{Object Hallucination Evaluation}
We adopt the evaluation pipeline of POPE~\cite{li2023evaluating}, a polling-based query method, to evaluate object hallucination in MoE-LLaVA. The results are presented in~\cref{tab:pope}, where MoE-LLaVA exhibits the best performance, indicating that MoE-LLaVA tends to generate objects consistent with the given image. Specifically, MoE-LLaVA-1.8B×4 surpasses LLaVA-1.5-13B by 1.0\%, 1.5\%, and 0.8\% in adversarial sampling, popular sampling, and random sampling, respectively, with 2.2B activated parameters. Additionally, we observe that the yes ratio of MoE-LLaVA remains relatively balanced, indicating that our sparse model is capable of providing accurate feedback based on the given questions.

\subsection{Quantitative Analysis}


\textbf{Routing Distributions.} In~\cref{fig:expert_load_prefer}, we present the expert loads (leftmost plot) and the modalities preferences of different experts (four subplots on the right) through MoE-LLaVA-2.7B×4-Top2 on ScienceQA. More visualization can be found in~\cref{appendix:router_dist}. To begin with, the expert loads in all MoE layers are totally balanced. However, as the model gradually becomes sparser, the expert 3 loads for layers 17 to 27 suddenly increase, and they even dominate the workload of almost all tokens. For the shallow layers (5-11), experts 2, 3, and 4 mainly collaborate. It is worth noting that expert 1 only works predominantly in the first few layers, and as the model becomes deeper, expert 1 gradually withdraws from the workload. Therefore, the experts in MoE-LLaVA have learned a certain pattern that allows them to divide their tasks in a specific manner. 

\begin{figure}[t]
\centering
    \includegraphics[width=1.0\linewidth]{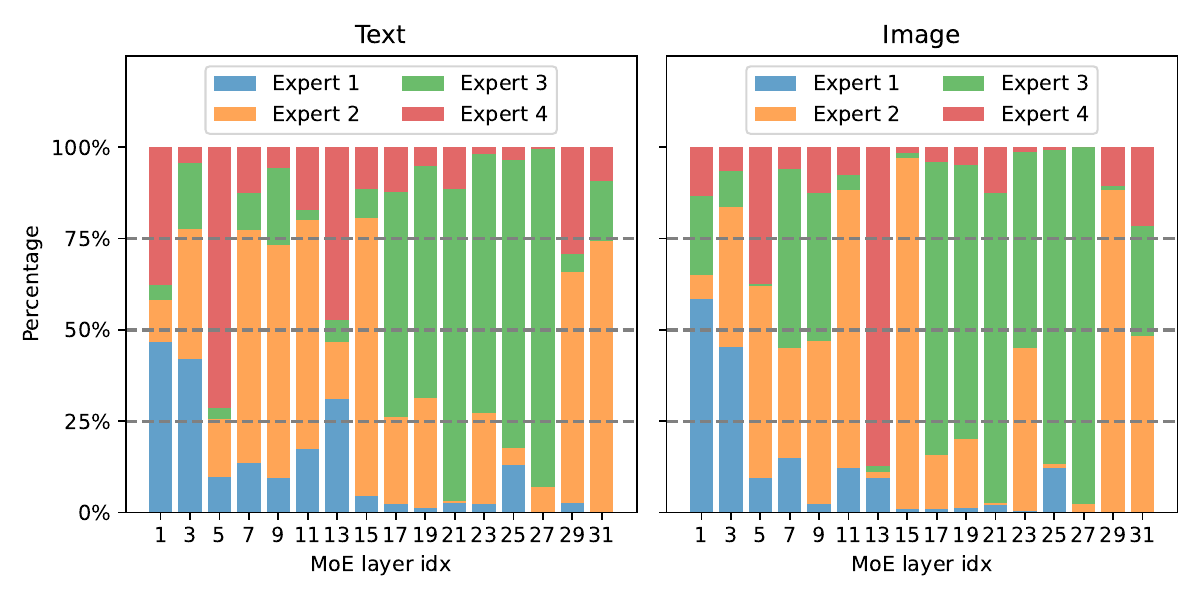}        
\vskip -0.1in
\caption{\textbf{Distribution of modalities across different experts.} Interrupted lines mean a perfectly balanced distribution of tokens.}
\label{fig:text_img_assign_to_expert}
\vskip -0.1in
\end{figure}

Furthermore, we show the distribution of modalities across different experts in~\cref{fig:text_img_assign_to_expert}. Similarly, experts develop their own preferences. Additionally, we find that the routing distributions for text and image are highly similar. For example, when expert 3 is actively working in layers 17-27, the proportions of text and image that MoE-LLaVA processes are similar. Each expert in MoE-LLaVA is capable of handling both text tokens and image tokens simultaneously, which demonstrates that MoE-LLaVA does not exhibit a clear preference for any modality. This serves as evidence of its strong interaction in multimodal learning.

\begin{figure}[t]
\vskip 0.1in
\centering
    \includegraphics[width=1.0\linewidth]{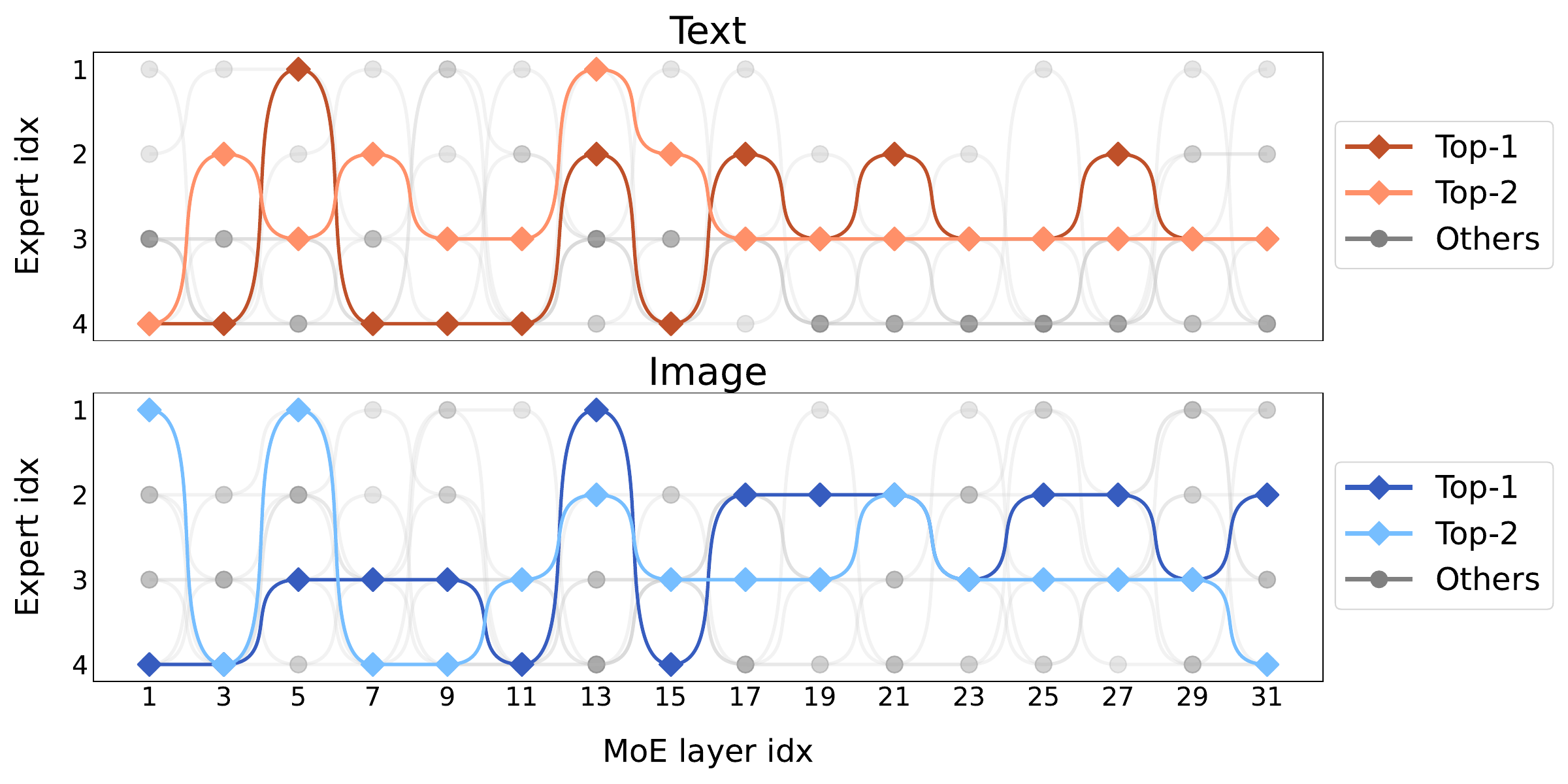}        
\vskip -0.1in
\caption{\textbf{Visualization of activated pathways.} We highlight the top-10 activated pathways on the text and image. Among them, the colorful paths represent the top-2 paths for text and image, respectively, while the gray paths represent the remaining 8 paths.}
\label{fig:bumpy_text_img}
\vskip -0.1in
\end{figure}

\textbf{Token Pathways.} Furthermore, we examine the behavior of experts at the token level. More visualization can be found in~\cref{appendix:token_pathway} and~\cref{appendix:board}. We track the trajectories of all tokens on downstream tasks. For all activated pathways, we employ PCA~\cite{pearson1901liii} to obtain the top-10 pathways, as shown in~\cref{fig:bumpy_text_img}. We found that for a given unseen text token or image tokens, MoE-LLaVA consistently tends to assign experts 2 and 3 to handle them in the deeper layers of the model.  Regarding experts 1 and 4, they tend to handle the tokens during the initialization phase. These findings contribute to a better understanding of the behavior of sparse models in multi-modal learning.

\subsection{Ablation Study}

In this section, we first validate the necessity of the three-stage training strategy. We then explore the impact of different base models and conduct ablation studies on the number of experts and active experts, and the MoE structure. We provide additional results in~\cref{appendix:capacity}.

\begin{table*}[t]
\caption{\textbf{Ablation study about training setting and architecture design decisions.} Settings for results in~\cref{tab:main_res} and~\cref{tab:pope} are highlighted in \colorbox{myblue}{blue}. We report the training time on 8 V100-32G.}
\vskip 0.1in
\label{tab:ablate_hyperparameters}
\centering
\small
    \subfloat[
	\label{tab:param}
	\textbf{Tuning the parameters of different subsets.}
	]{
		\centering
        \setlength\tabcolsep{1.35mm}
		\begin{minipage}{0.46\linewidth}{\begin{center}
              \begin{tabular}{c|ccccc|c}
                \toprule
                 \textbf{Subset} & \textbf{GQA} & \textbf{VisWiz} & \textbf{VQA$^\text{T}$ }& \textbf{POPE}  & \textbf{LLaVA$^\text{W}$} & \textbf{Time} \\
                \midrule
                \rowcolor{myblue} FFN  & \textbf{61.5} & \textbf{32.6} & \textbf{48.0} & \textbf{87.0}  & \textbf{88.7} & \textbf{20h} \\
                All & 61.3 & 31.9  & 47.6 & \textbf{87.0}  & 88.1 & 27h \\
                \bottomrule
              \end{tabular}
		\end{center}}\end{minipage}
	}
    \subfloat[
	\textbf{The number of experts.}
	\label{tab:experts}
	]{
        \setlength\tabcolsep{1.4mm}
		\begin{minipage}{0.55\linewidth}{\begin{center}
              \begin{tabular}{c|ccccc|c}
                \toprule
                \textbf{Experts} & \textbf{GQA}  & \textbf{SQA$^\text{I}$} & \textbf{VQA$^\text{T}$} & \textbf{POPE} & \textbf{LLaVA$^\text{W}$} &\textbf{Time} \\
                \midrule
                 1  & 60.9  & 60.2 & \textbf{48.3} & 86.4   & 86.3 & \textbf{13h} \\
                 \rowcolor{myblue} 2 & \textbf{61.2} & \textbf{60.8} & 47.0 & \textbf{87.5}   & \textbf{86.5} & 14h \\
                \bottomrule
              \end{tabular}
		\end{center}}\end{minipage}
	}
    \vskip -0.05in
    \subfloat[
	\textbf{The value of top-k.}
	\label{tab:topk}
	]{
        \setlength\tabcolsep{1.8mm}
		\begin{minipage}{0.43\linewidth}{\begin{center}
              \begin{tabular}{c|ccccc|c}
                \toprule
                 \textbf{Top-k}& \textbf{VQA$^\text{v2}$}  & \textbf{GQA} & \textbf{SQA$^\text{I}$} & \textbf{VQA$^\text{T}$} & \textbf{POPE} & \textbf{Time} \\
                \midrule
                1 & 74.5 & 58.4 & 58.0 & 44.0 & 85.7 & \textbf{19h} \\
                \rowcolor{myblue} 2 & \textbf{76.2} & \textbf{61.5} & \textbf{63.1}  & \textbf{48.0} & \textbf{88.7} & 20h \\
                \bottomrule
              \end{tabular}
		\end{center}}\end{minipage}
	}
    \subfloat[
	\textbf{The architectures of MoE-LLaVA.}
	\label{tab:arch}
	]{
        \setlength\tabcolsep{1.2mm}
		\begin{minipage}{0.61\linewidth}{\begin{center}
              \begin{tabular}{c|ccccc|c}
                \toprule
                 \textbf{Architecture}& \textbf{VQA$^\text{v2}$}  & \textbf{GQA} & \textbf{SQA$^\text{I}$} & \textbf{VQA$^\text{T}$}  & \textbf{POPE} & \textbf{Time} \\
                \midrule
                 First-Half & 75.9 &  61.3 & 62.4 & 47.0 & 86.9  & \textbf{20h} \\
                 Second-Half & \textbf{76.3} & 61.2 & 62.6  & 47.2 & 86.9 &  \textbf{20h} \\
                 \rowcolor{myblue} Interval & 76.2 & 61.5 & \textbf{63.1}  & \textbf{48.0} & \textbf{88.7} & \textbf{20h} \\
                 All & 74.5 & \textbf{61.5} & 62.1  & 47.1 & 87.0 & 32h \\
                \bottomrule
              \end{tabular}
		\end{center}}\end{minipage}
	}
\vskip -0.3in
\end{table*}

\begin{table}[t]
  \small
  \setlength\tabcolsep{0.9mm}
  \caption{Ablation study about \textbf{different training strategies.} ``LA'' and ``Hb'' represent LLaVA-FT and Hybrid-FT in~\cref{tab:data}.}
  \vskip 0.1in
  \label{tab:init}
  \centering
  \begin{tabular}{c|ccc|cccc}
    \toprule
      & \textbf{MoE} & \textbf{Stage \uppercase\expandafter{\romannumeral2}} & \textbf{Stage \uppercase\expandafter{\romannumeral3}} & \textbf{GQA} & \textbf{SQA$^\text{I}$} & \textbf{POPE} & \textbf{LLaVA$^\text{W}$} \\
    
     
    \midrule
    
     (a) & \ding{52} & -  & LV+Hb & 58.4 & 58.1 & 81.9 & 88.0 \\
     \rowcolor{myblue} (b) & \ding{52} & Hb & LV & \textbf{61.5} & \textbf{63.1} & \textbf{87.0} & \underline{88.7} \\
     (c) & \ding{55} & LV+Hb & - & 60.9 & 60.2 & 86.4 & 86.3 \\
     (d) & \ding{55} & Hb & LV & \underline{60.9} & \underline{62.5} & \underline{86.9} & \textbf{90.1} \\
    \bottomrule
  \end{tabular}
\vskip -0.1in
\end{table}

\begin{table}[t]
  \small
  \setlength\tabcolsep{1.2mm}
  \caption{Ablation study about \textbf{the model size of MoE-LLaVA.} }
  \vskip 0.1in
  \label{tab:scale}
  \centering
  \begin{tabular}{cc|ccccc}
    \toprule
     \textbf{Model}& \textbf{MoE}&  \textbf{VQA$^\text{v2}$}  & \textbf{SQA$^\text{I}$} & \textbf{VQA$^\text{T}$}  & \textbf{MMB}&   \textbf{LLaVA$^\text{W}$}\\
    \midrule
     \multirow{2}{*}{StableLM}  & \ding{55} & 74.5 & 62.0  & 48.8 & 58.2 &83.2\\
       & \cellcolor{myblue}\ding{52} & \cellcolor{myblue}\textbf{76.7} & \cellcolor{myblue}\textbf{62.6}  &\cellcolor{myblue}\textbf{50.1} & \cellcolor{myblue}\textbf{60.2} & \cellcolor{myblue}\textbf{86.8}\\
    \midrule
     \multirow{2}{*}{Qwen}  & \ding{55} & 74.9 &  60.2 & \textbf{48.3} & \textbf{60.6} & 86.3\\
       & \cellcolor{myblue}\ding{52} & \cellcolor{myblue}\textbf{76.2} & \cellcolor{myblue}\textbf{63.1}  &\cellcolor{myblue}48.0 & \cellcolor{myblue}59.7 & \cellcolor{myblue}\textbf{88.7}\\
    \midrule
     \multirow{2}{*}{Phi-2}  & \ding{55} & 75.6 &  67.8 & 50.0 & 65.0 & 91.3\\
       & \cellcolor{myblue}\ding{52} & \cellcolor{myblue}\textbf{77.6} & \cellcolor{myblue}\textbf{68.5}  & \cellcolor{myblue}\textbf{51.4} &  \cellcolor{myblue}\textbf{65.2} & \cellcolor{myblue}\textbf{94.1}\\
    \bottomrule
  \end{tabular}
\vskip -0.2in
\end{table}

\textbf{Effect of Training Strategy.} In~\cref{tab:init}, we conduct three variant experiments to demonstrate the rationale behind using the second-stage instruction tuning as the initialization for the third-stage MoE tuning. When adapting MoE to LVLMs, a straightforward approach is to replace the classic LLaVA's FFN with a MoE layer and train it according to the original second-stage script, denoted as variant (a). However, variant (a) performs the worst, suggesting that the current multi-modal instruction dataset is insufficient to support both the conversion from LLM to LVLM and the conversion from LVLM to a sparse model simultaneously. Therefore, we collect more data, referred to as Hybrid-FT, and initially convert LLM to LVLM in the second stage. Subsequently, in the third stage, LVLM is sparsified by using the LLaVA-FT dataset, resulting in variant (b). Additionally, we expand the data of the original LLaVA's second stage for fair comparison, denoted as variant (c). The results indicate that variants (b) outperformed variants (a) and (c). These findings demonstrate that providing a reasonable LVLM initialization allows the model to transition rapidly from a dense model to a sparse model, validating the principle behind our three-stage training strategy.

\textbf{Effect of Tuning the Parameters of Different Subsets.} In~\cref{tab:param}, we examine the performance of fine-tuning different parts of the parameters. ``FFN'' represents fine-tuning all FFN layers and MoE layers in the model. ``All'' indicates fine-tuning all parameters. The results indicate tuning the FFN is sufficient to achieve results comparable to full-parameter tuning, but it requires only approximately 75\% of the time. Therefore, to enhance generalization and reduce training costs, we only fine-tune FFN layers.

\textbf{Effect of the Number of Experts.} Typically, increasing the number of experts directly leads to higher performance~\cite{lepikhin2020gshard, fedus2022switch}. In~\cref{tab:experts}, we change the number of experts while keeping the number of activated experts the same, so the number of activated parameters for both models remains the same. More sparse experts outperform the single expert dense model by 1.1\% on POPE and 0.6\% on SQA$^\text{I}$, respectively. The results demonstrate that sparse experts can deliver superior performance.

\textbf{Effect of the Number of Activated Experts.} To evaluate the effect of the number of activated experts, we compare the performance of using different top-$k$ strategies. With the number of activated experts changing from 1 to 2, it brings a significant improvement with only 1h training time increasing.
These results show that activating more experts can improve the MOE-LLaVA ability.
To leverage the advantages of the MoE scheme, we set the number of activated experts to 2.

\textbf{Effect of the Architectures.} In~\cref{tab:arch}, we explore four variations of MoE architecture. Specifically, ``First-Half'' indicates that MoE layers are applied only to the first half of the model while the second half retains the original dense architecture. ``Second-Half'' means that MoE layers are placed in the second half of the model while the first half remains dense. ``Interval'' represents alternating occurrences of MoE layers and dense layers. ``All'' indicates that all layers are sparse MoE layers. Intuitively, it is expected that incorporating all MoE will enhance performance. However, using ``All'' does not yield better results and results in longer training times compared to other architectures. Therefore, MoE-LLaVA alternates the insertion of MoE layers.

\textbf{Effect of the Model Size.} As shown in~\cref{tab:scale}, we compare the performance of models with different parameter sizes as the foundation models for MoE-LLaVA. For smaller models such as Phi2-MoE and Qwen-MoE, the performance with MoE surpasses that of dense models. We provide additional results in~\cref{appendix:model_scale}.

\section{Conclusion and Future Directions}
\label{sec:conclu}
In this work, we propose the MoE-Tuning to adapting the MoE architecture to LVLMs, and construct the MoE-based spare model MoE-LLaVA, which can find a sparse pathway by simultaneously handling image and text features. Our framework demonstrates strong ability of multi-modal understanding and rich potential for hallucination inhibition, achieving comparable performance of LLaVA-1.5-7B with only 3B activated parameters.


While MoE-LLaVA demonstrates competitive capabilities, we observe some difficulties in training stability, particularly with 16-bit float precision. Furthermore, due to the presence of multiple experts specializing in different abilities, MoE-LLaVA can easily be expanded to handle additional tasks such as detection, segmentation, generation, or handling more modalities such as video, depth, and thermal. 


\clearpage

\section*{Impact Statements}

\subsection*{Broader Impacts}
While MoE-LLaVA holds great potential and application value in multi-modal understanding, it may also have some negative social impacts:
\begin{itemize}
\item Information credibility: MoE-LLaVA can generate realistic texts, including false information and misleading content.
\item Bias and discrimination: The training data for MoE-LLaVA often comes from the internet, where various biases and discriminatory content may exist. If these unequal patterns are learned and amplified by the model, they may be reflected in the generated responses.
\item Social influence: People may become overly reliant on MoE-LLaVA for information and problem-solving, instead of actively thinking and seeking multiple sources of information. This can lead to increased dependency, reduced autonomy in thinking, and judgment skills.
\end{itemize}

\subsection*{Reproducibility}

In~\cref{appendix:train_detail}, we have provided a detailed list of all the training hyperparameters. We have open-sourced all models and codes. Reproducibility can be achieved by using the code provided in the materials. 

\subsection*{Compute}
For the main results, we conducte experiments on 8 A800-80G. For the ablation study, we measure the time on 8 V100-32G.

\subsection*{Licenses}

The majority of this project is released under the Apache 2.0 license.

\begin{itemize}
\item The service is a research preview intended for non-commercial use only, subject to the model License of LLaMA (\url{https://github.com/facebookresearch/llama/blob/main/MODEL_CARD.md}). 
\item Terms of Use of the data generated by OpenAI (\url{https://openai.com/policies/terms-of-use}). 
\item Privacy Practices of ShareGPT (\url{https://chrome.google.com/webstore/detail/sharegpt-share-your-chatg/daiacboceoaocpibfodeljbdfacokfjb}). 
\end{itemize}

\nocite{langley00}

\bibliography{example_paper}
\bibliographystyle{icml2024}

\newpage
\appendix
\onecolumn

\def\apptoptitlebar{\hrule height0pt}
\def\appbottomtitlebar{\vskip .22in }

{\center\baselineskip 18pt
   \apptoptitlebar{\Large\textbf{Appendix for MoE-LLaVA}}\appbottomtitlebar}
                       
\section{Implementation Details}

\subsection{More Model Architecture} 
\label{appendix:more_model}

In~\cref{tab:more_model}, we present additional variants of the MoE-LLaVA. We introduce how the total parameters is calculated. When the number of activated experts is 2, setting $Experts=2$ yields the number of activated parameters.

\vskip -0.15in
\begin{equation}
\begin{aligned}
    Total\text{\_}Parameters = &Embedding \cdot Width \\
    &+ Layers \cdot (4 \cdot Width \cdot Width + Width \cdot FFN \cdot FFN\text{\_}Factor + 2 \cdot Width) \\
    &+ Width + Width \cdot Embedding \\
    &+ MoE\text{\_}Layers \cdot (Experts - 1) \cdot (Width \cdot FFN \cdot FFN\text{\_}Factor + 2 \cdot Width) \\
    &+ MoE\text{\_}Layers \cdot (Width \cdot Experts)
\end{aligned}
\end{equation}
\vspace{-0.5cm}

\begin{table}[h]
  \small
  \setlength\tabcolsep{0.85mm}
  \caption{\textbf{More architecture details of the MoE-LLaVA model.} ``FFN Factor`` represents the number of linear layers in the FFN. ``*'' donates the dimension of the hidden states for the keys (k) and values (v) is 1024. ``1.6B×4-Top2'' represents a dense foundation model with 1.6B parameters, which will be equipped with a total of four experts, with two of them being activated. ``$\dag$'' donates all layers will equipped with MoE layer.}
  \vskip 0.1in
  \label{tab:more_model}
  \centering
  \begin{tabular}{l|ccc|cccccc|cc}
    \toprule
    \multirow{2}{*}{\textbf{Name}} & \multirow{2}{*}{\textbf{Experts}} & \multirow{2}{*}{\textbf{Top-k}} & \textbf{MoE} & \multirow{2}{*}{\textbf{Embedding}} & \multirow{2}{*}{\textbf{Width}} & \multirow{2}{*}{\textbf{Layers}} & \multirow{2}{*}{\textbf{FFN}} & \textbf{FFN} & \multirow{2}{*}{\textbf{Heads}} & \textbf{Activated} & \textbf{Total} \\
     &  &  & \textbf{Layers} & & &  & & \textbf{Factor} &  & \textbf{Param} & \textbf{Param} \\
    \midrule
    StableLM-1.6B~\cite{StableLM-2-1.6B} & - & - & - & 100352 & 2560 & 32 & 10240 & 2 & 32 & 1.6B & 1.6B \\
    MoE-LLaVA-1.6B×4-Top2 & 4 & 2 & 16 & 100352 & 2560 & 32 & 10240 & 2 & 32 & 2.0B & 2.9B \\
    MoE-LLaVA-1.6B×4-Top2$^\dag$ & 4 & 2 & 32 & 100352 & 2560 & 32 & 10240 & 2 & 32 & 2.5B & 4.1B \\
    \midrule
    Qwen-1.8B~\cite{bai2023qwen} & - & - & - & 151936 & 2048 & 24 & 5504 & 3 & 16 & 1.8B & 1.8B \\
    MoE-LLaVA-1.8B×4-Top2 & 4 & 2 & 12 & 151936 & 2048 & 24 & 5504 & 3 & 16 & 2.2B & 3.1B \\
    MoE-LLaVA-1.8B×4-Top2$^\dag$ & 4 & 2 & 24 & 151936 & 2048 & 24 & 5504 & 3 & 16 & 2.6B & 4.3B \\
    \midrule
    Phi2-2.7B~\cite{phi2} & - & - & - & 51200 & 2560 & 32 & 10240 & 2 & 32 & 2.7B & 2.7B \\
    MoE-LLaVA-2.7B×4-Top2 & 4 & 2 & 16 & 51200 & 2560 & 32 & 10240 & 2 & 32 & 3.6B & 5.3B \\
    MoE-LLaVA-2.7B×4-Top2$^\dag$ & 4 & 2 & 32 & 51200 & 2560 & 32 & 10240 & 2 & 32 & 4.5B & 7.8B \\
    \midrule
    OpenChat-7B~\cite{wang2023openchat} & - & - & - & 32000 & 4096$^*$ & 32 & 14336 & 3 & 32 & 6.7B & 6.7B \\
    MoE-LLaVA-7B×4-Top2 & 4 & 2 & 16 & 32000 & 4096$^*$ & 32 & 14336 & 3 & 32 & 9.6B & 15.2B \\
    MoE-LLaVA-7B×4-Top2$^\dag$ & 4 & 2 & 32 & 32000 & 4096$^*$ & 32 & 14336 & 3 & 32 & 12.4B & 23.7B \\
    \bottomrule
  \end{tabular}
\vskip -0.1in
\end{table}

\begin{wraptable}{r}{0.45\textwidth}
  \small
  \vskip -0.51in
  \setlength\tabcolsep{0.75mm}
  \caption{\textbf{Training hyperparameters.}}
  \vskip 0.1in
  \label{tab:train_detail}
  \centering
  \begin{tabular}{l|ccc}
    \toprule
            Config & Stage \uppercase\expandafter{\romannumeral1} & Stage \uppercase\expandafter{\romannumeral2} & Stage \uppercase\expandafter{\romannumeral3} \\
            \midrule
            Experts & -  & - & 4 \\
            Top-k & -  & - & 2 \\
            \midrule
            Deepspeed & Zero2  & Zero2 & Zero2\_offload \\
            Data & LLaVA-PT  & Hybird-PT & LLaVA-FT\\
            Image resolution & \multicolumn{3}{c}{336×336} \\
            Image encoder & \multicolumn{3}{c}{CLIP-Large/336} \\
            Feature select layer & \multicolumn{3}{c}{-2} \\
            Image projector & \multicolumn{3}{c}{2 Linear layers with GeLU} \\
            Epoch & \multicolumn{3}{c}{1} \\
            Learning rate &  1e-3 & 2e-5 & 2e-5 \\
            Learning rate schdule & \multicolumn{3}{c}{Cosine} \\
            Weight decay & \multicolumn{3}{c}{0.0}  \\
            Text max length & \multicolumn{3}{c}{2048} \\
            Batch size per GPU &  32 & 16 & 16 \\
            GPU & \multicolumn{3}{c}{8 × A800-80G}  \\
            Precision & \multicolumn{3}{c}{Bf16} \\
    \bottomrule
  \end{tabular}
\vskip -0.1in
\end{wraptable}

\subsection{Training Details} 
\label{appendix:train_detail}

As shown in~\cref{tab:train_detail}, we present the training hyperparameters for all models, which are applicable to Qwen, StableLM, Phi and OpenChat. For the training process in all stages, we consistently train for 1 epoch, as we find that the models overfit when training for 2 epochs. The batch size for the first stage is 256 and 128 for the second and third stages. We use an image resolution of 336x336 for all three stages. \textbf{Additionally, for smaller models like Qwen-1.8B, it is feasible to train them on 8 V100-32G GPUs. However, during the training process, using fp16 may sometimes lead to loss becoming NaN.} Since our models are smaller than 7B, we can train them in \textit{zero2} mode. However, for stage 3, deepspeed temporarily does not support training MoE architecture in \textit{zero3} mode. Therefore, we choose \textit{zero2\_offload} to further reduce the memory requirements and enable running on 8 A800-80G GPUs. We enable the gradient checkpoint mode for all training stage.


\section{Additional Results and Visualization}

\subsection{Model Scaling}
\label{appendix:model_scale}

\begin{wraptable}{r}{0.55\textwidth}
  \small
  \vskip -0.5in
  \setlength\tabcolsep{1.9mm}
  \caption{Ablation study about \textbf{the model size of MoE-LLaVA.}}
  \vskip 0.1in
  \label{tab:openchat}
  \centering
  \begin{tabular}{cc|ccccc}
    \toprule
     \textbf{Model}& \textbf{MoE}&  \textbf{VQA$^\text{v2}$}  & \textbf{SQA$^\text{I}$} & \textbf{VQA$^\text{T}$}  & \textbf{MMB}&   \textbf{LLaVA$^\text{W}$} \\
    \midrule
     \multirow{2}{*}{StableLM}  & \ding{55} & 74.5 & 62.0  & \textbf{48.8} & 58.2 &83.2 \\
       & \cellcolor{myblue}\ding{52} &\cellcolor{myblue} \textbf{76.0} & \cellcolor{myblue}\textbf{62.6}  &\cellcolor{myblue} 47.8 & \cellcolor{myblue}\textbf{59.4} & \cellcolor{myblue}\textbf{85.9}\\
    \midrule
     \multirow{2}{*}{Qwen}  & \ding{55} & 74.9 &  60.2 & \textbf{48.3} & \textbf{60.6} & 86.3\\
       & \cellcolor{myblue}\ding{52} & \cellcolor{myblue}\textbf{76.2} & \cellcolor{myblue}\textbf{63.1}  &\cellcolor{myblue}48.0 & \cellcolor{myblue}59.7 & \cellcolor{myblue}\textbf{88.7}\\
    \midrule
    \multirow{2}{*}{Phi-2}  & \ding{55} & 75.6 &  67.8 & 50.0 & 65.0 & 91.3\\
       & \cellcolor{myblue}\ding{52} & \cellcolor{myblue}\textbf{77.6} & \cellcolor{myblue}\textbf{68.5}  & \cellcolor{myblue}\textbf{51.4} & \cellcolor{myblue}\textbf{65.2} & \cellcolor{myblue}\textbf{94.1}\\
    \midrule
    \midrule
     \multirow{2}{*}{OpenChat} & \cellcolor{myblue}\ding{55} & \cellcolor{myblue}77.9  & \cellcolor{myblue}\textbf{69.0} & \cellcolor{myblue}\textbf{54.7} & \cellcolor{myblue}\textbf{66.9} &\cellcolor{myblue}\textbf{89.7} \\
      & \ding{52} & \textbf{78.9} & 62.8 & 52.5 &  65.9 & 86.3\\
    \bottomrule
  \end{tabular}
\vskip -0.1in
\end{wraptable}

As shown in~\cref{tab:openchat}, for models smaller than 7B, we demonstrate a strong scale of law. MoE-LLaVA exhibits improved performance as the model size increases, as exemplified by StableLM-1.6B, Qwen-1.8B, and Phi-2.7B. But surprisingly, the overall performance of OpenChat-MoE is significantly inferior to dense models. We speculate that this may be due to the insufficient data for current multi-modal instruction tuning to support sparse pattern learning in 10B-level models, which should be addressed in future work when scaling up to larger MoE-LLaVA models.

\subsection{Training Capacity}
\label{appendix:capacity}

For MoE layers, we employ the Batch Priority Routing (BPR) strategy~\cite{riquelme2021scaling}. This strategy utilizes the routing results to determine which tokens should be dropped, ensuring a more balanced workload among the experts. During the training process, the BPR strategy dynamically adjusts the routing results for each expert based on their capacity. When the tokens assigned to an expert exceed its predefined capacity, the excess tokens are dropped. We conduct a ablation study on the hyperparameter capacity, as shown in~\cref{tab:capacity}. Increasing the capacity consistently improves performance for different sizes of MoE-LLaVA.

\vskip -0.2in
\begin{table*}[h]
\small
  \setlength\tabcolsep{1.33mm}
\caption{Ablation study about \textbf{the capacity of MoE-LLaVA.} ``Res.'' represent the input image resolution. $^*$ donates that there is some overlap in the training data.}
\vskip 0.1in
  \label{tab:capacity}
  \centering
  \begin{tabular}{l|cc|ccccc|cccc|c}
    \toprule
     \multirow{2}{*}{\textbf{Methods}}& \multirow{2}{*}{\textbf{Res.}}  & \multirow{2}{*}{\textbf{Capacity}}  & \multicolumn{5}{c|}{\textbf{Image Question Answering}} & \multicolumn{4}{c}{\textbf{Benchmark Toolkit}} &  \\
      &   &  & VQA$^\text{v2}$ & GQA & VisWiz & SQA$^\text{I}$ & VQA$^\text{T}$ & POPE & MMB & LLaVA$^\text{W}$ & MM-Vet & Avg\\
    \midrule
    \multirow{2}{*}{MoE-LLaVA-1.6B×4-Top2} & \multirow{2}{*}{336}  & \cellcolor{myblue}\textbf{1.5} & \cellcolor{myblue}\textbf{76.7$^*$} & \cellcolor{myblue}{60.3$^*$} & \cellcolor{myblue}{36.2} & \cellcolor{myblue}\textbf{62.6} & \cellcolor{myblue}\textbf{50.1} & \cellcolor{myblue}\textbf{85.7} & \cellcolor{myblue}\textbf{60.2} & \cellcolor{myblue}\textbf{86.8} & \cellcolor{myblue}\textbf{26.9} & \cellcolor{myblue}\textbf{60.6} \\
     &  & 1.0 & {76.0$^*$} & \textbf{60.4$^*$} & \textbf{37.2} & \textbf{62.6} & {47.8} & {84.3} & {59.4} & {85.9} & {26.1} & 59.9 \\
    \midrule
    \multirow{2}{*}{MoE-LLaVA-2.7B×4-Top2} & \multirow{2}{*}{336}  & \cellcolor{myblue}\textbf{1.5} & \cellcolor{myblue}\textbf{77.6$^*$} & \cellcolor{myblue}\textbf{61.4$^*$} & \cellcolor{myblue}\textbf{43.9} & \cellcolor{myblue}{68.5} & \cellcolor{myblue}\textbf{51.4} & \cellcolor{myblue}\textbf{86.3} & \cellcolor{myblue}{65.2} & \cellcolor{myblue}\textbf{94.1} & \cellcolor{myblue}\textbf{34.3} & \cellcolor{myblue}\textbf{64.7} \\
     &    & 1.0 & {77.1$^*$} & {61.1$^*$} & {43.4} & \textbf{68.7} & {50.2} & {85.0} & \textbf{65.5} & {93.2} & {31.1} & 63.9\\
    \midrule
    \multirow{2}{*}{MoE-LLaVA-2.7B×4-Top2} & \multirow{2}{*}{384}  & \cellcolor{myblue}\textbf{1.5} & \cellcolor{myblue}\textbf{79.9$^*$} & \cellcolor{myblue}{62.6$^*$} & \cellcolor{myblue}\textbf{43.7} & \cellcolor{myblue}\textbf{70.3} & \cellcolor{myblue}\textbf{57.0} & \cellcolor{myblue}\textbf{85.7} & \cellcolor{myblue}\textbf{68.0} & \cellcolor{myblue}\textbf{97.3} & \cellcolor{myblue}\textbf{35.9} & \cellcolor{myblue}\textbf{66.7} \\
     &   & 1.0 & {79.4$^*$} & \textbf{62.7$^*$} & {42.1} & \textbf{70.3} & {55.7} & {85.5} & {67.9} & {95.1} & {33.6} & 65.8 \\
    \bottomrule
  \end{tabular}
\vskip -0.08in
\end{table*}

\subsection{Routing Distributions}
\label{appendix:router_dist}
In this section, we present the routing distributions of MoE-LLaVA-OpenChat-7B×4-Top2, MoE-LLaVA-Phi-2.7B×4-Top2, MoE-LLaVA-Qwen-1.8B×4-Top2, and MoE-LLaVA-StableLM-1.6B×4-Top2 on six benchmarks (ScienceQA-IMG~\cite{lu2022learn}, TextVQA~\cite{singh2019towards}, POPE~\cite{li2023evaluating}, MMBench~\cite{liu2023mmbench}, VisWiz~\cite{gurari2018vizwiz}, MM-Vet~\cite{yu2023mm}). These routing distributions are based on the training up to the final checkpoint.

\begin{figure}[h]
	\centering
	\subfloat[ScienceQA-IMG]{\includegraphics[width=.49\columnwidth]{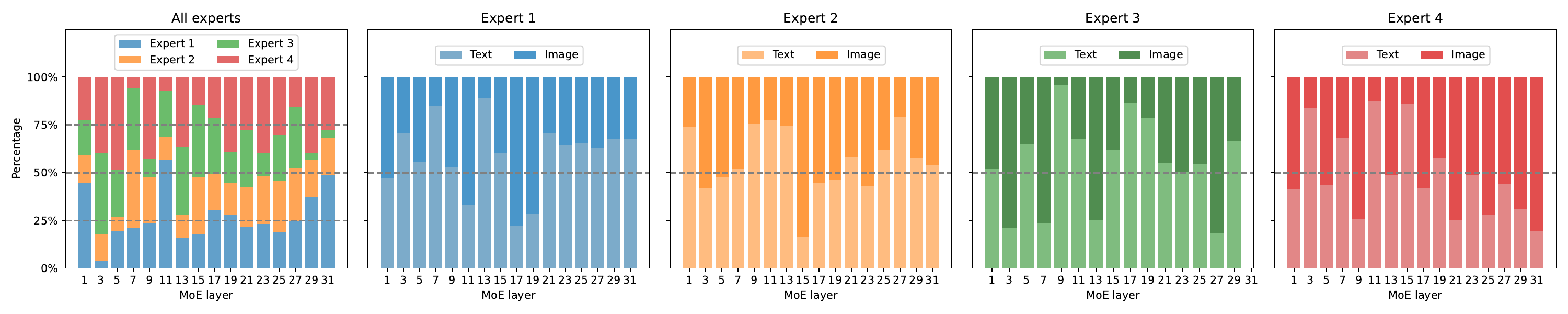}}\hspace{5pt}
	\subfloat[TextQA]{\includegraphics[width=.49\columnwidth]{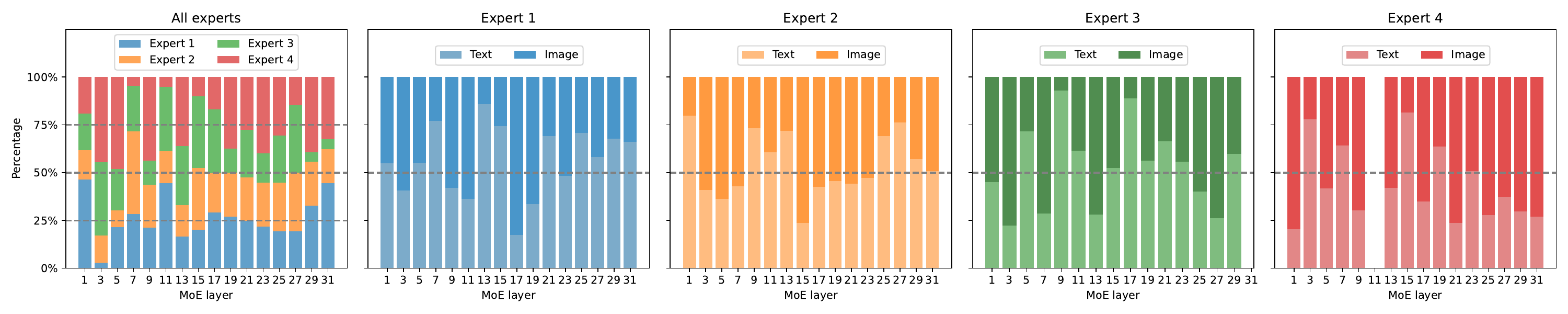}}\\
    \vskip -0.1in
	\subfloat[POPE]{\includegraphics[width=.49\columnwidth]{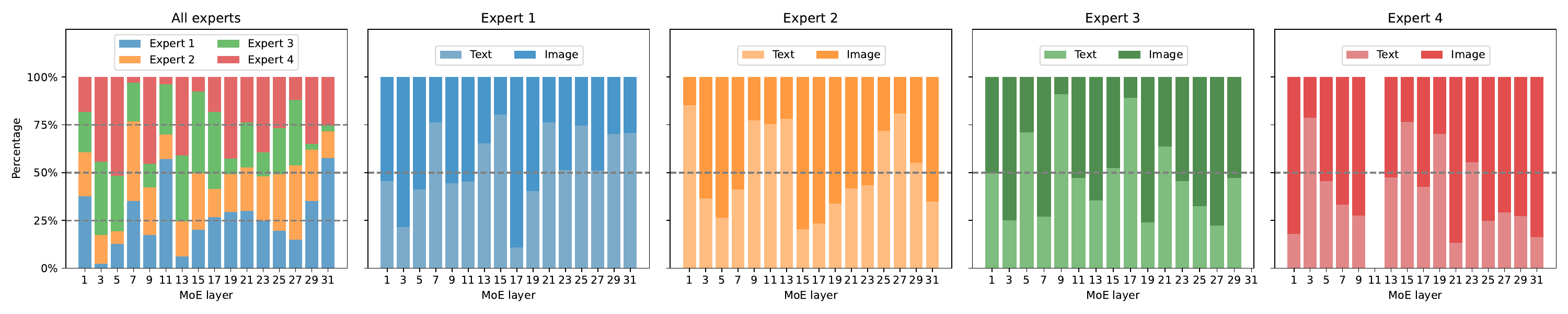}}\hspace{5pt}
	\subfloat[MMBench]{\includegraphics[width=.49\columnwidth]{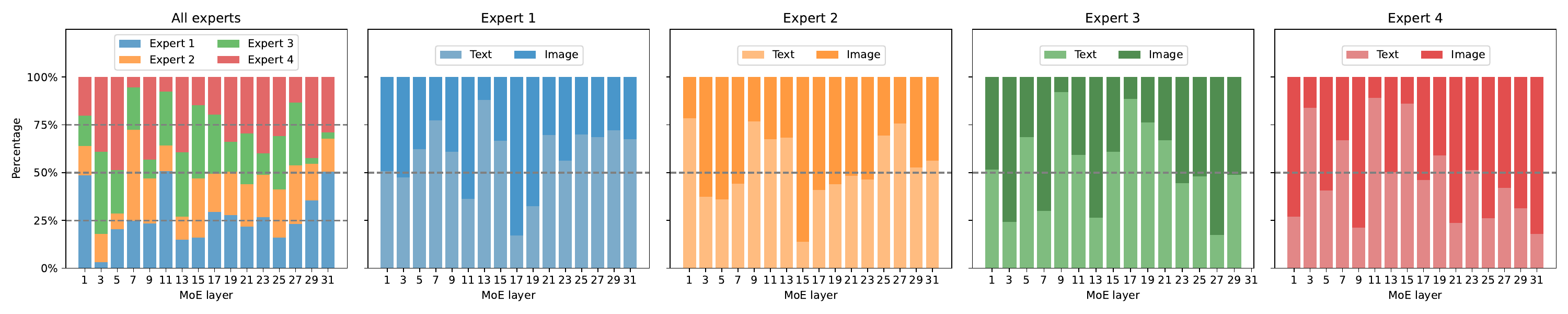}}\\
    \vskip -0.1in
	\subfloat[Viswiz]{\includegraphics[width=.49\columnwidth]{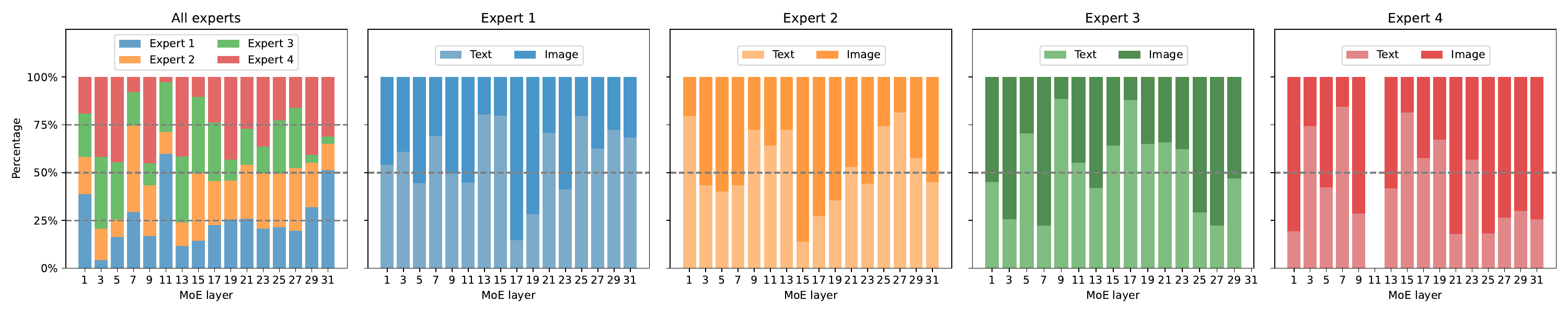}}\hspace{5pt}
	\subfloat[MM-Vet]{\includegraphics[width=.49\columnwidth]{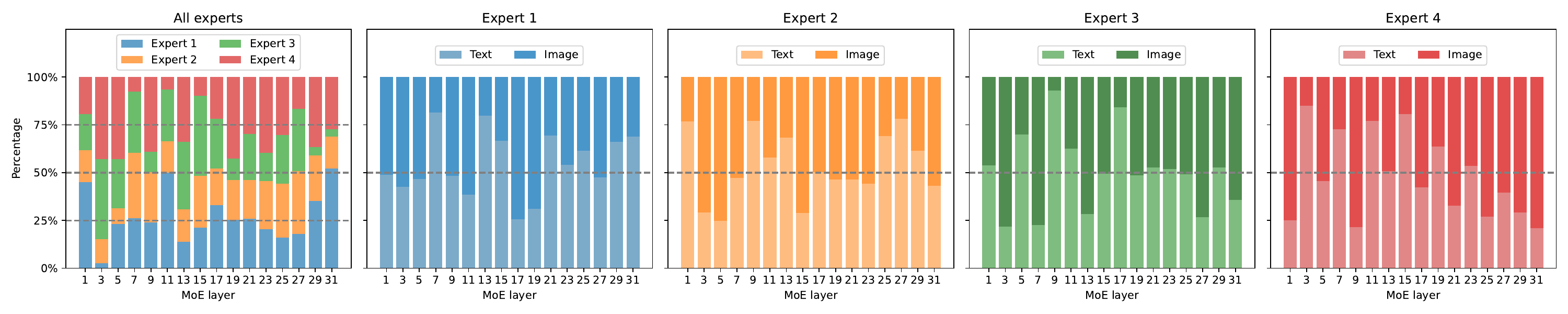}}
\vskip -0.1in
	\caption{Distribution of expert loadings and expert preferences on \textbf{MoE-LLaVA-OpenChat-7B×4-Top2}.}
        \label{fig:expert_text_img_dist_openchat}
\vskip -0.2in
\end{figure}

\begin{figure}[h]
	\centering
	\subfloat[ScienceQA-IMG]{\includegraphics[width=.49\columnwidth]{fig/expert_text_img_dist/phi_sciqa.pdf}}\hspace{5pt}
	\subfloat[TextQA]{\includegraphics[width=.49\columnwidth]{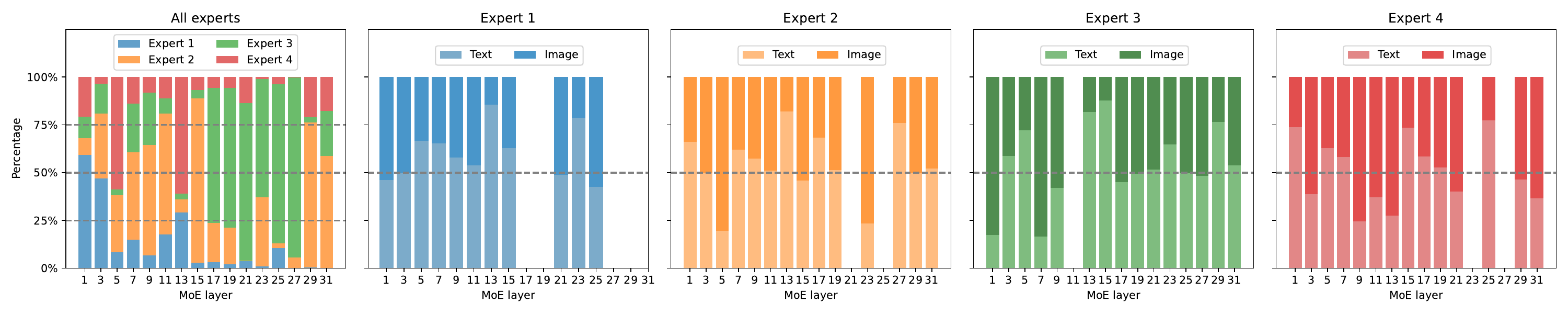}}\\
    \vskip -0.1in
	\subfloat[POPE]{\includegraphics[width=.49\columnwidth]{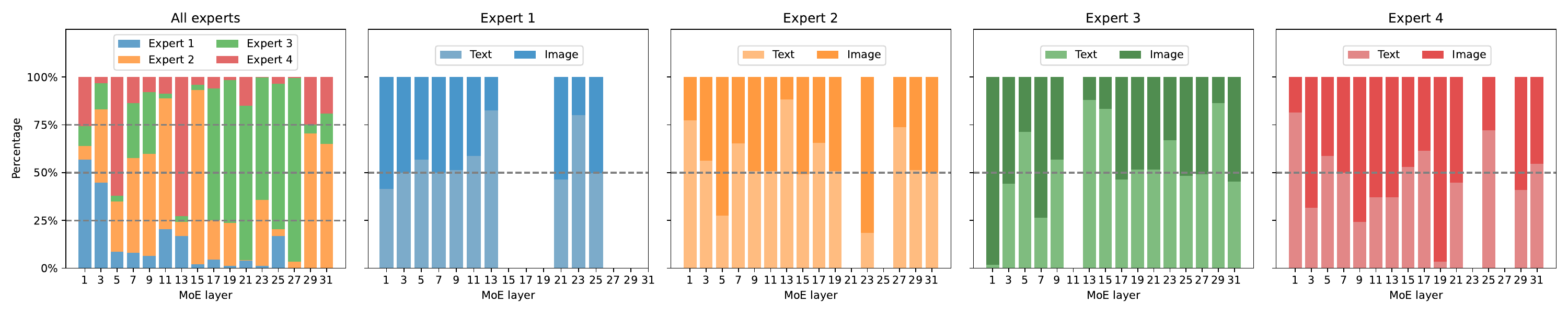}}\hspace{5pt}
	\subfloat[MMBench]{\includegraphics[width=.49\columnwidth]{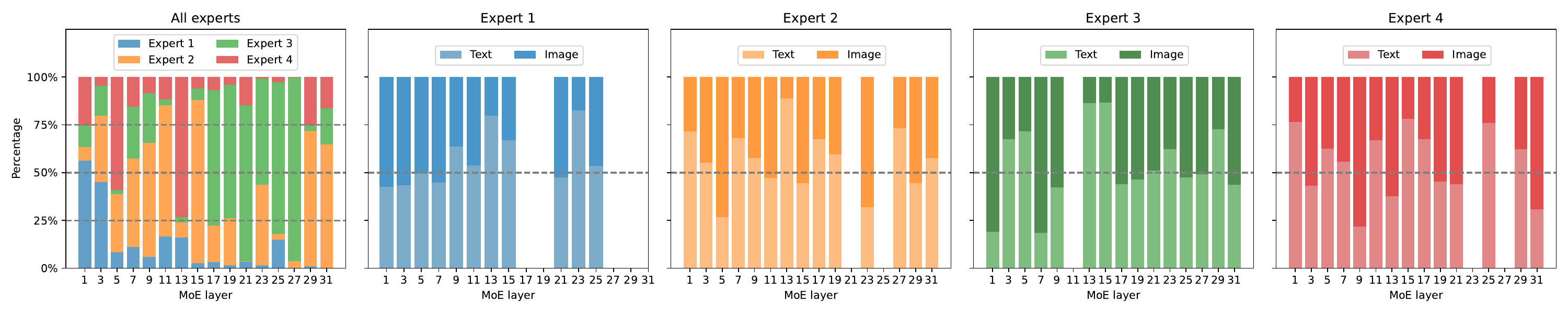}}\\
    \vskip -0.1in
	\subfloat[Viswiz]{\includegraphics[width=.49\columnwidth]{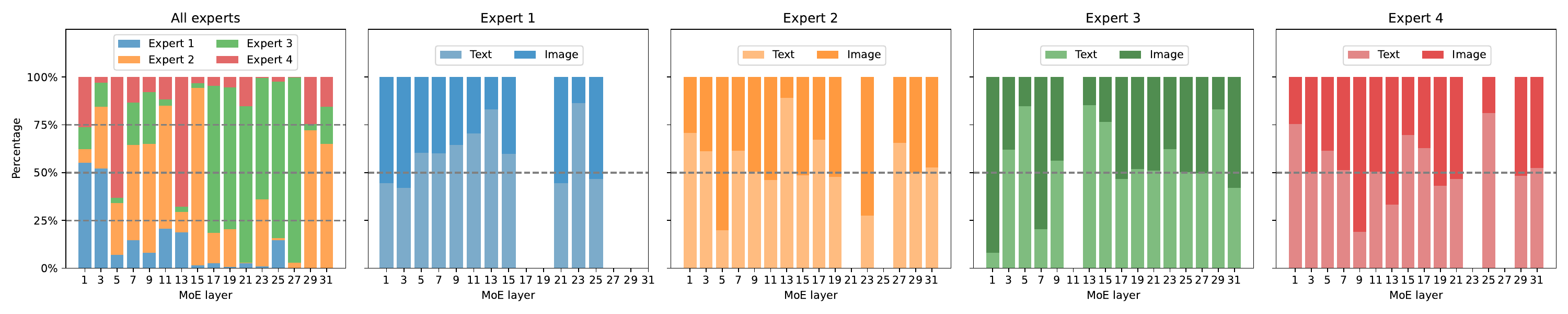}}\hspace{5pt}
	\subfloat[MM-Vet]{\includegraphics[width=.49\columnwidth]{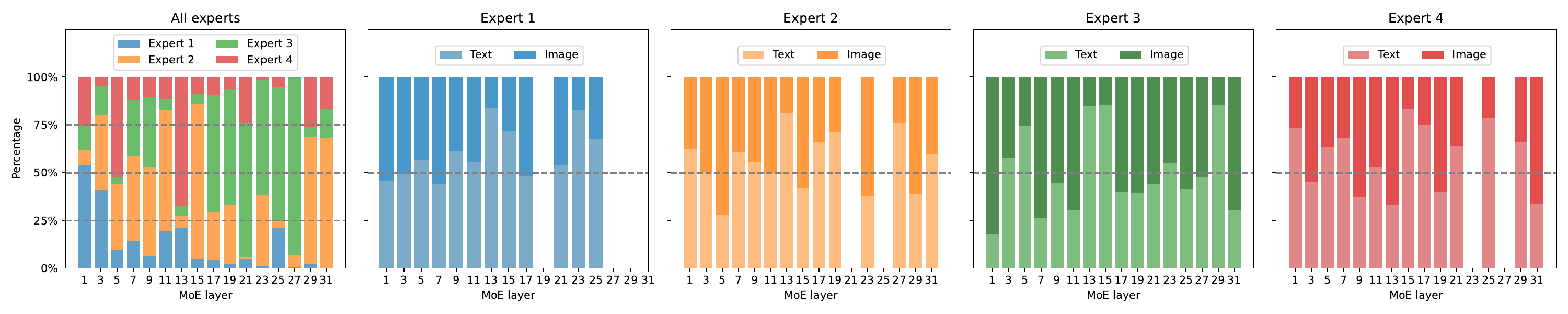}}
\vskip -0.1in
	\caption{Distribution of expert loadings and expert preferences on \textbf{MoE-LLaVA-Phi-2.7B×4-Top2}.}
        \label{fig:expert_text_img_dist_phi}
\vskip -0.2in
\end{figure}

\begin{figure}[h]
	\centering
	\subfloat[ScienceQA-IMG]{\includegraphics[width=.49\columnwidth]{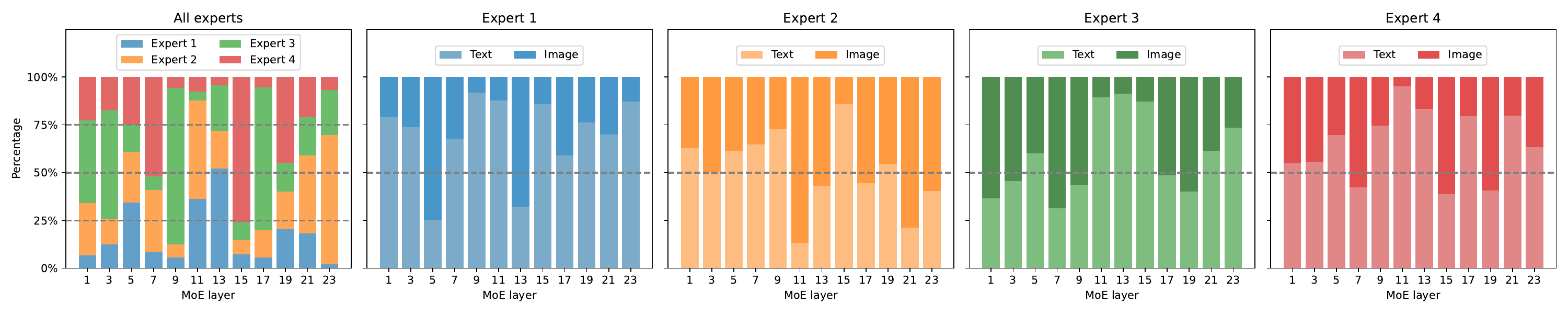}}\hspace{5pt}
	\subfloat[TextQA]{\includegraphics[width=.49\columnwidth]{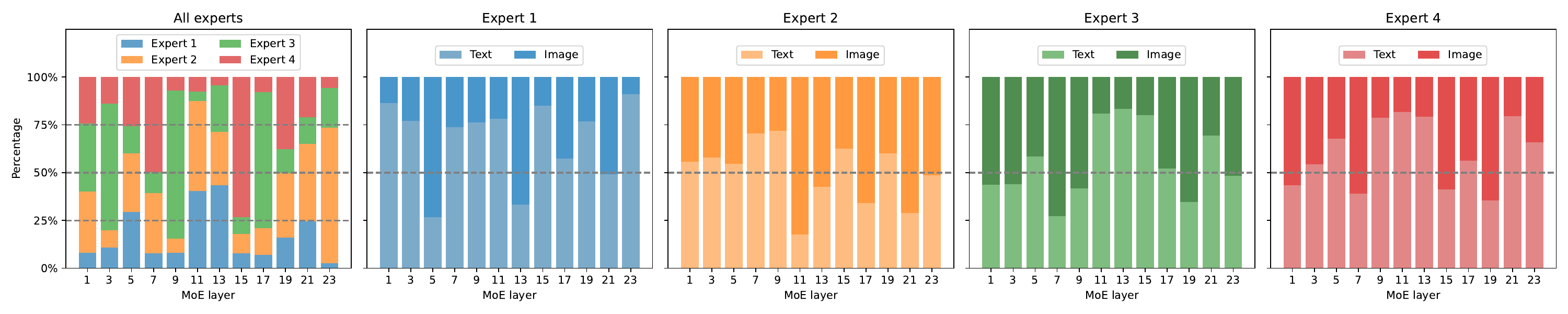}}\\
    \vskip -0.1in
	\subfloat[POPE]{\includegraphics[width=.49\columnwidth]{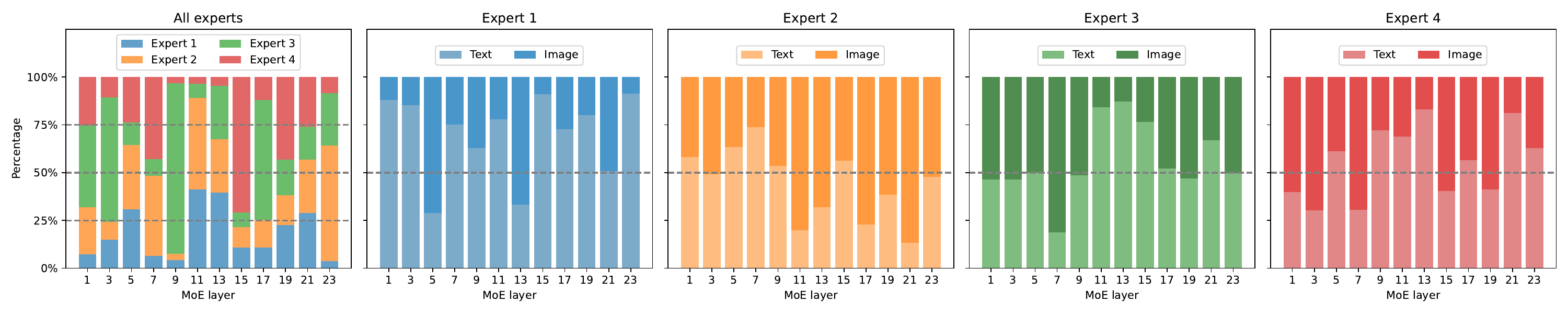}}\hspace{5pt}
	\subfloat[MMBench]{\includegraphics[width=.49\columnwidth]{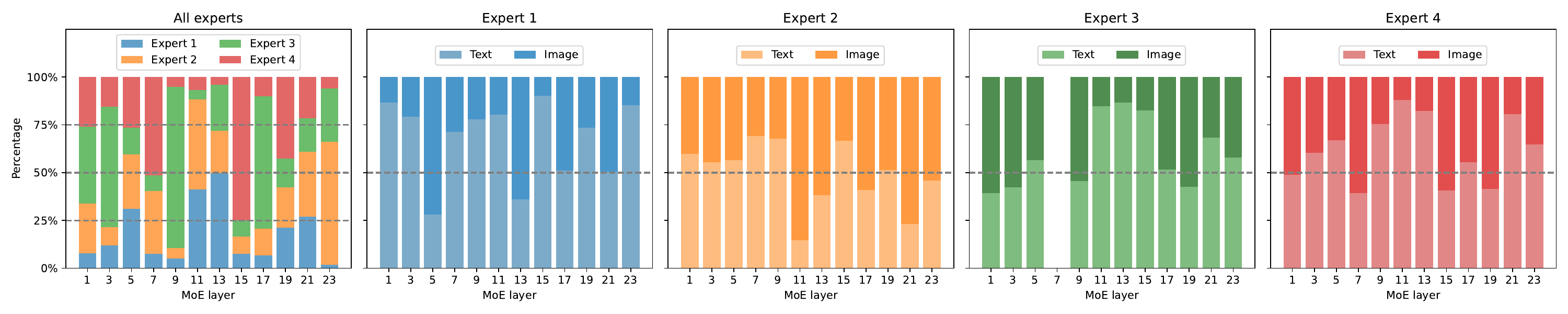}}\\
    \vskip -0.1in
	\subfloat[Viswiz]{\includegraphics[width=.49\columnwidth]{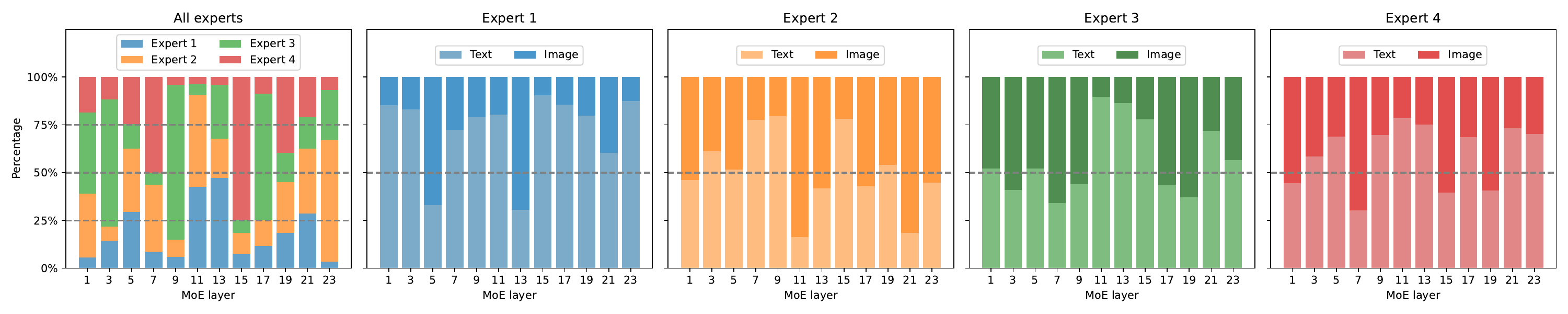}}\hspace{5pt}
	\subfloat[MM-Vet]{\includegraphics[width=.49\columnwidth]{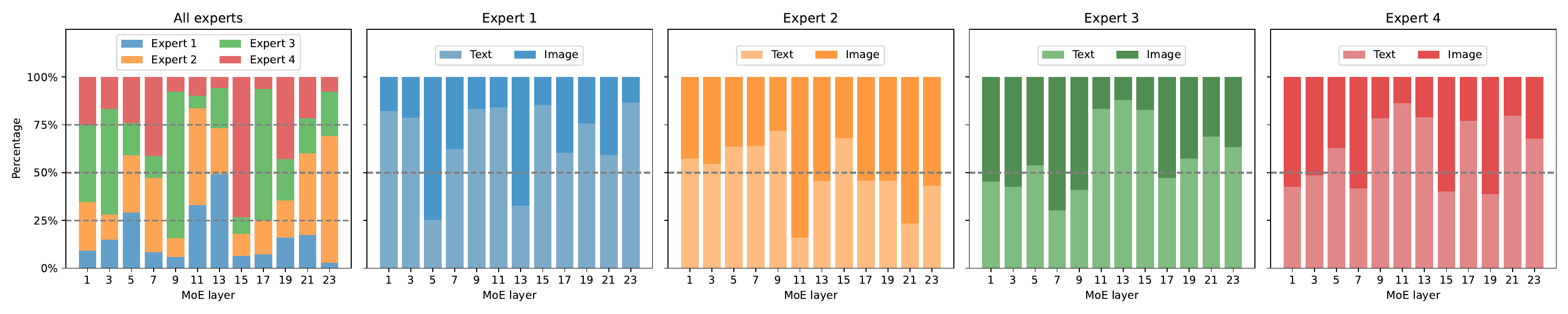}}
\vskip -0.1in
	\caption{Distribution of expert loadings and expert preferences on \textbf{MoE-LLaVA-Qwen-1.8B×4-Top2}.}
        \label{fig:expert_text_img_dist_qwen}
\vskip -0.2in
\end{figure}

\begin{figure}[h]
	\centering
	\subfloat[ScienceQA-IMG]{\includegraphics[width=.49\columnwidth]{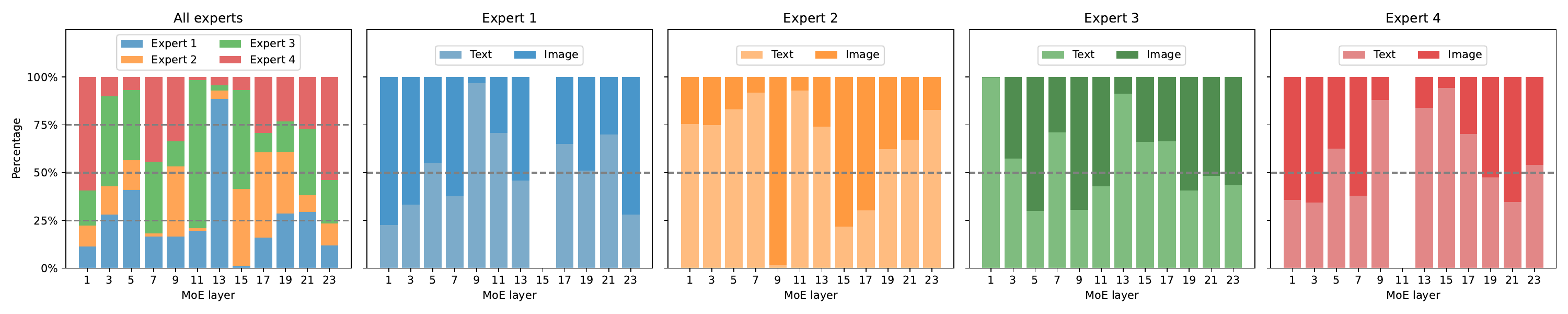}}\hspace{5pt}
	\subfloat[TextQA]{\includegraphics[width=.49\columnwidth]{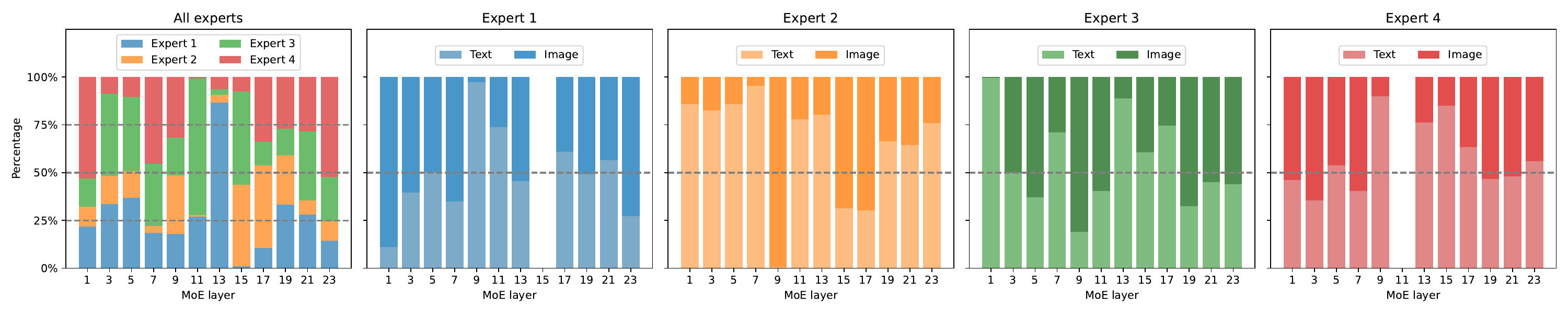}}\\
    \vskip -0.1in
	\subfloat[POPE]{\includegraphics[width=.49\columnwidth]{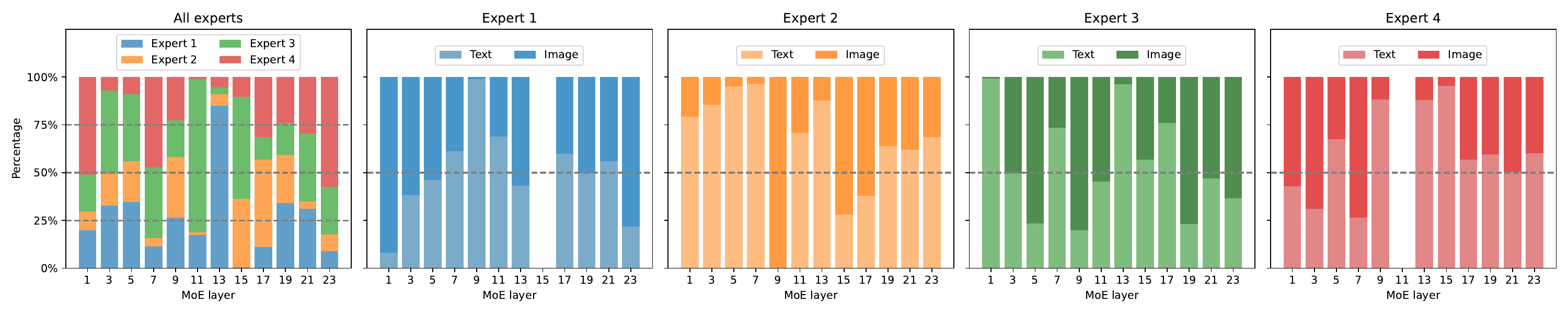}}\hspace{5pt}
	\subfloat[MMBench]{\includegraphics[width=.49\columnwidth]{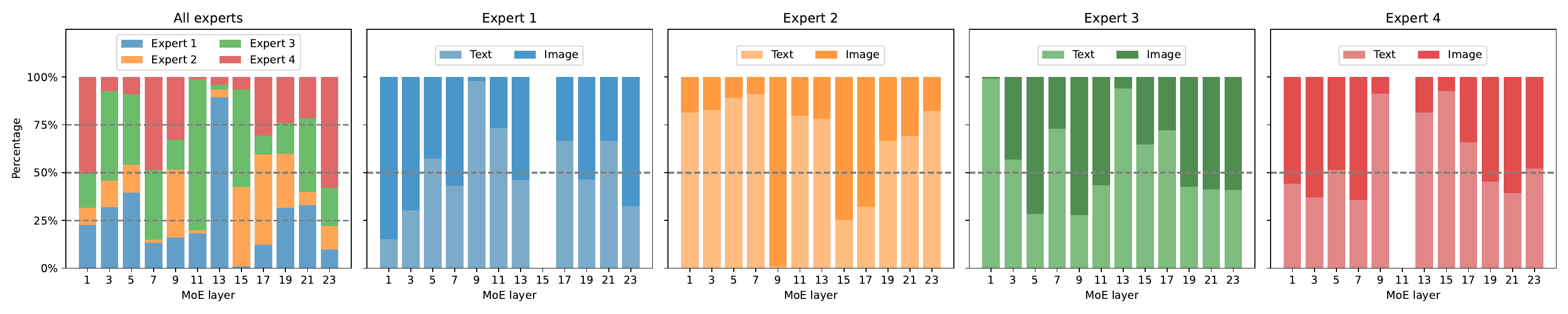}}\\
    \vskip -0.1in
	\subfloat[Viswiz]{\includegraphics[width=.49\columnwidth]{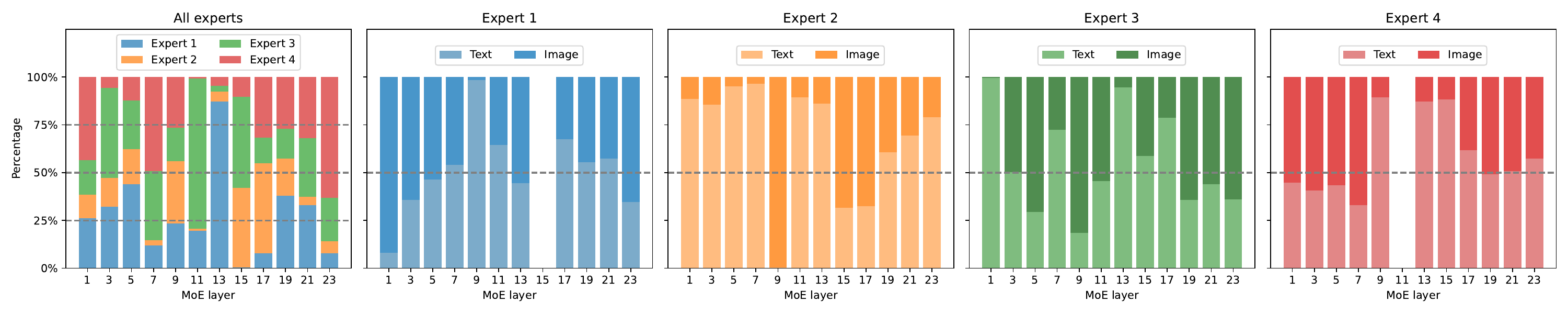}}\hspace{5pt}
	\subfloat[MM-Vet]{\includegraphics[width=.49\columnwidth]{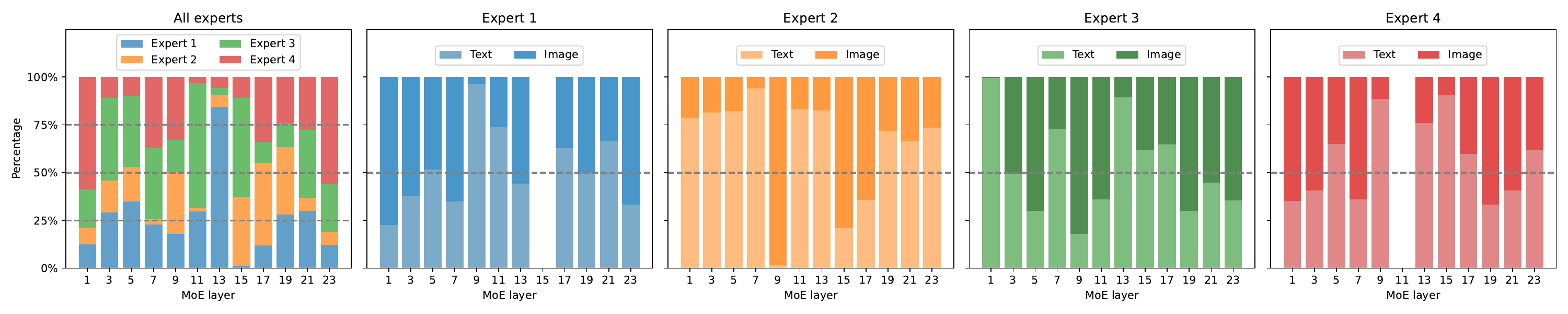}}
\vskip -0.1in
	\caption{Distribution of expert loadings and expert preferences on \textbf{MoE-LLaVA-StableLM-1.6B×4-Top2}.}
        \label{fig:expert_text_img_dist_stablelm}
\vskip -0.2in
\end{figure}

For MoE-LLaVA-OpenChat-7B×4-Top2, it is a truly large model compared to our setting. However, as shown in~\cref{appendix:model_scale}, its performance is not as good as expected. We provide the routing distribution of MoE-LLaVA-OpenChat after sparsification in~\cref{fig:expert_text_img_dist_openchat}. We can observe that even after three stages of training, the routing distributions of MoE-LLaVA-OpenChat and MoE-LLaVA-Phi (~\cref{fig:expert_text_img_dist_phi}) differ significantly. MoE-LLaVA-OpenChat exhibits a relatively balanced distribution overall, in terms of both expert loads and expert preferences for different modalities. On the other hand, MoE-LLaVA-Phi, along with other smaller models such as MoE-LLaVA-Qwen and MoE-LLaVA-StableLM, show some specific patterns or, in other words, their distributions are more \textbf{disordered}. For example, (1) in~\cref{fig:expert_text_img_dist_phi}, MoE-LLaVA-Phi exhibits a prominent expert 3 in layers 17-23, which dominates the majority of the workload. (2) In~\cref{fig:expert_text_img_dist_qwen}, MoE-LLaVA-Qwen shows a strong preference for the image modality in expert 1. (3) In Figure~\cref{fig:expert_text_img_dist_stablelm}, experts 2 and 3 of MoE-LLaVA-StableLM are actively engaged in the middle layers of the model. We believe this is highly likely due to the insufficient amount of current multimodal fine-tuning data (655k in our setting) to enable sparsification for 10B-level models, even starting from a well-initialized LVLM.

In fact, we should reflect on what behavior is expected for a sparse MoE model. \textbf{Should it exhibit specific patterns for each expert, like MoE-LLaVA-Phi, or should it have similar behavior among the experts, like MoE-LLaVA-OpenChat?} If we consider that in a sparse model, the behavior of each expert should be similar at initialization, as they are initialized from a shared FFN and the router has not yet learned any inductive biases, then if the routing distribution continues to remain balanced as the network learns, it would be similar to the initialization and may lead to confusion in the model. \textbf{Therefore, we speculate that the lack of sufficient data may be a reason for the poor performance of MoE-LLaVA-OpenChat.} However, due to the current limitations in data and computational resources, we are unable to further explore this, and we hope that future work can make progress in this direction.

\begin{figure}[h]
	\centering
	\subfloat[ScienceQA-IMG]{\includegraphics[width=.32\columnwidth]{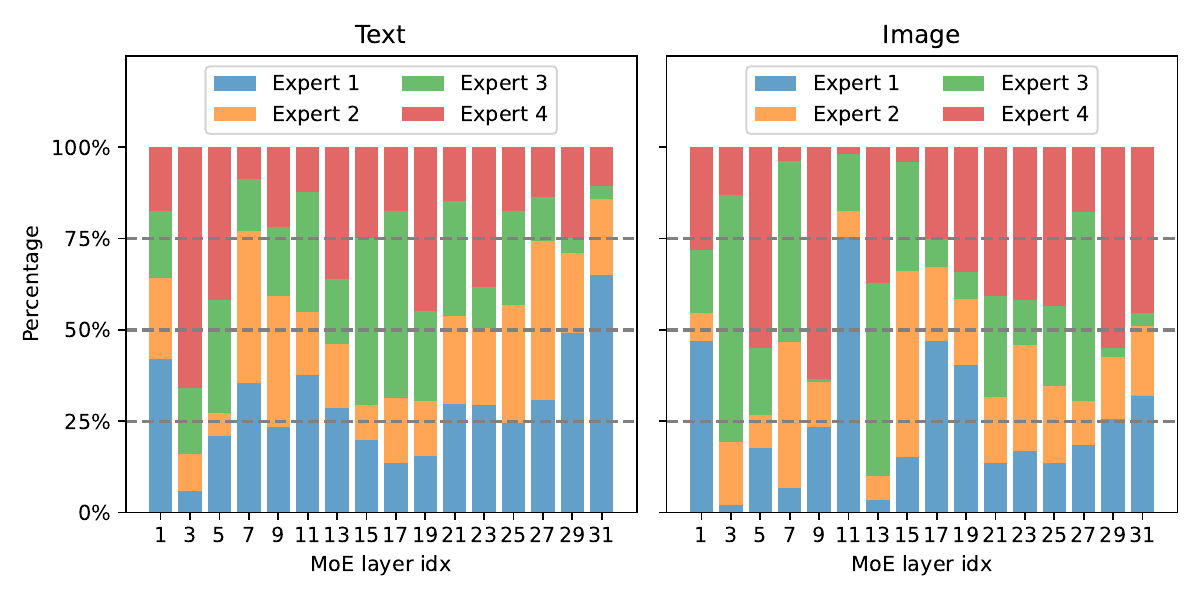}}\hspace{5pt}
	\subfloat[TextQA]{\includegraphics[width=.32\columnwidth]{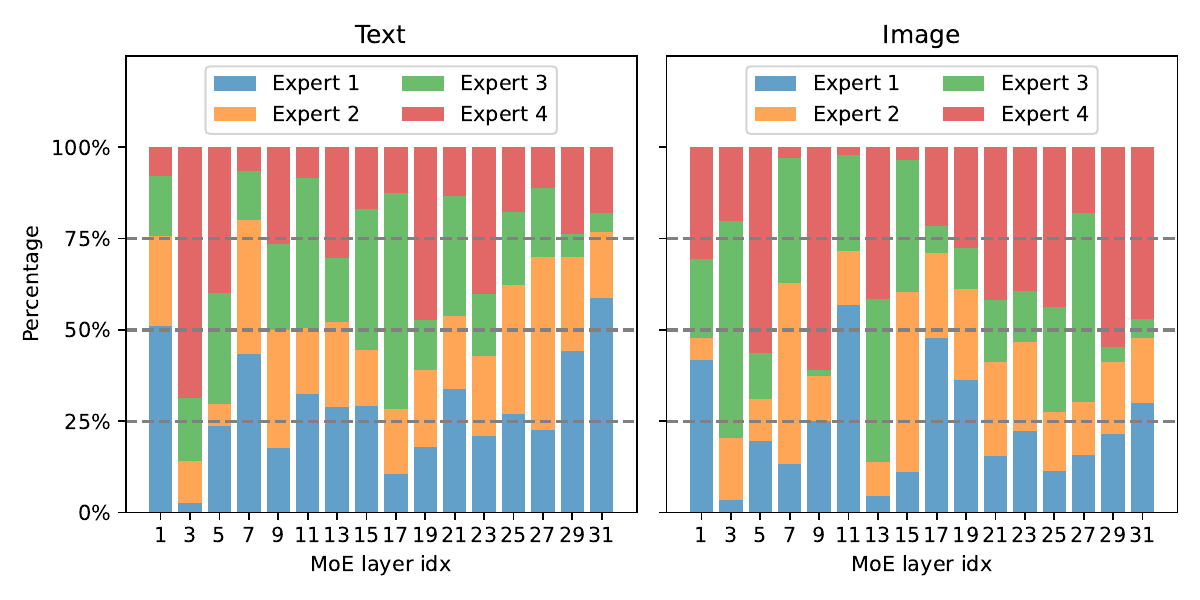}}\hspace{5pt}
	\subfloat[POPE]{\includegraphics[width=.32\columnwidth]{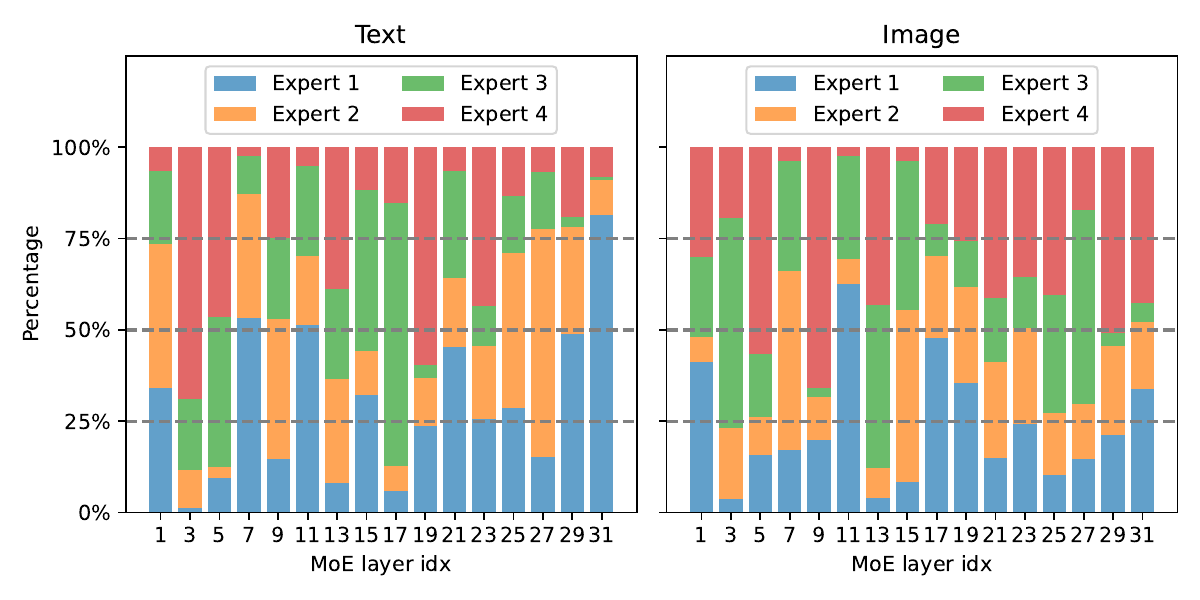}}\\
    \vskip -0.1in
	\subfloat[MMBench]{\includegraphics[width=.32\columnwidth]{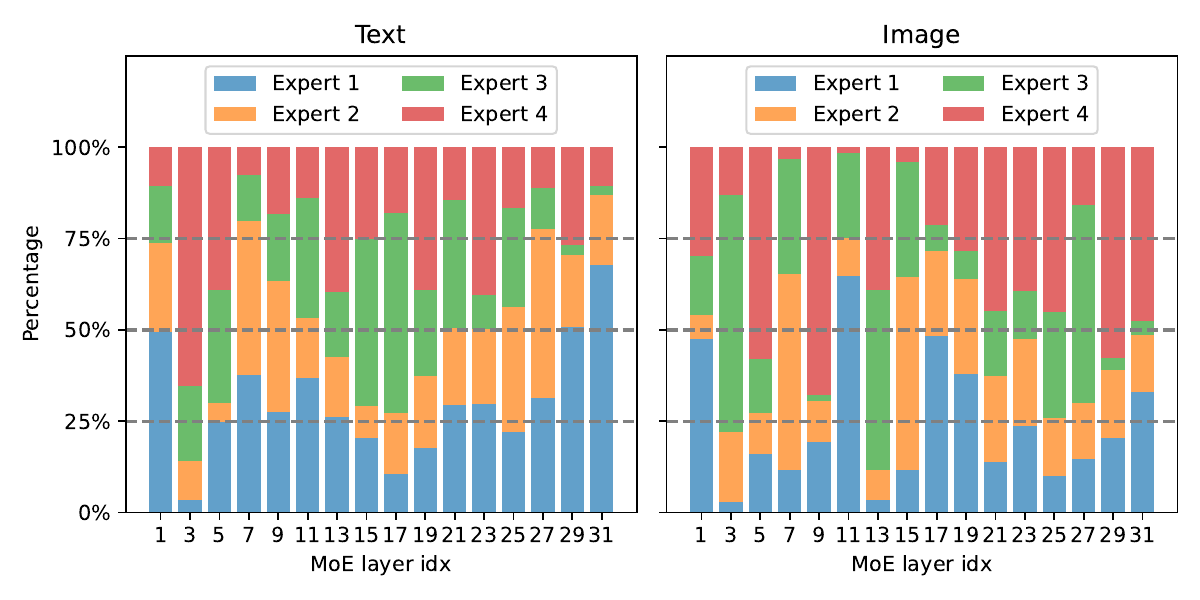}}\hspace{5pt}
	\subfloat[Viswiz]{\includegraphics[width=.32\columnwidth]{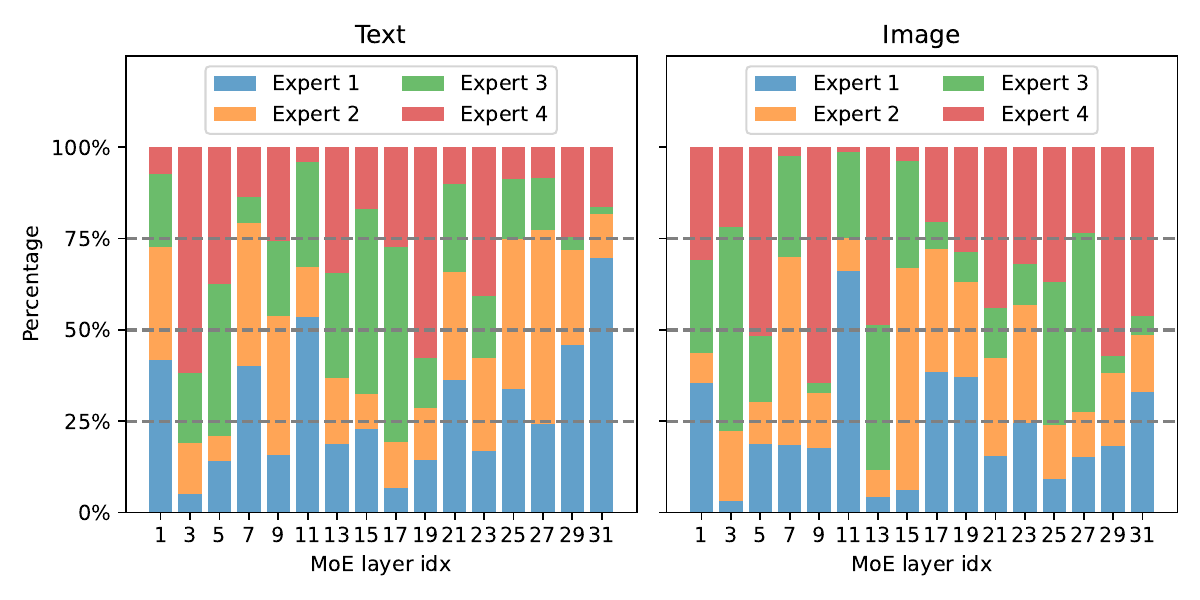}}\hspace{5pt}
	\subfloat[MM-Vet]{\includegraphics[width=.32\columnwidth]{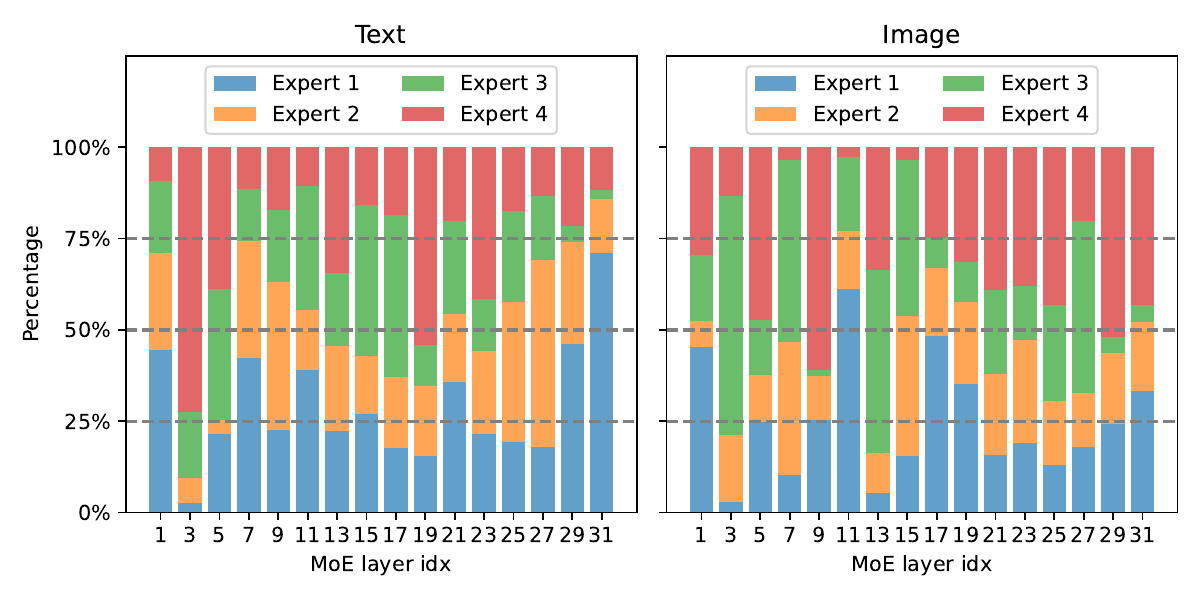}}
\vskip -0.1in
	\caption{Distribution of modalities across different experts on \textbf{MoE-LLaVA-OpenChat-7B×4-Top2}.}
        \label{fig:text_img_assign_openchat}
\vskip -0.2in
\end{figure}

\begin{figure}[h]
	\centering
	\subfloat[ScienceQA-IMG]{\includegraphics[width=.32\columnwidth]{fig/text_img_assign/phi_sciqa.pdf}}\hspace{5pt}
	\subfloat[TextQA]{\includegraphics[width=.32\columnwidth]{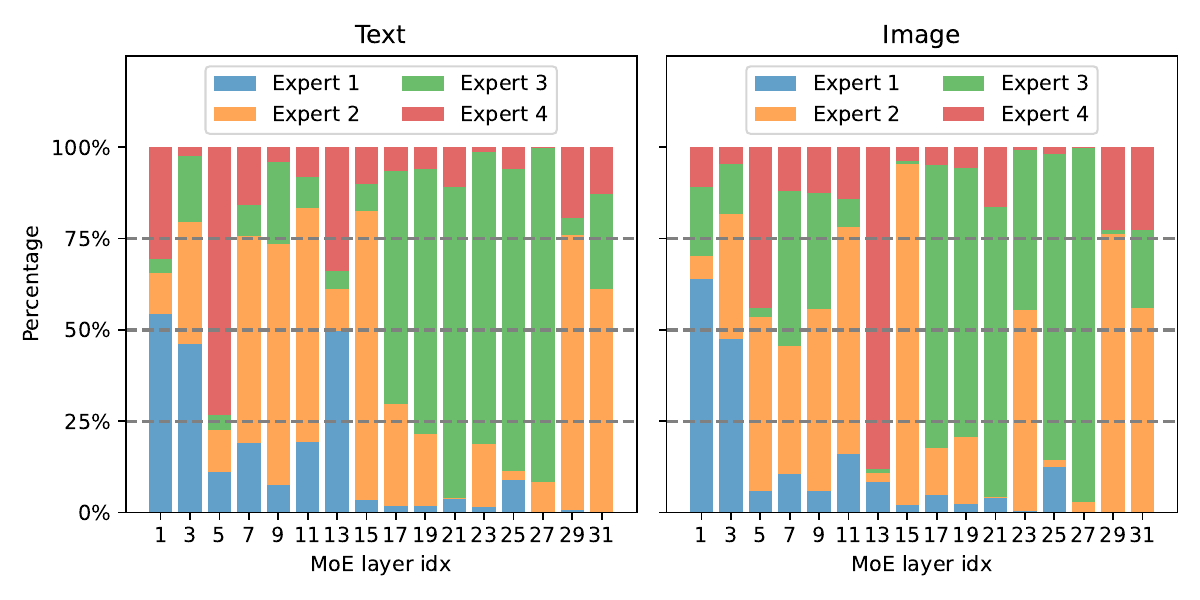}}\hspace{5pt}
	\subfloat[POPE]{\includegraphics[width=.32\columnwidth]{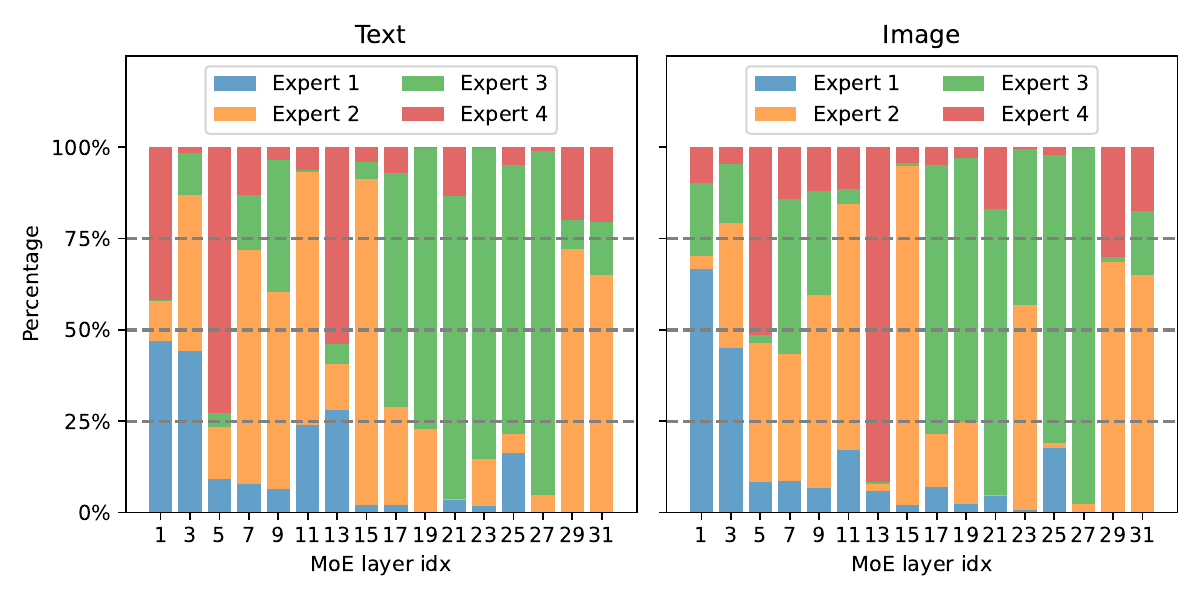}}\\
    \vskip -0.1in
	\subfloat[MMBench]{\includegraphics[width=.32\columnwidth]{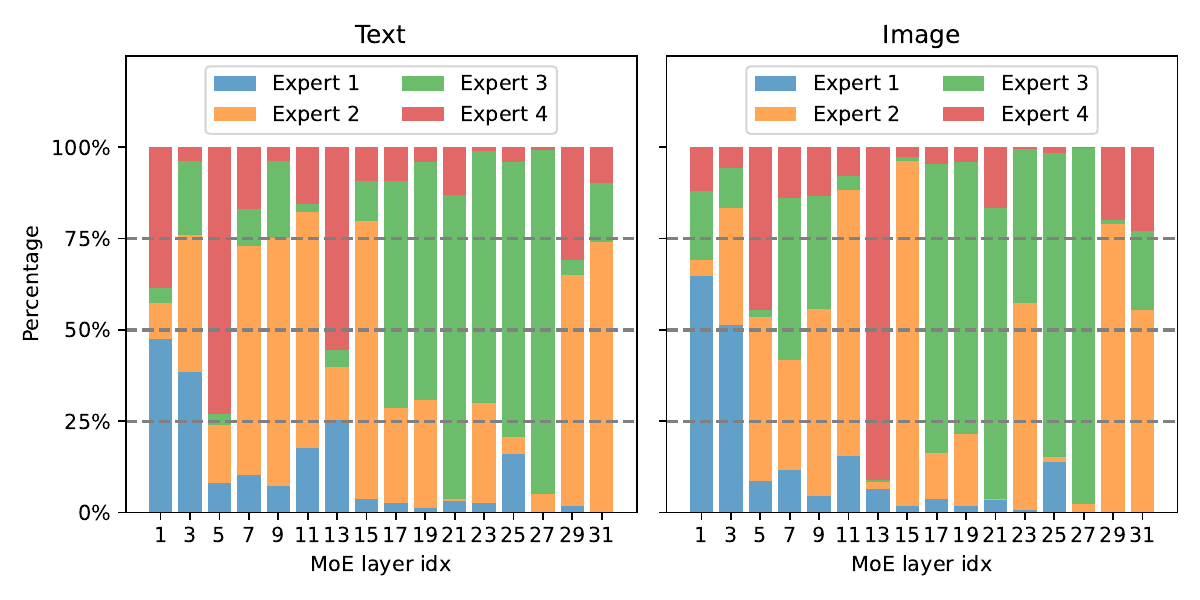}}\hspace{5pt}
	\subfloat[Viswiz]{\includegraphics[width=.32\columnwidth]{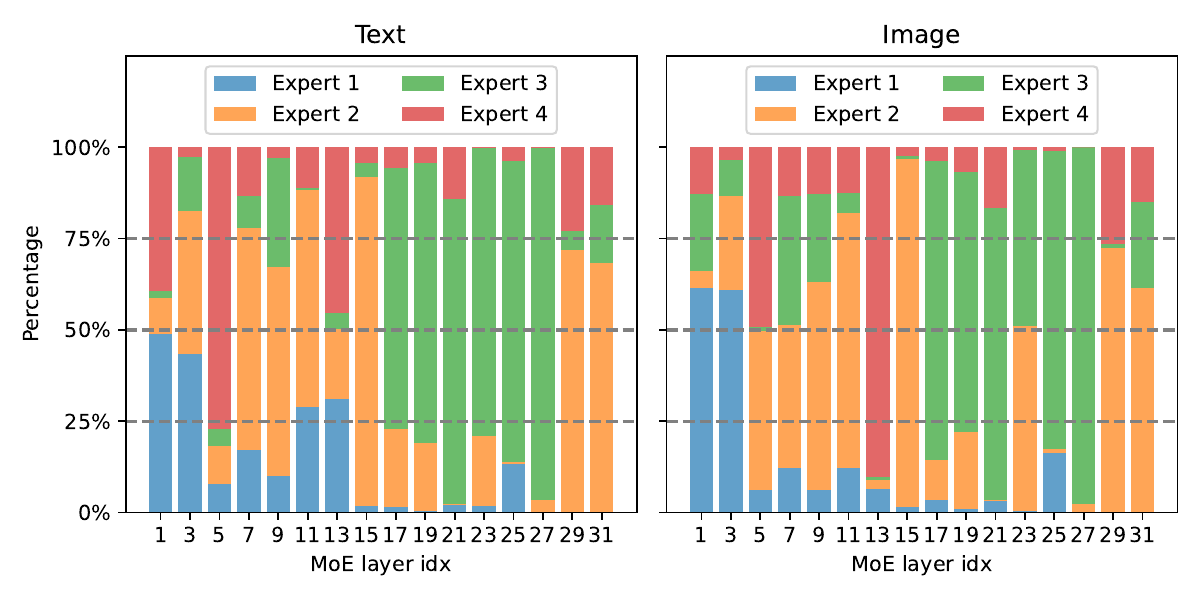}}\hspace{5pt}
	\subfloat[MM-Vet]{\includegraphics[width=.32\columnwidth]{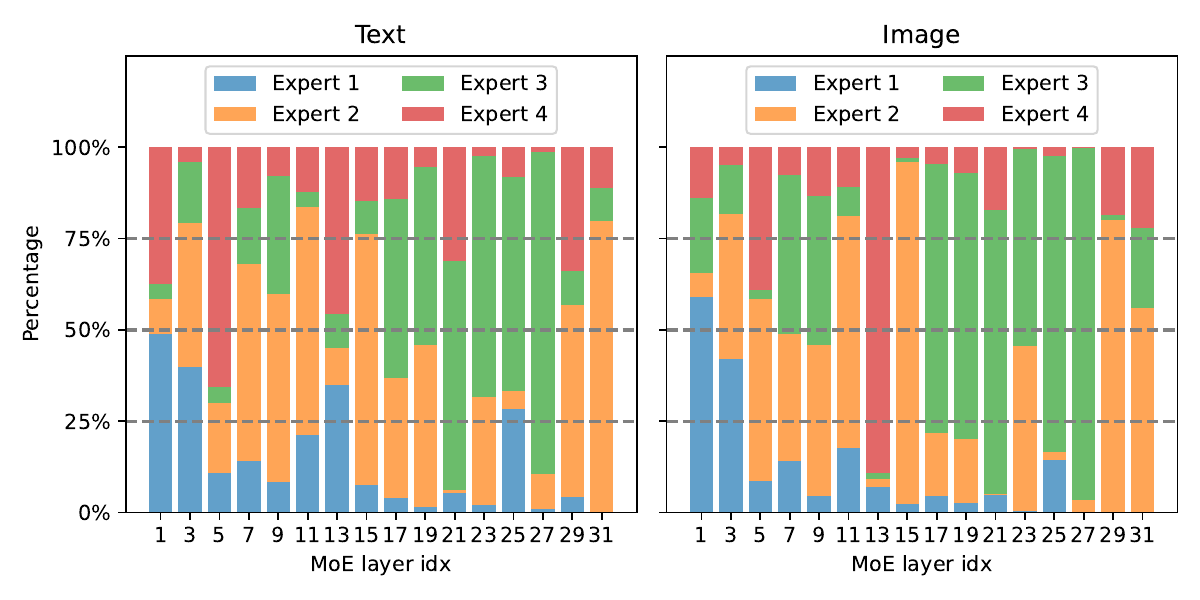}}
\vskip -0.1in
	\caption{Distribution of modalities across different experts on \textbf{MoE-LLaVA-Phi-2.7B×4-Top2}.}
        \label{fig:text_img_assign_phi}
\vskip -0.2in
\end{figure}

\begin{figure}[h]
	\centering
	\subfloat[ScienceQA-IMG]{\includegraphics[width=.32\columnwidth]{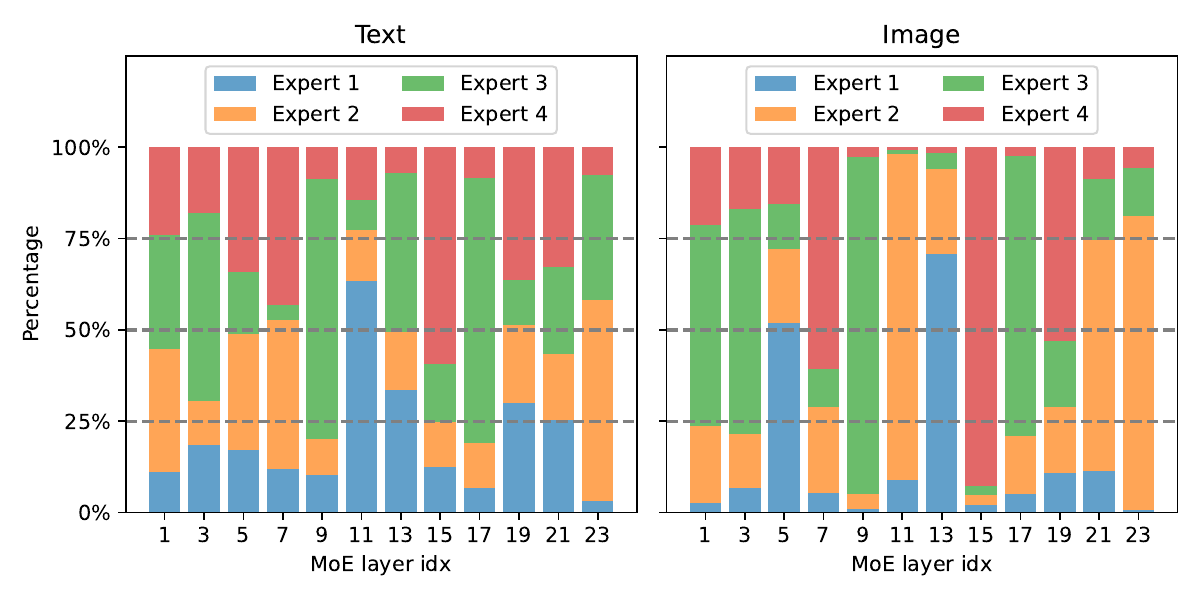}}\hspace{5pt}
	\subfloat[TextQA]{\includegraphics[width=.32\columnwidth]{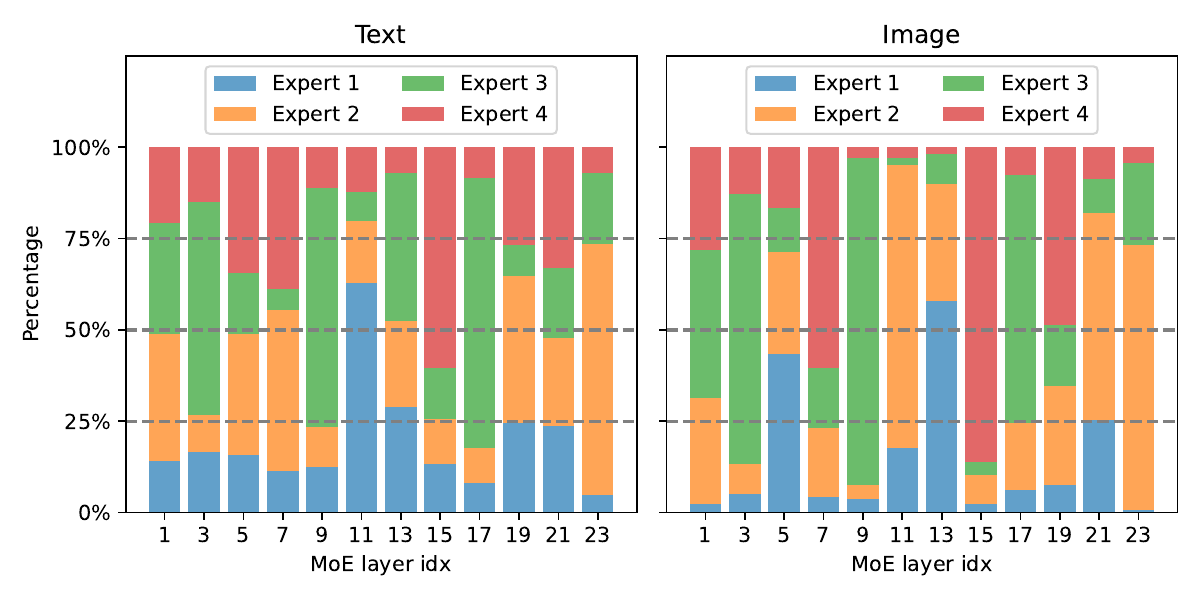}}\hspace{5pt}
	\subfloat[POPE]{\includegraphics[width=.32\columnwidth]{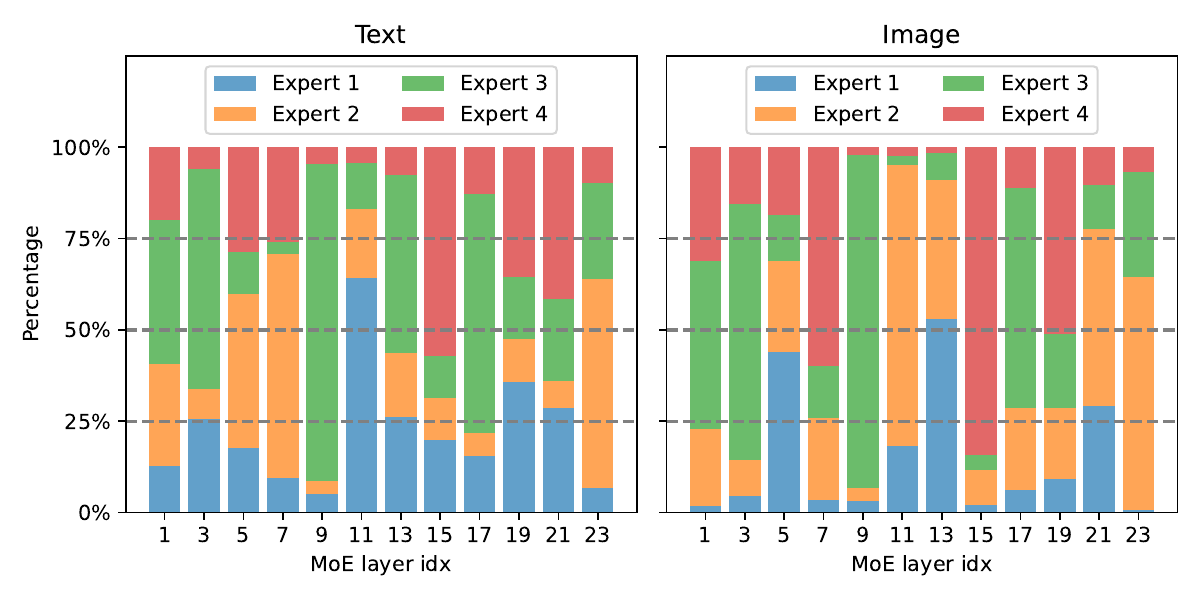}}\\
    \vskip -0.1in
	\subfloat[MMBench]{\includegraphics[width=.32\columnwidth]{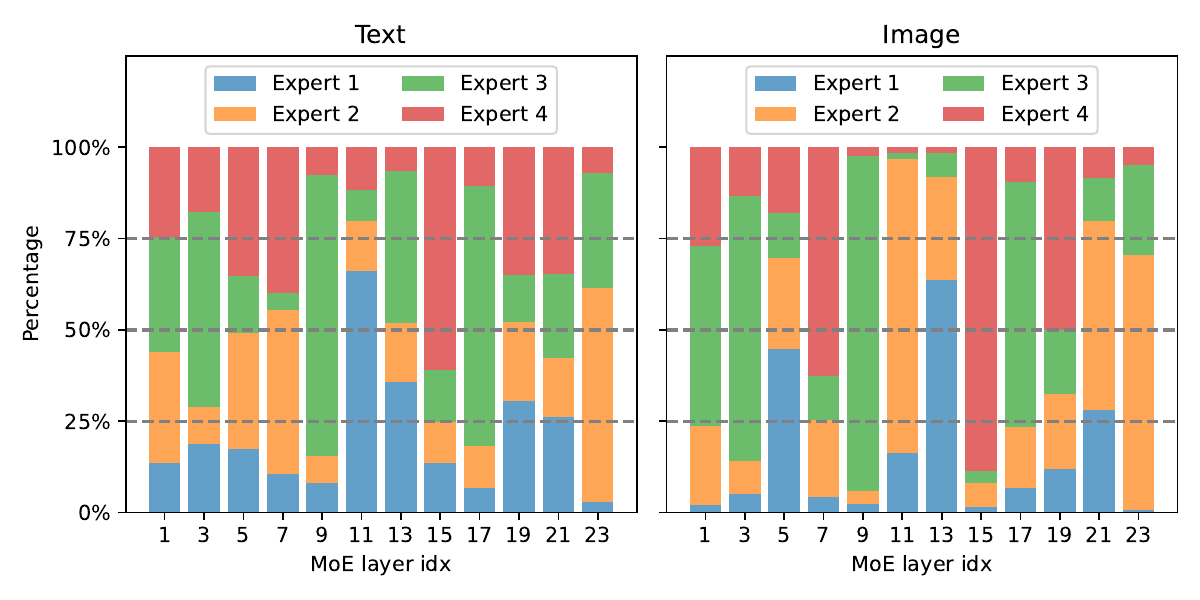}}\hspace{5pt}
	\subfloat[Viswiz]{\includegraphics[width=.32\columnwidth]{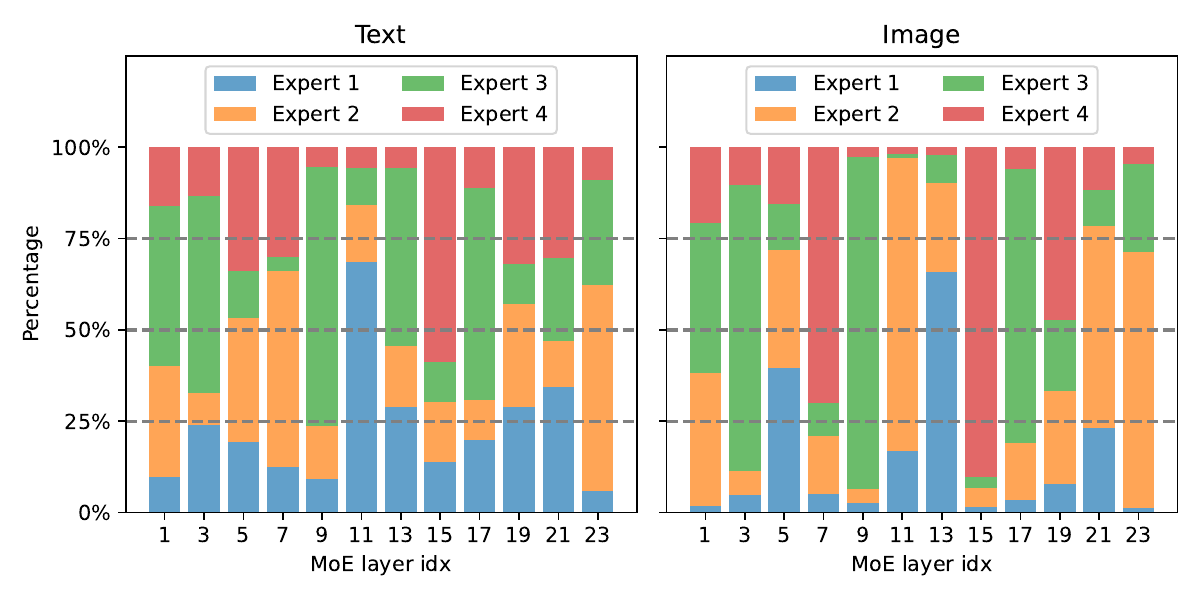}}\hspace{5pt}
	\subfloat[MM-Vet]{\includegraphics[width=.32\columnwidth]{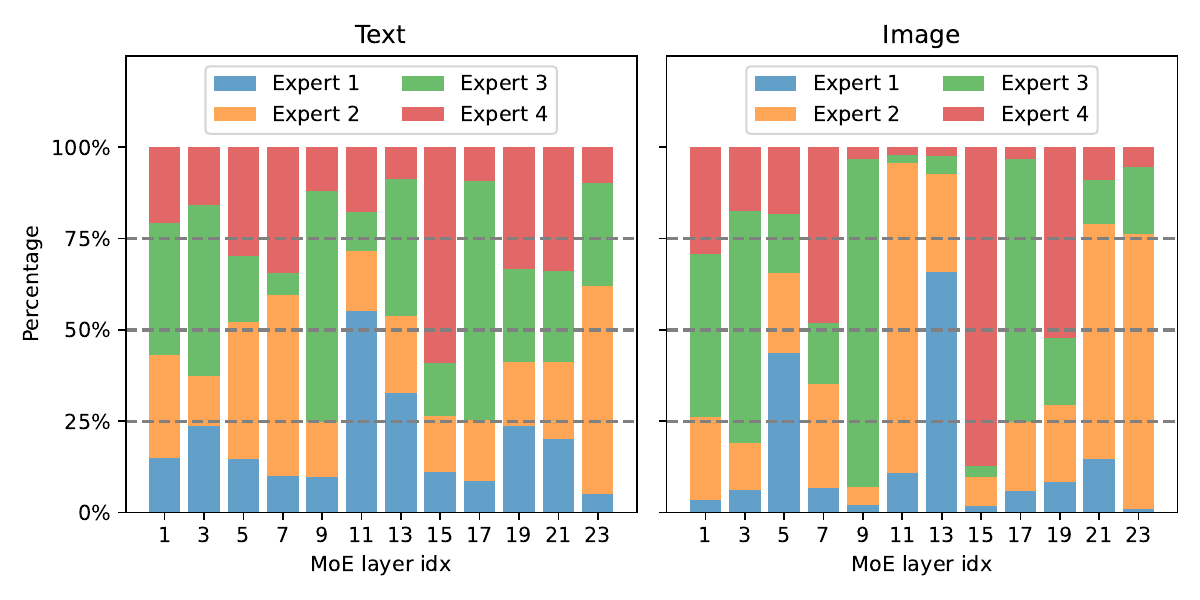}}
\vskip -0.1in
	\caption{Distribution of modalities across different experts on \textbf{MoE-LLaVA-Qwen-1.8B×4-Top2}.}
        \label{fig:text_img_assign_qwen}
\vskip -0.2in
\end{figure}

\begin{figure}[h]
	\centering
	\subfloat[ScienceQA-IMG]{\includegraphics[width=.32\columnwidth]{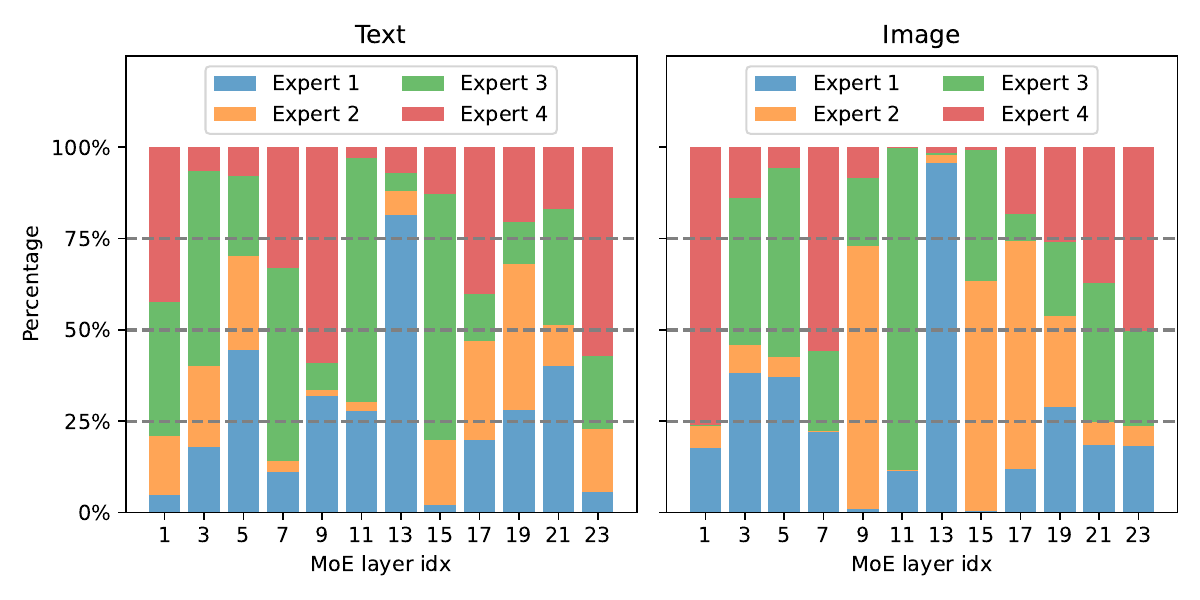}}\hspace{5pt}
	\subfloat[TextQA]{\includegraphics[width=.32\columnwidth]{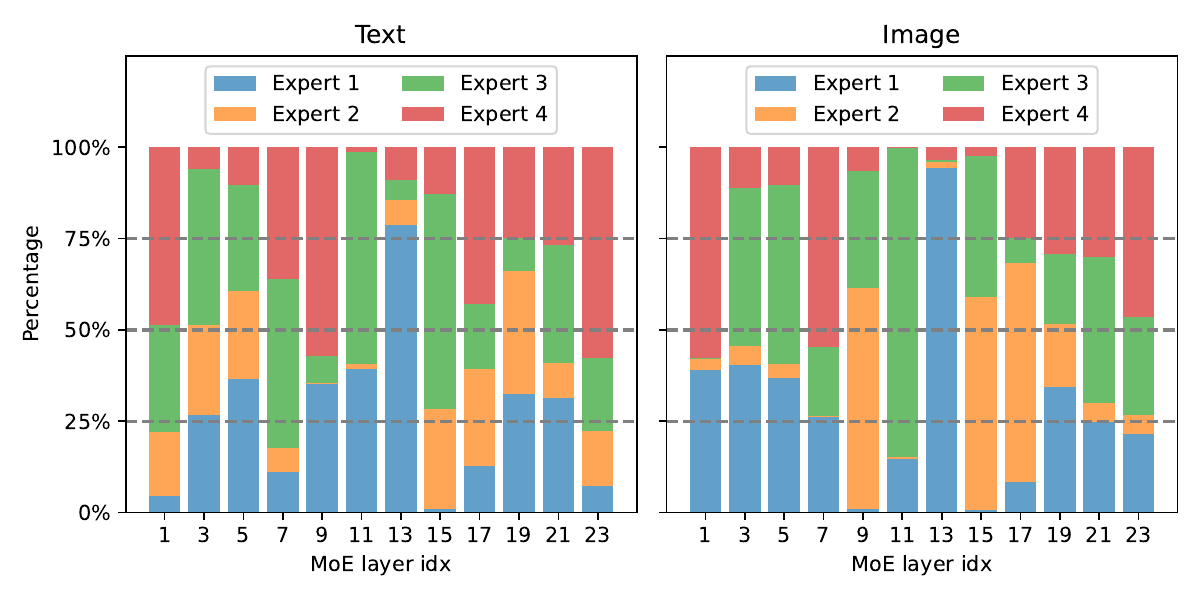}}\hspace{5pt}
	\subfloat[POPE]{\includegraphics[width=.32\columnwidth]{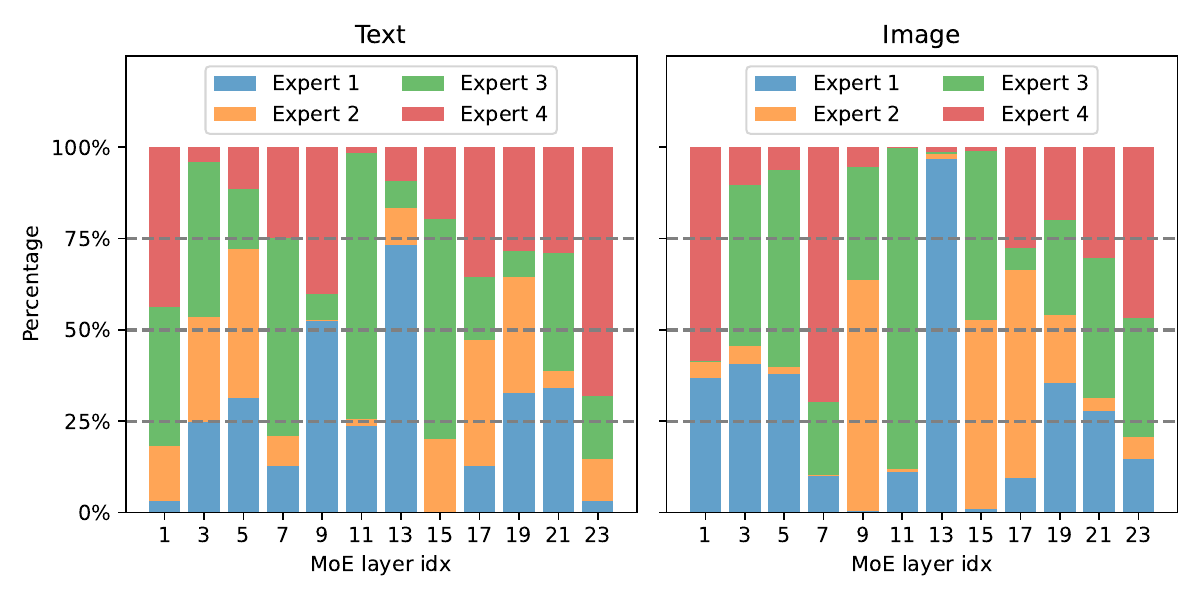}}\\
    \vskip -0.1in
	\subfloat[MMBench]{\includegraphics[width=.32\columnwidth]{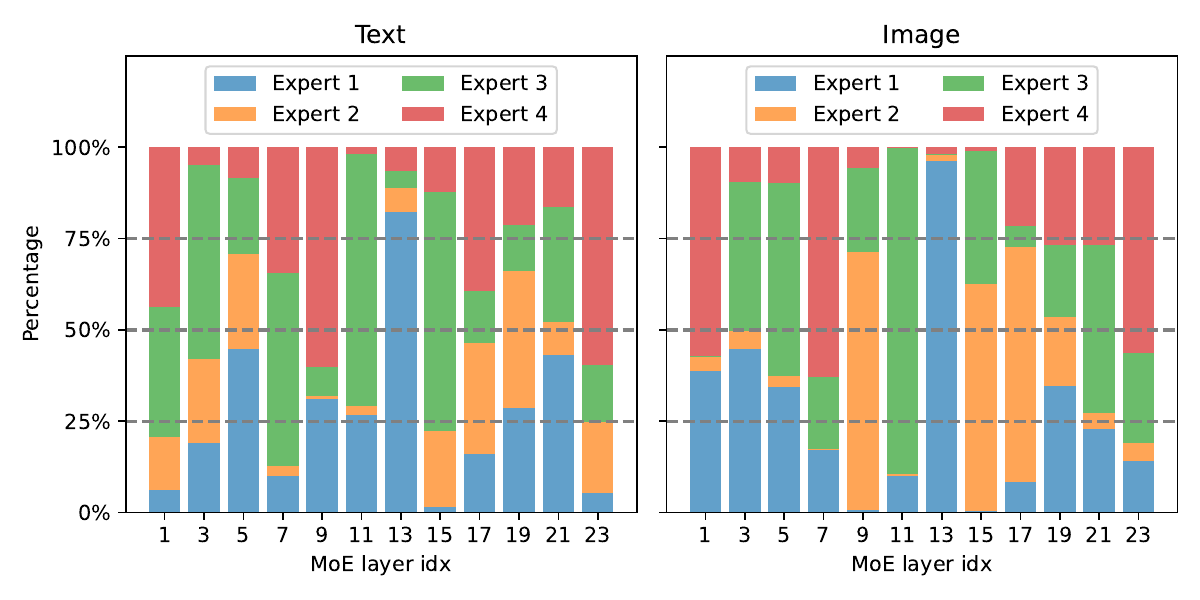}}\hspace{5pt}
	\subfloat[Viswiz]{\includegraphics[width=.32\columnwidth]{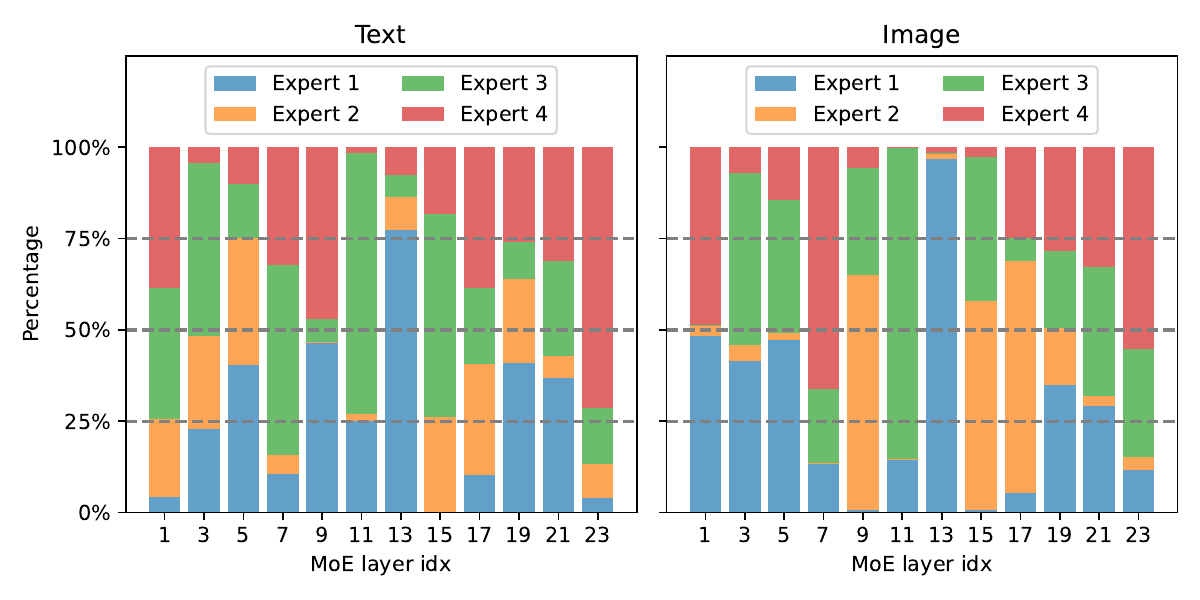}}\hspace{5pt}
	\subfloat[MM-Vet]{\includegraphics[width=.32\columnwidth]{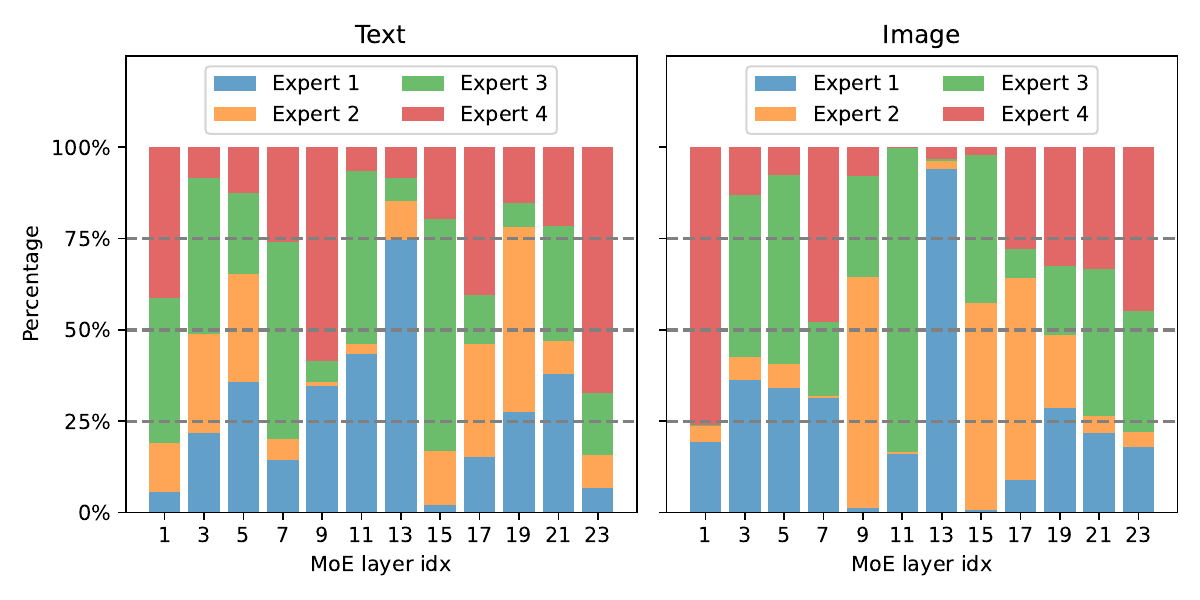}}
\vskip -0.1in
	\caption{Distribution of modalities across different experts on \textbf{MoE-LLaVA-StableLM-1.6B×4-Top2}.}
        \label{fig:text_img_assign_stablelm}
\vskip -0.2in
\end{figure}

Additionally, we provide more details in~\cref{fig:text_img_assign_openchat},~\cref{fig:text_img_assign_phi},~\cref{fig:text_img_assign_qwen}, and~\cref{fig:text_img_assign_stablelm}.

\vskip -0.2in

\subsection{Token Pathways}
\label{appendix:token_pathway}

In~\cref{fig:text_img_assign_openchat},~\cref{fig:text_img_assign_phi},~\cref{fig:text_img_assign_qwen}, and~\cref{fig:text_img_assign_stablelm}, we track the paths of each token for MoE-LLaVA-OpenChat-7B×4-Top2, MoE-LLaVA-Phi-2.7B×4-Top2, MoE-LLaVA-Qwen-1.8B×4-Top2, and MoE-LLaVA-StableLM-1.6B×4-Top2, respectively. In general, the overall trends of the token paths align with the analysis in~\cref{appendix:router_dist}. The paths of MoE-LLaVA-OpenChat-7B×4-Top2 appear more disorderly and diverse, which is attributed to a more balanced expert assignment. On the other hand, MoE-LLaVA-Phi-2.7B×4-Top2, MoE-LLaVA-Qwen-1.8B×4-Top2, and MoE-LLaVA-StableLM-1.6B×4-Top2 each exhibit their specific patterns.

\begin{figure}[htbp]
	\centering
	\subfloat[ScienceQA-IMG]{\includegraphics[width=.32\columnwidth]{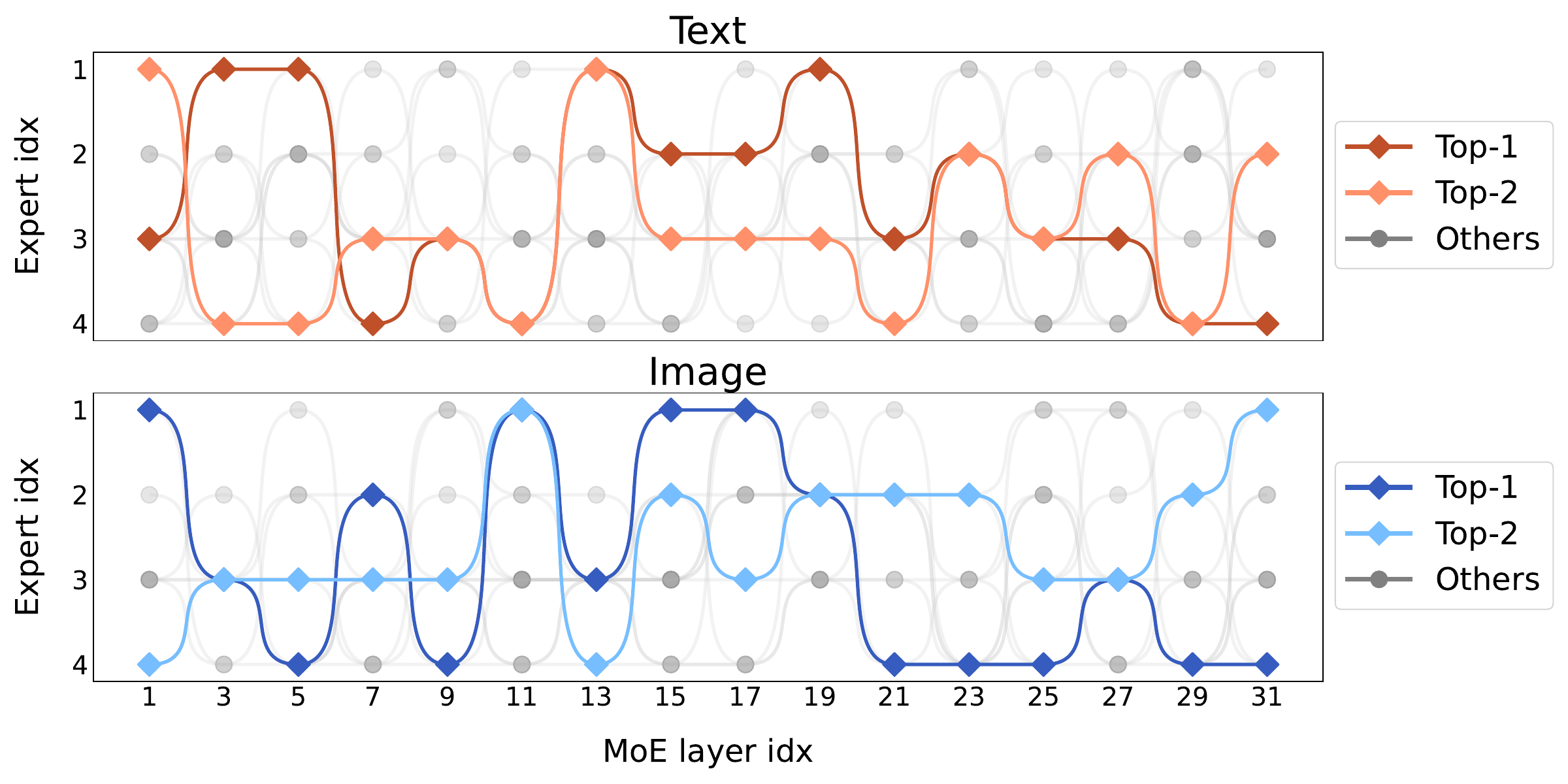}}\hspace{5pt}
	\subfloat[TextQA]{\includegraphics[width=.32\columnwidth]{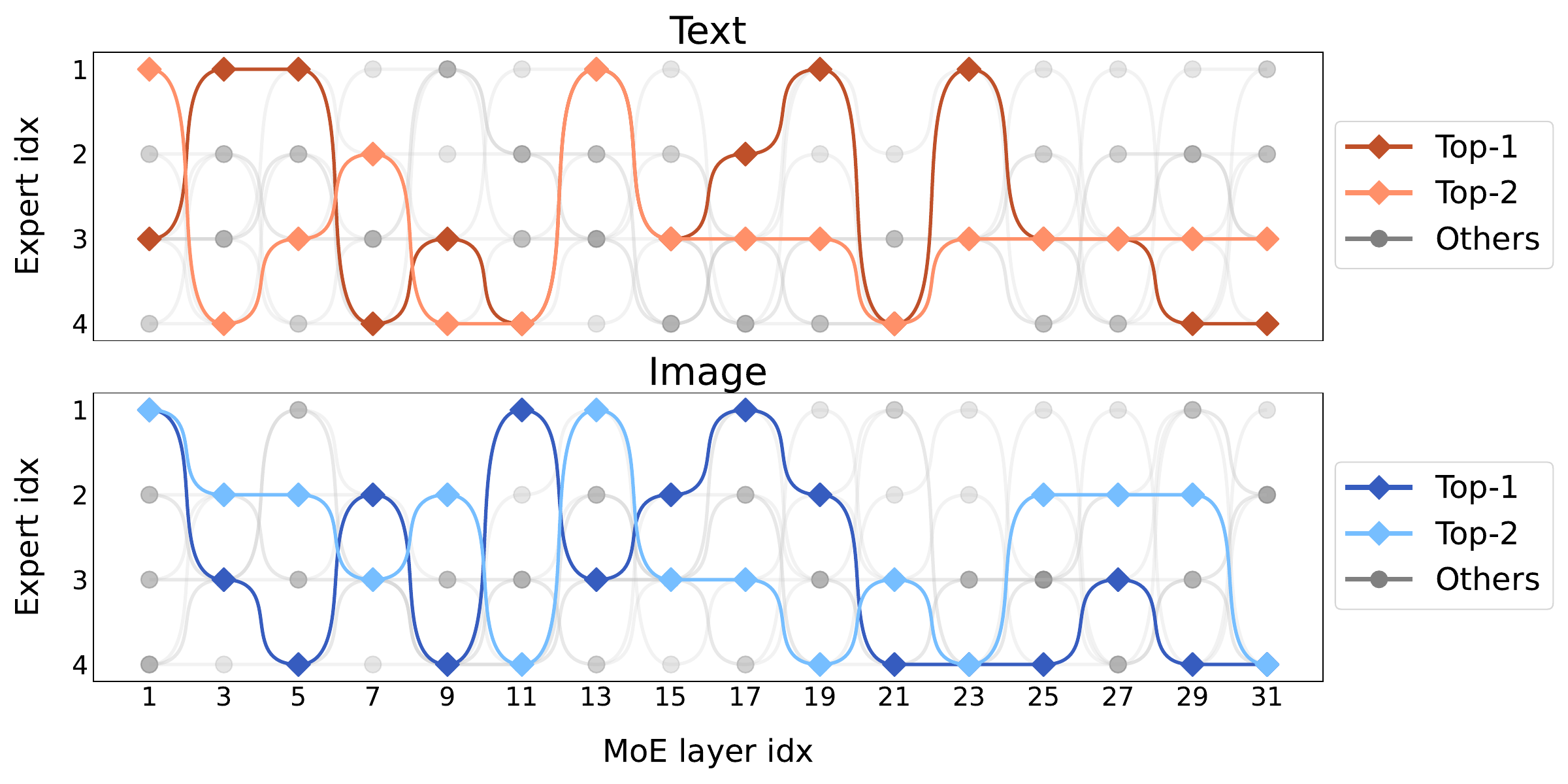}}\hspace{5pt}
	\subfloat[POPE]{\includegraphics[width=.32\columnwidth]{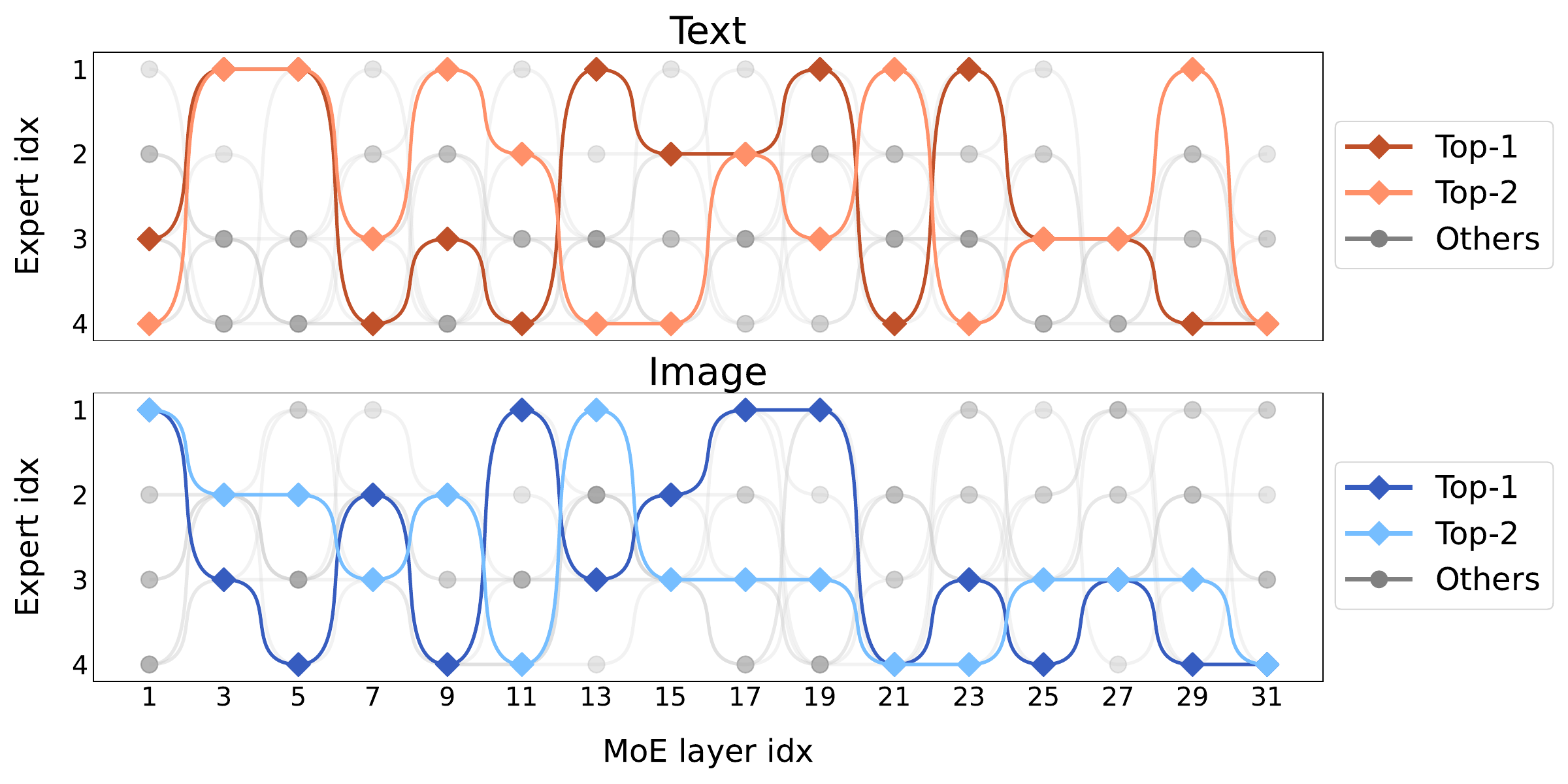}}\\
    \vskip -0.1in
	\subfloat[MMBench]{\includegraphics[width=.32\columnwidth]{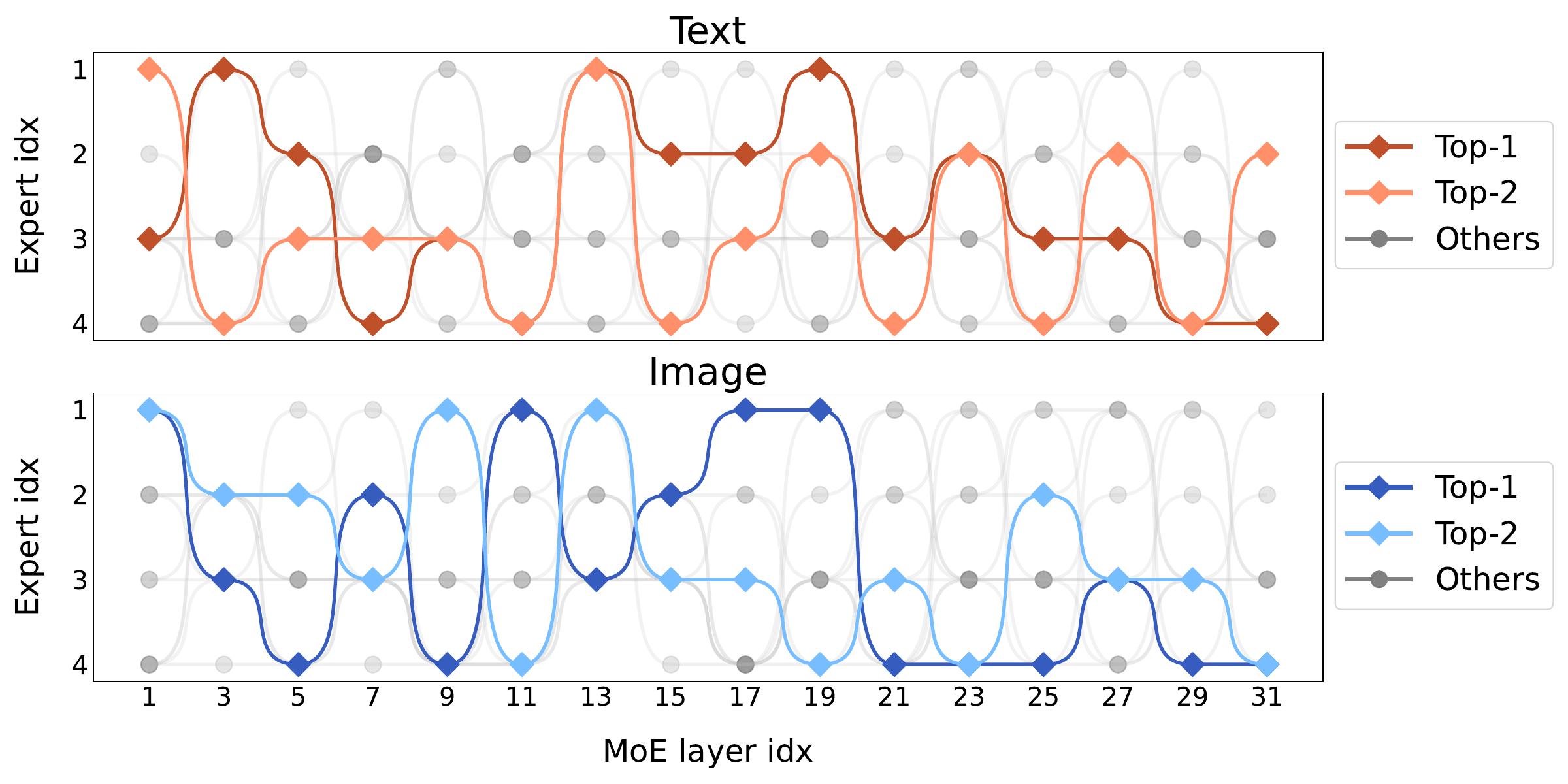}}\hspace{5pt}
	\subfloat[Viswiz]{\includegraphics[width=.32\columnwidth]{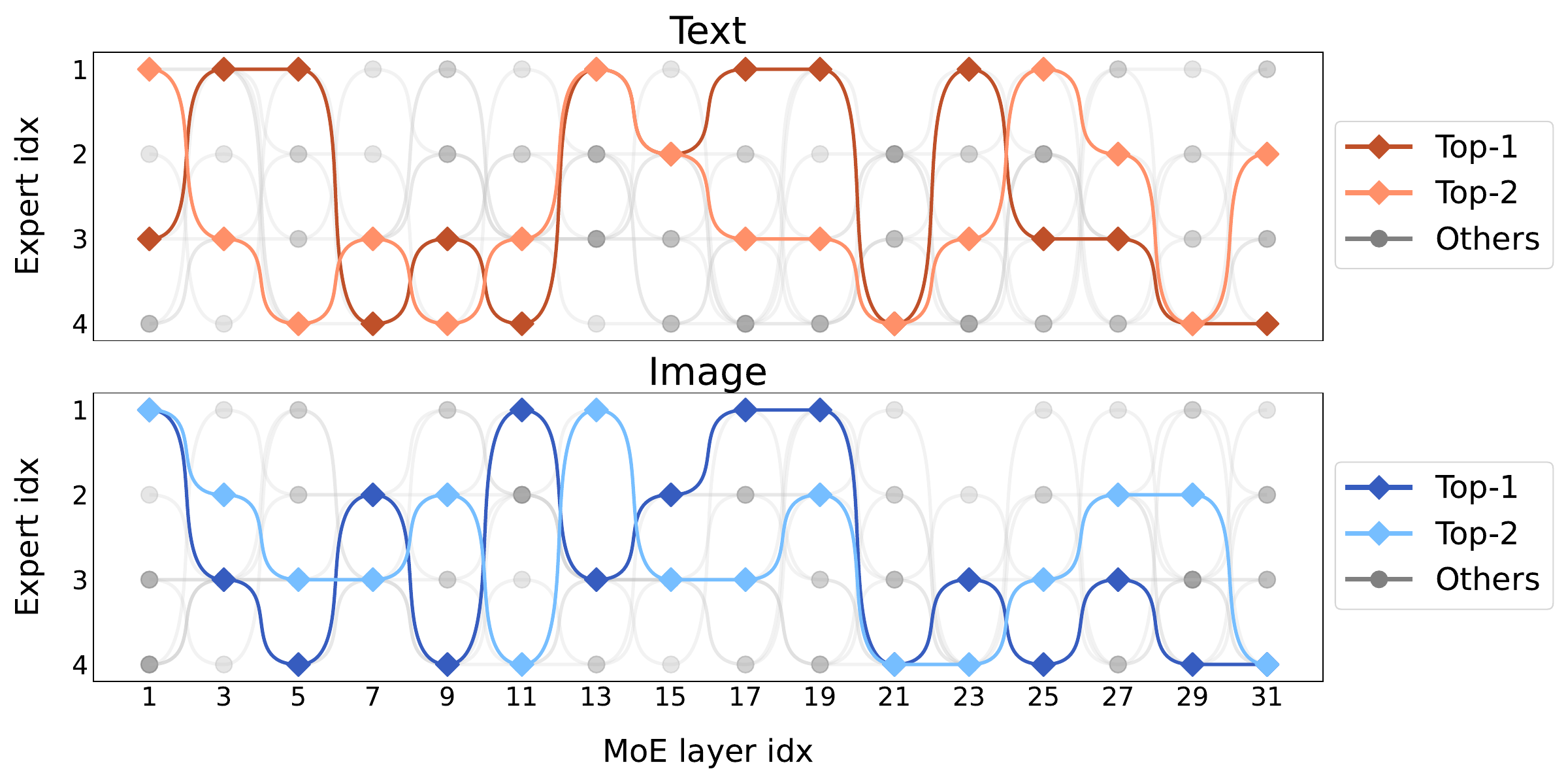}}\hspace{5pt}
	\subfloat[MM-Vet]{\includegraphics[width=.32\columnwidth]{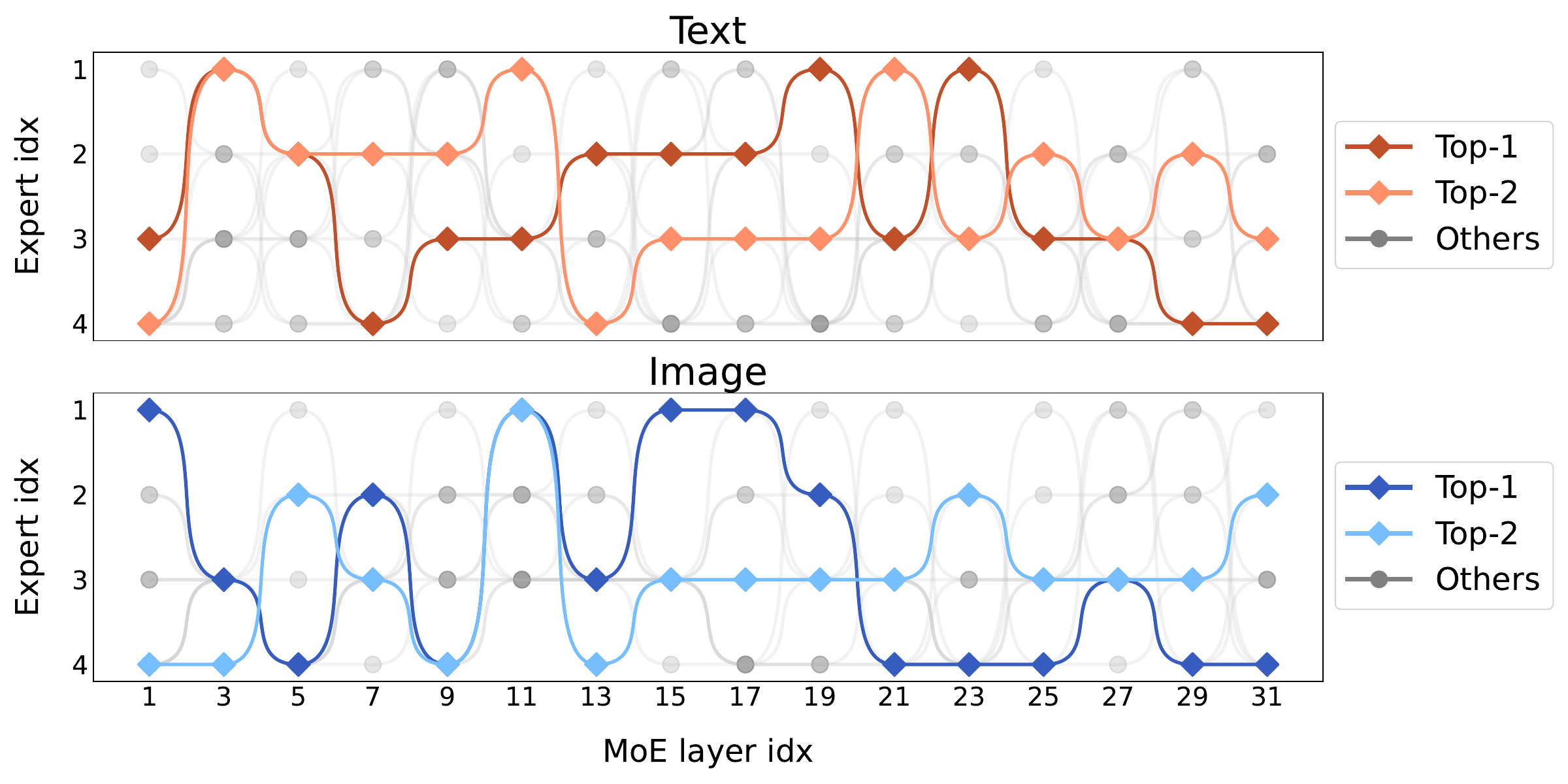}}
\vskip -0.1in
	\caption{Visualization of activated pathways on \textbf{MoE-LLaVA-OpenChat-7B×4-Top2}.}
        \label{fig:bumpy_text_img_phi}
\vskip -0.2in
\end{figure}

\begin{figure}[htbp]
	\centering
	\subfloat[ScienceQA-IMG]{\includegraphics[width=.32\columnwidth]{fig/bumpy_text_img/phi_sciqa.pdf}}\hspace{5pt}
	\subfloat[TextQA]{\includegraphics[width=.32\columnwidth]{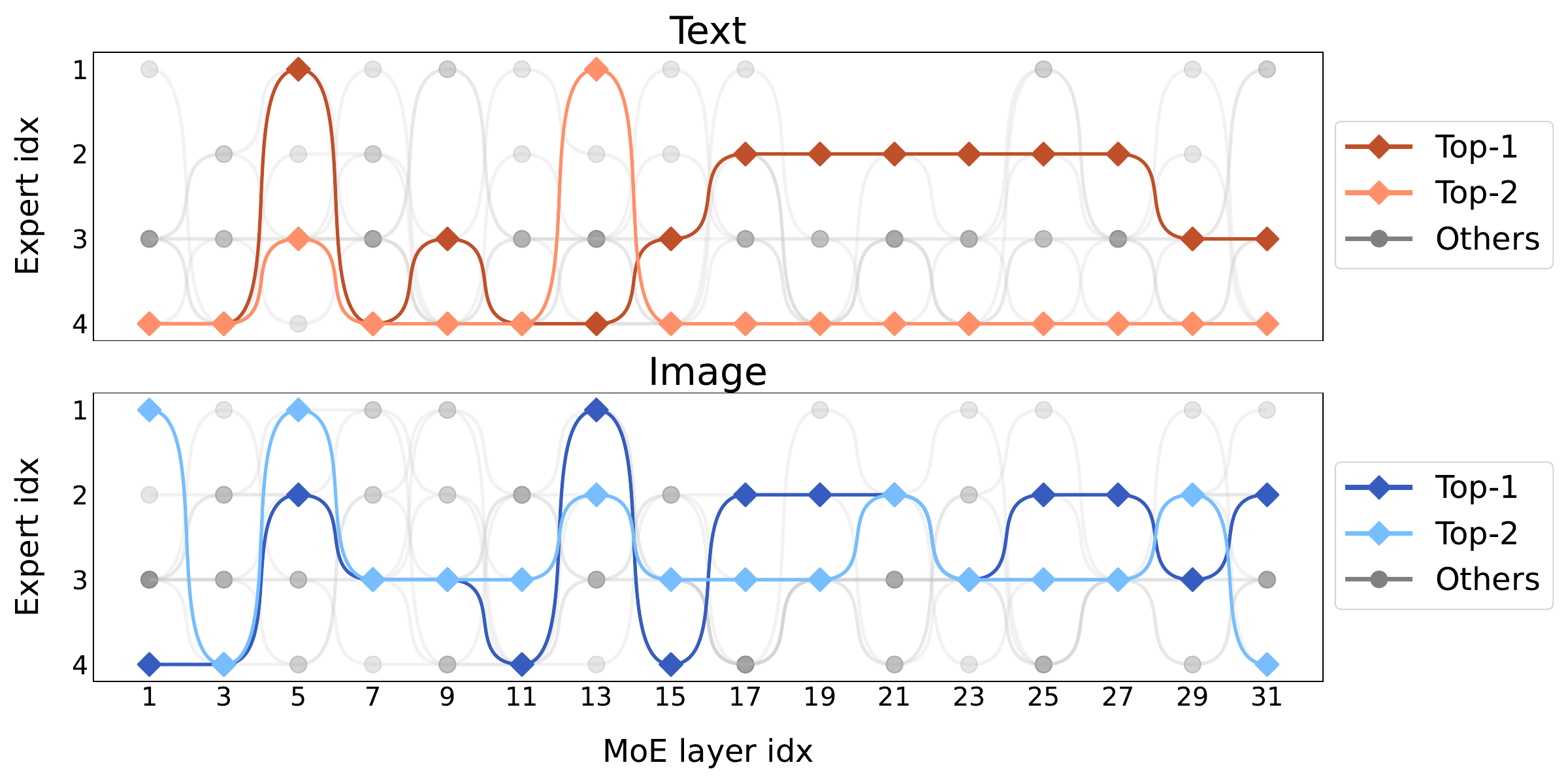}}\hspace{5pt}
	\subfloat[POPE]{\includegraphics[width=.32\columnwidth]{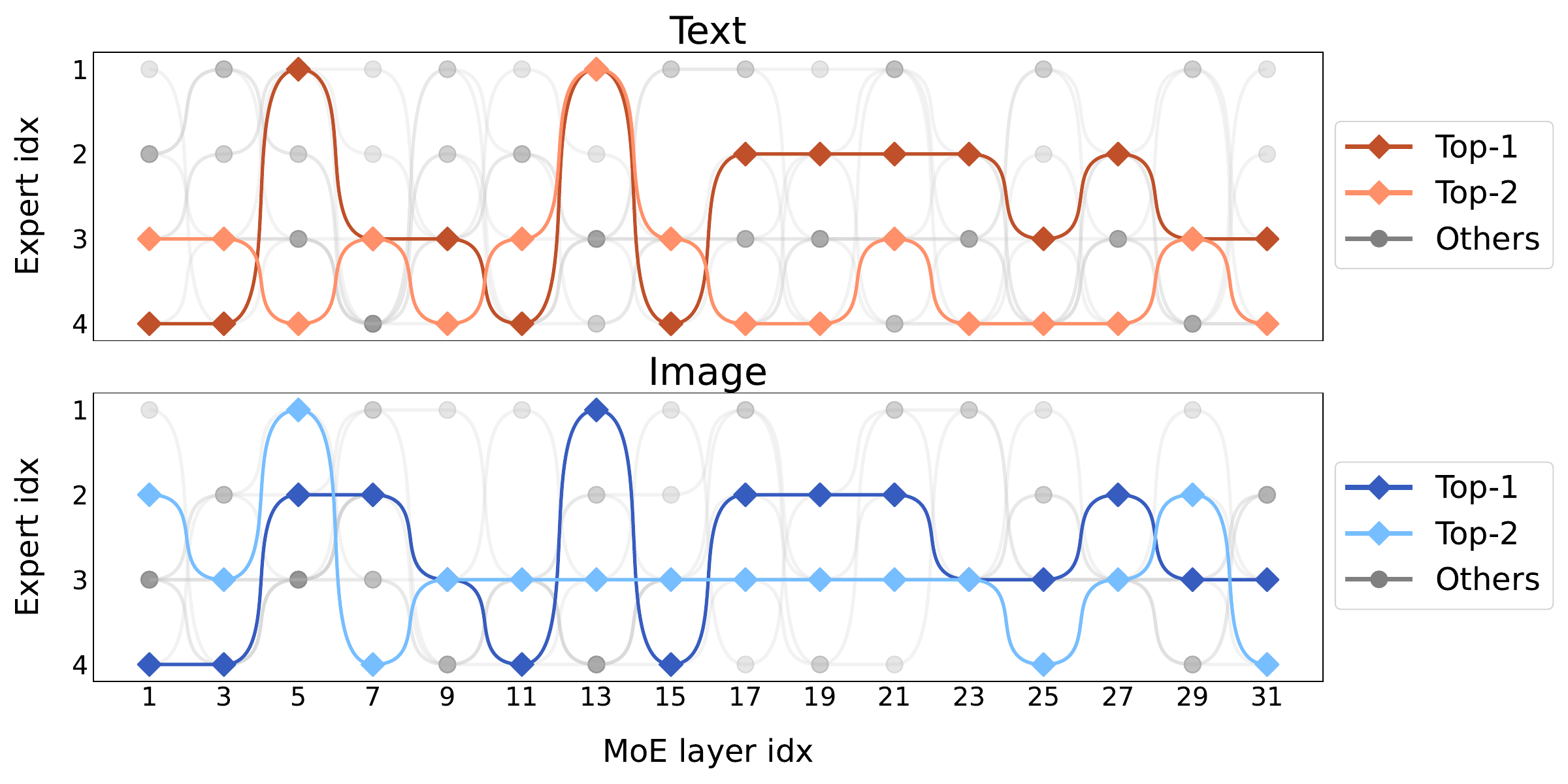}}\\
    \vskip -0.1in
	\subfloat[MMBench]{\includegraphics[width=.32\columnwidth]{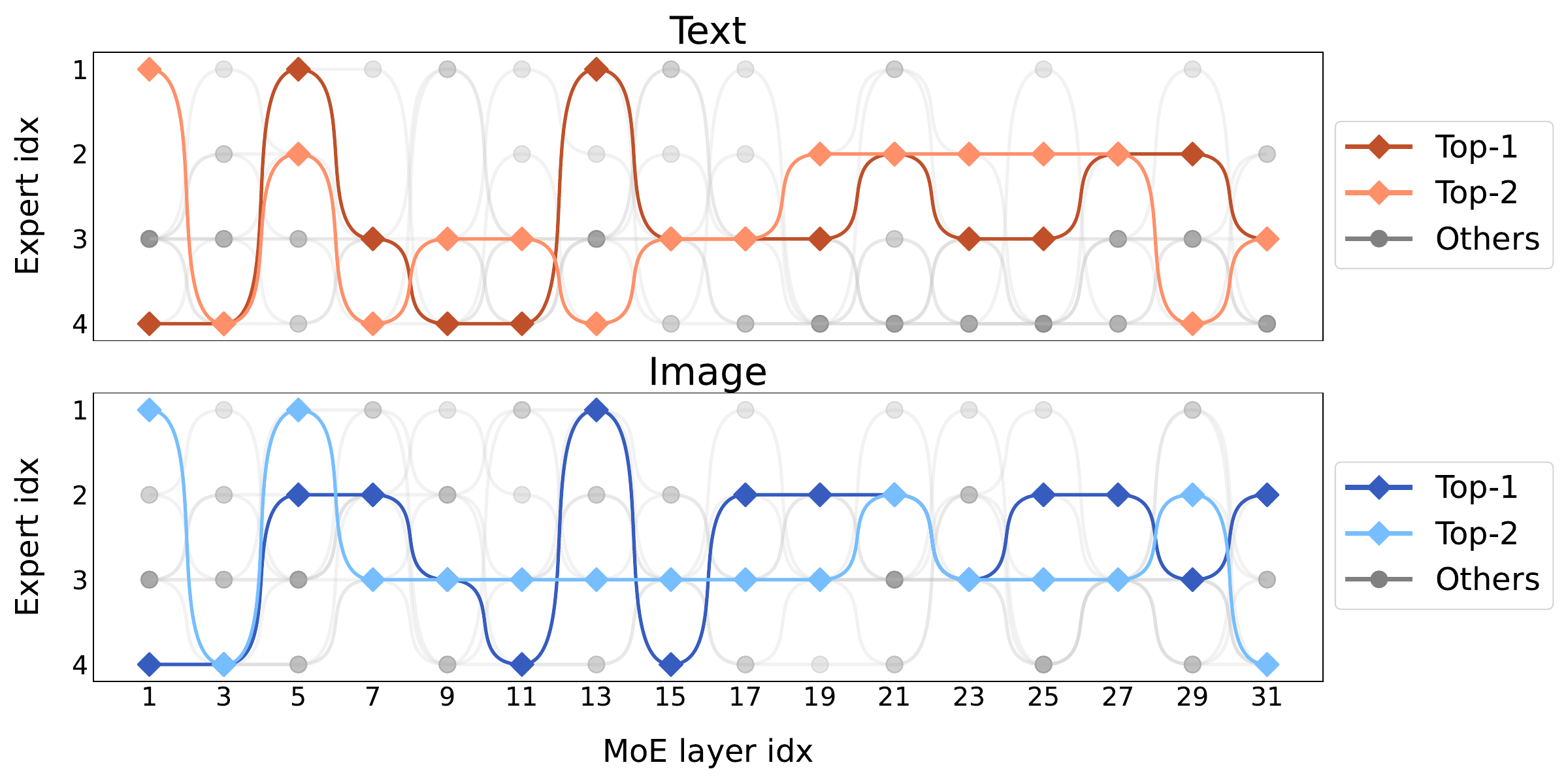}}\hspace{5pt}
	\subfloat[Viswiz]{\includegraphics[width=.32\columnwidth]{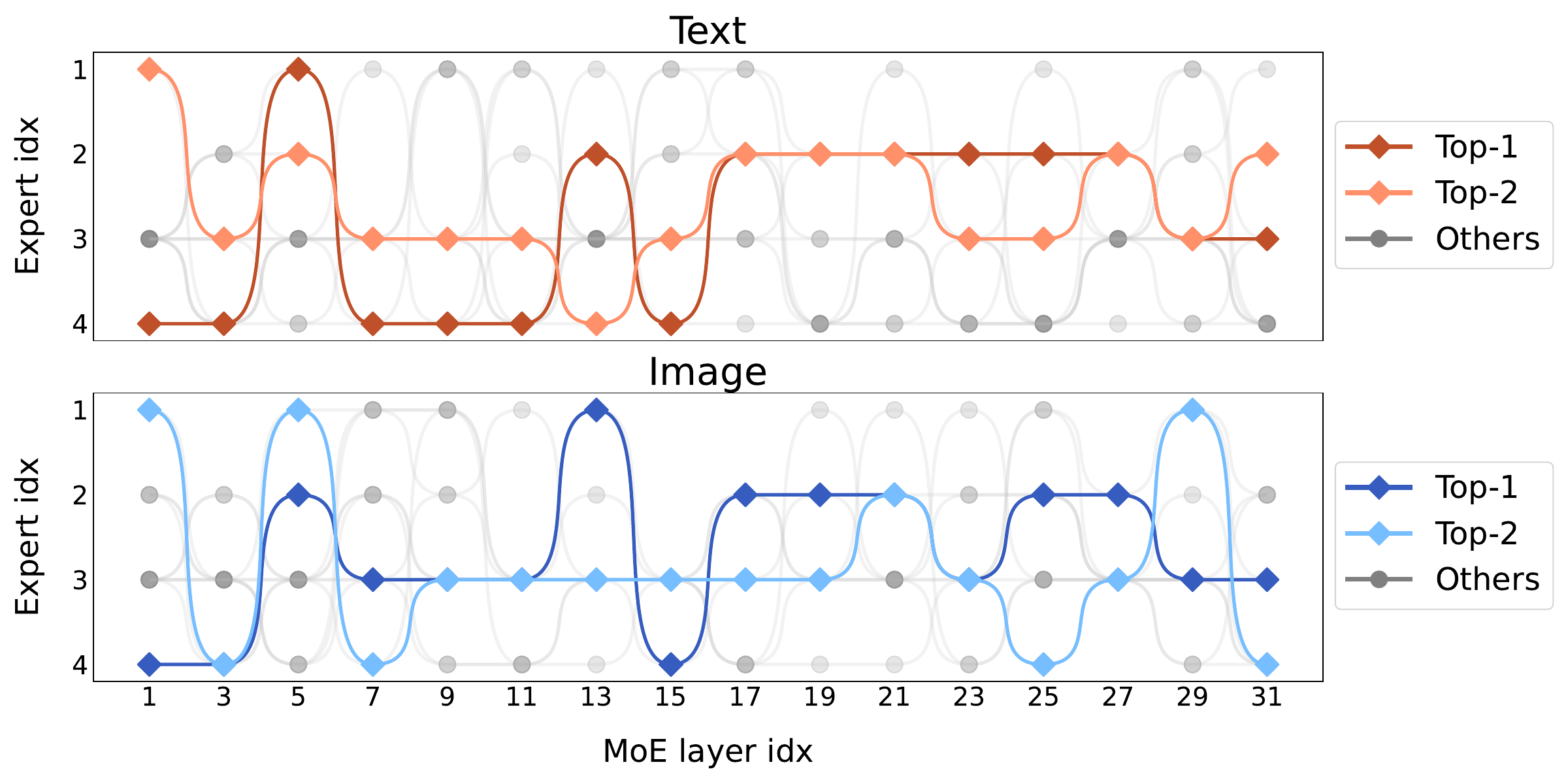}}\hspace{5pt}
	\subfloat[MM-Vet]{\includegraphics[width=.32\columnwidth]{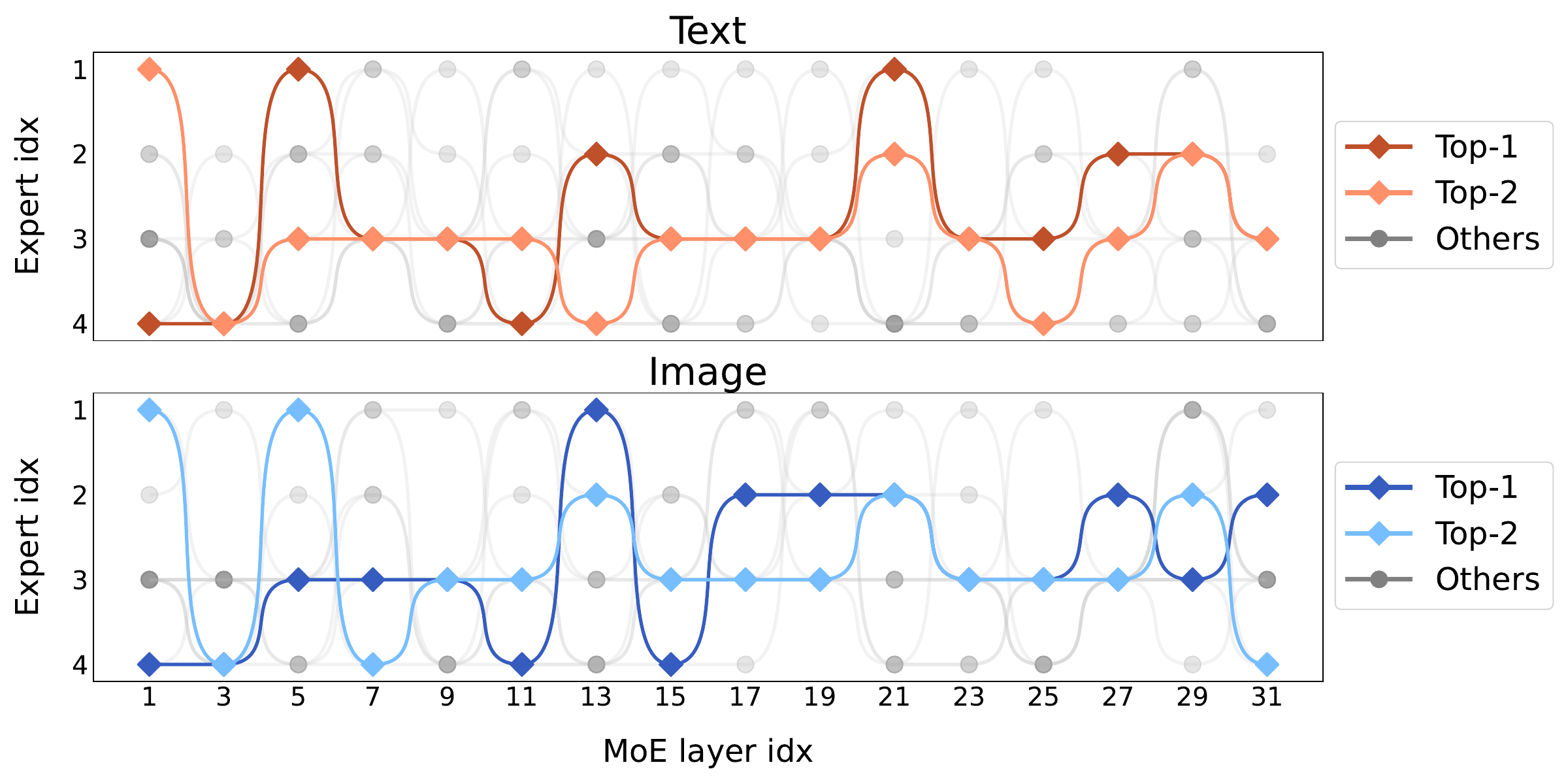}}
\vskip -0.1in
	\caption{Visualization of activated pathways on \textbf{MoE-LLaVA-Phi-2.7B×4-Top2}.}
        \label{fig:bumpy_text_img_phi}
\vskip -0.2in
\end{figure}

\begin{figure}[htbp]
	\centering
	\subfloat[ScienceQA-IMG]{\includegraphics[width=.32\columnwidth]{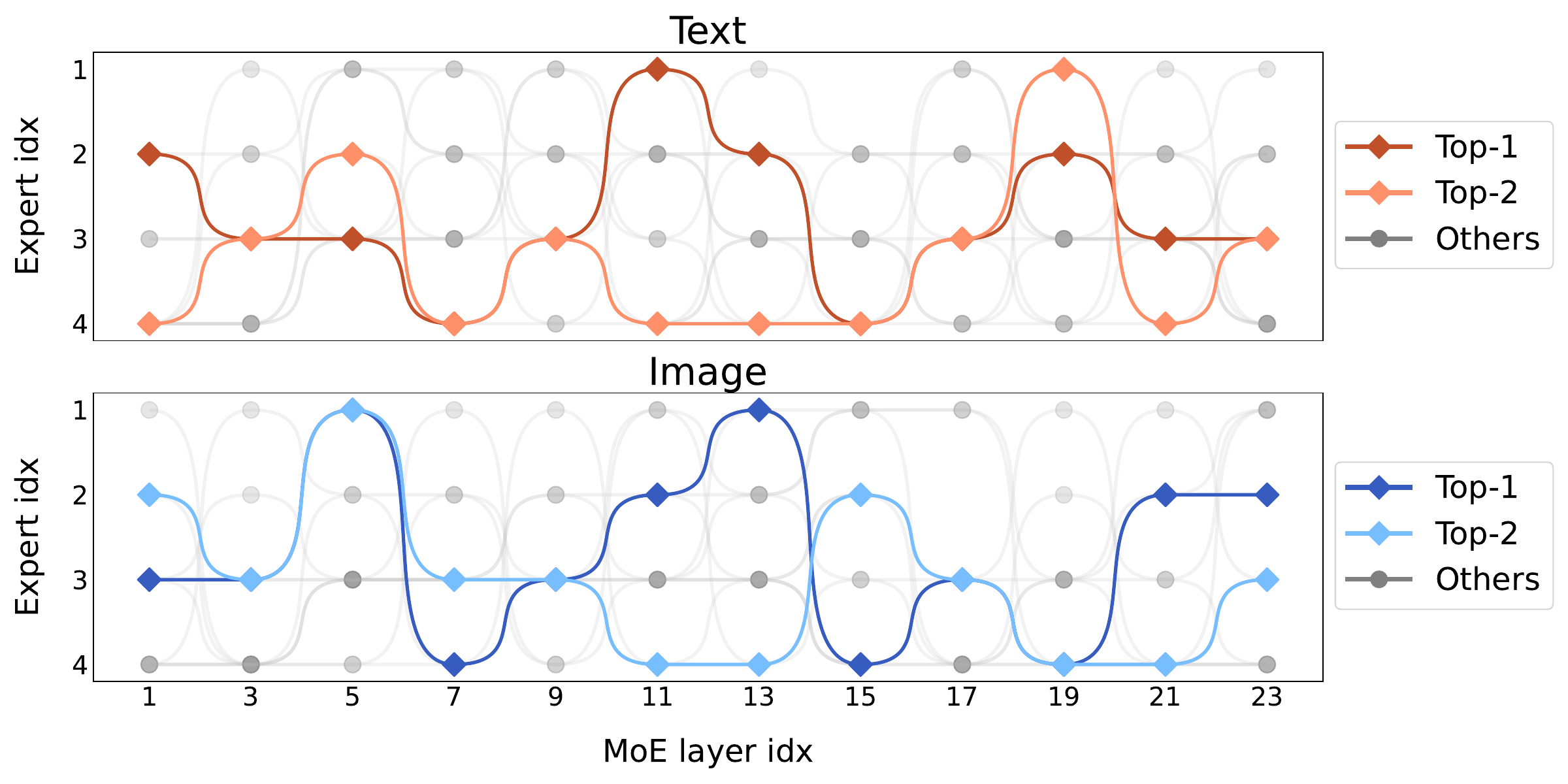}}\hspace{5pt}
	\subfloat[TextQA]{\includegraphics[width=.32\columnwidth]{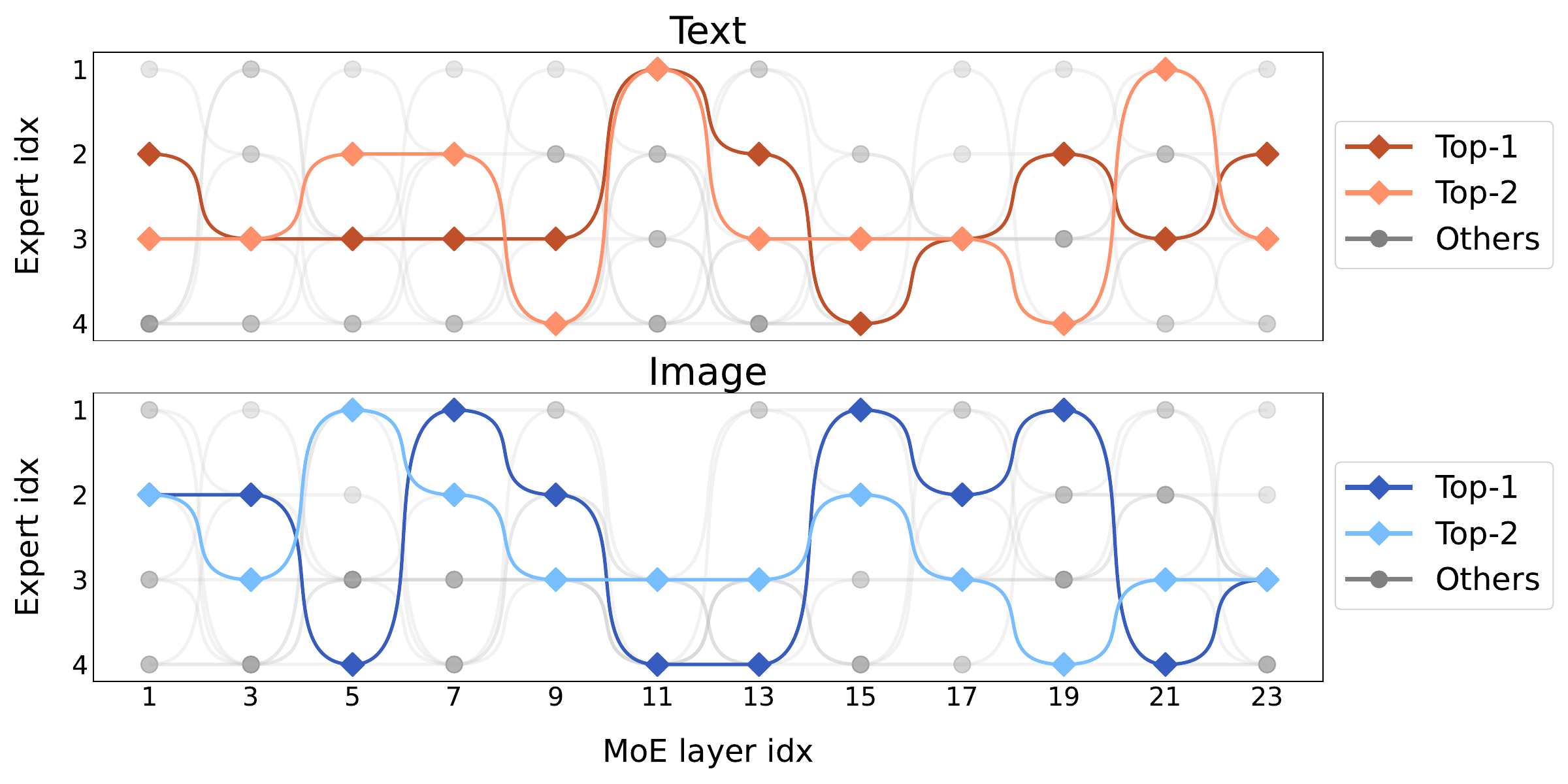}}\hspace{5pt}
	\subfloat[POPE]{\includegraphics[width=.32\columnwidth]{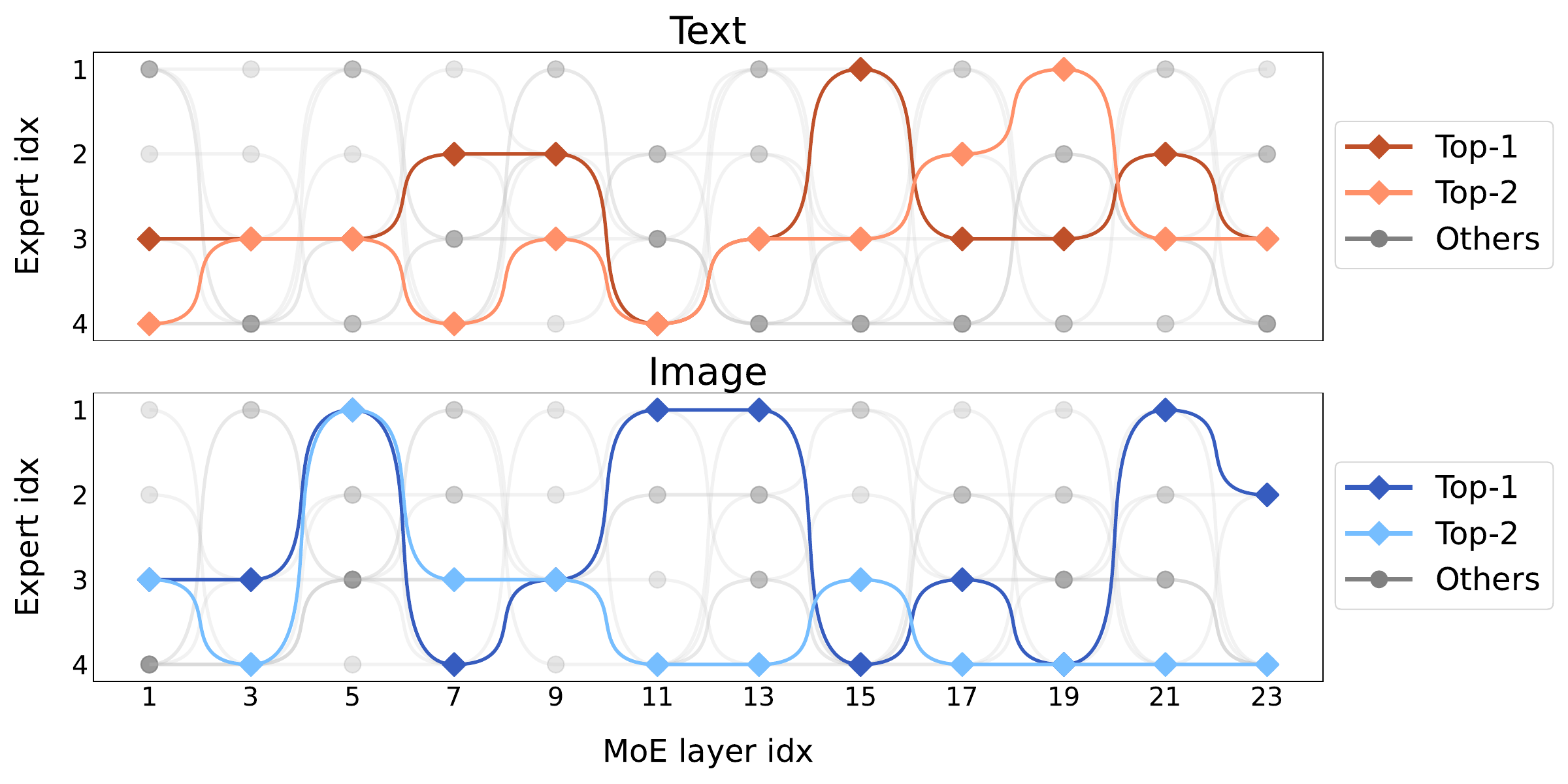}}\\
    \vskip -0.1in
	\subfloat[MMBench]{\includegraphics[width=.32\columnwidth]{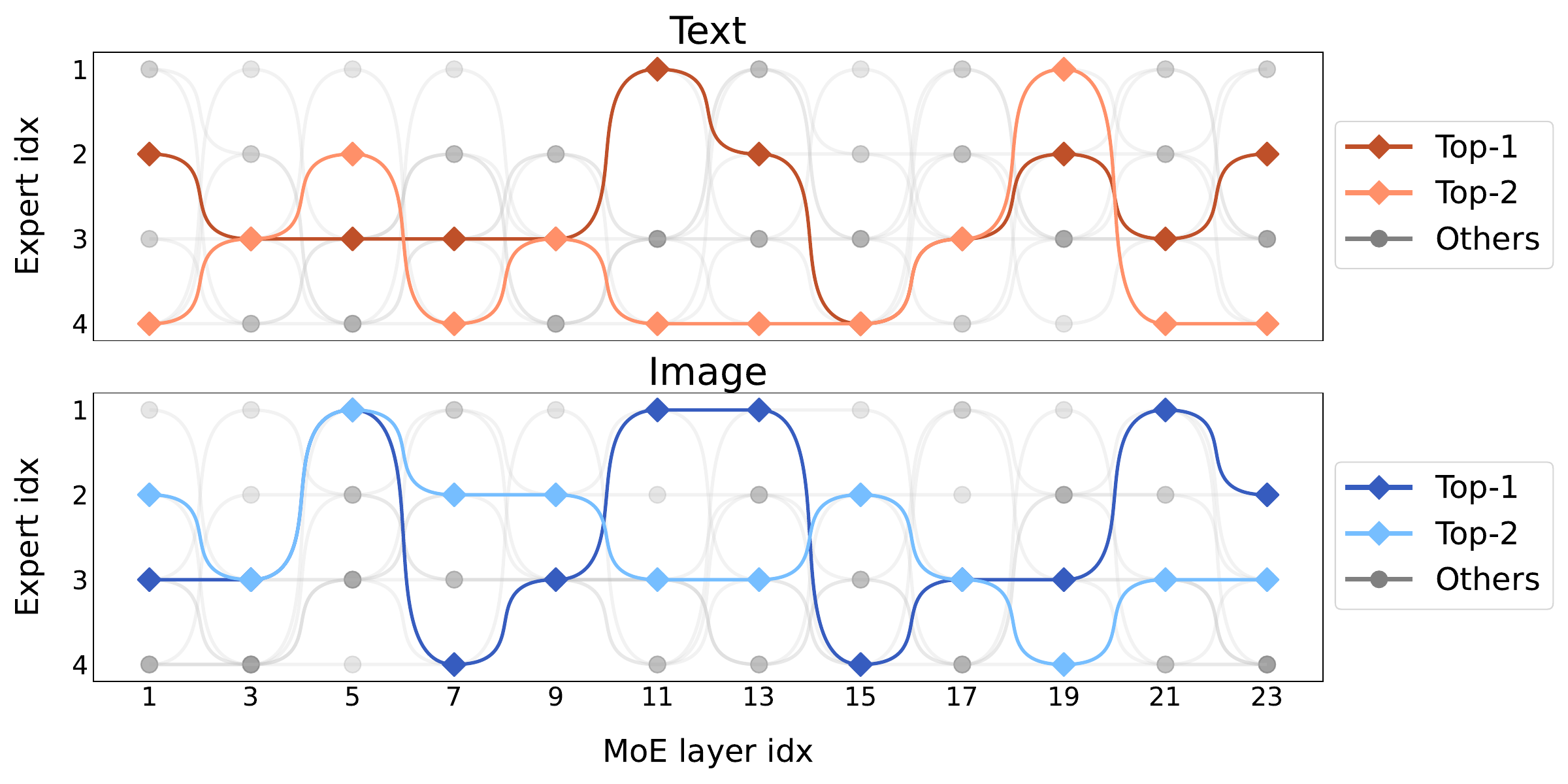}}\hspace{5pt}
	\subfloat[Viswiz]{\includegraphics[width=.32\columnwidth]{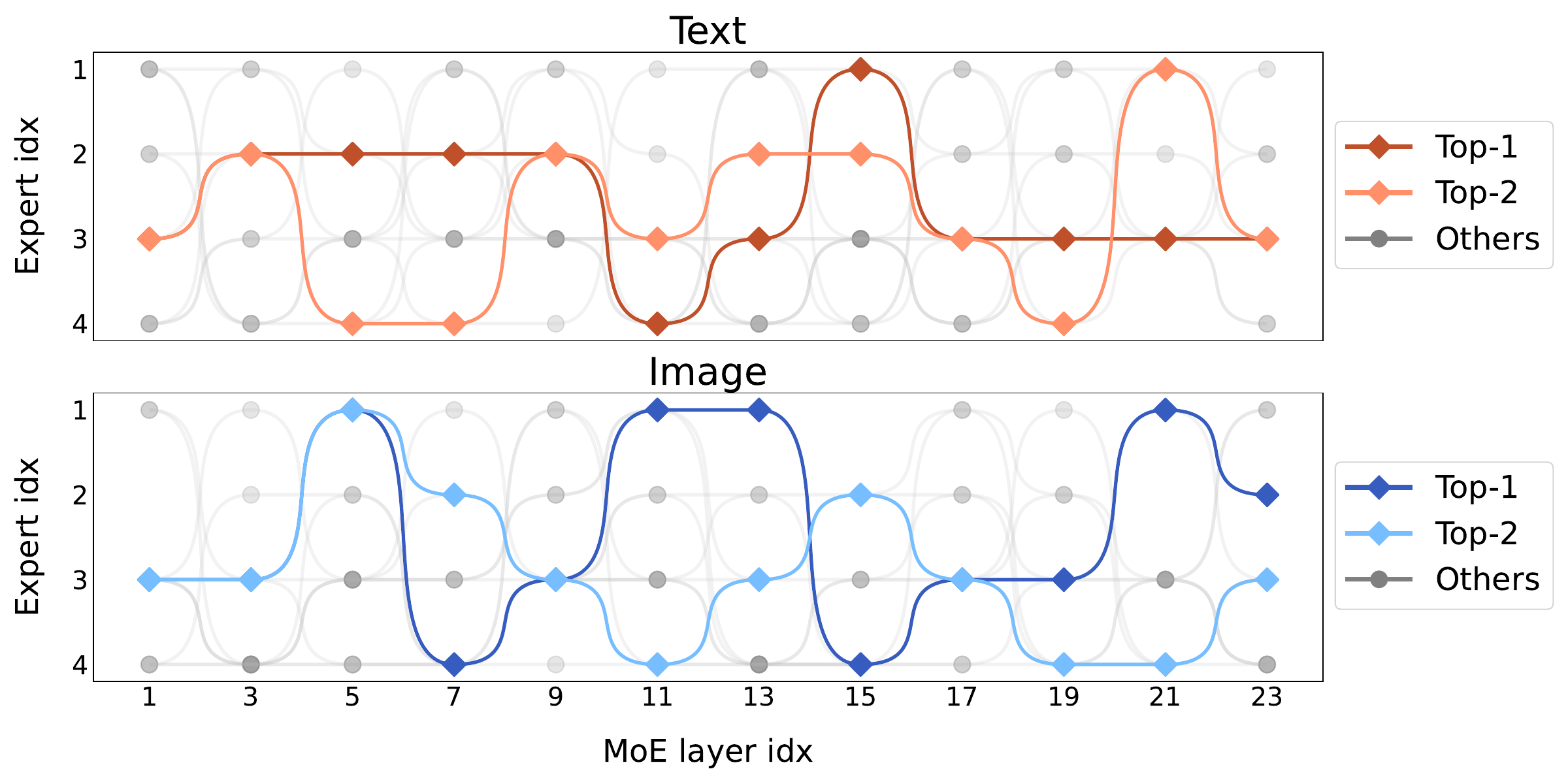}}\hspace{5pt}
	\subfloat[MM-Vet]{\includegraphics[width=.32\columnwidth]{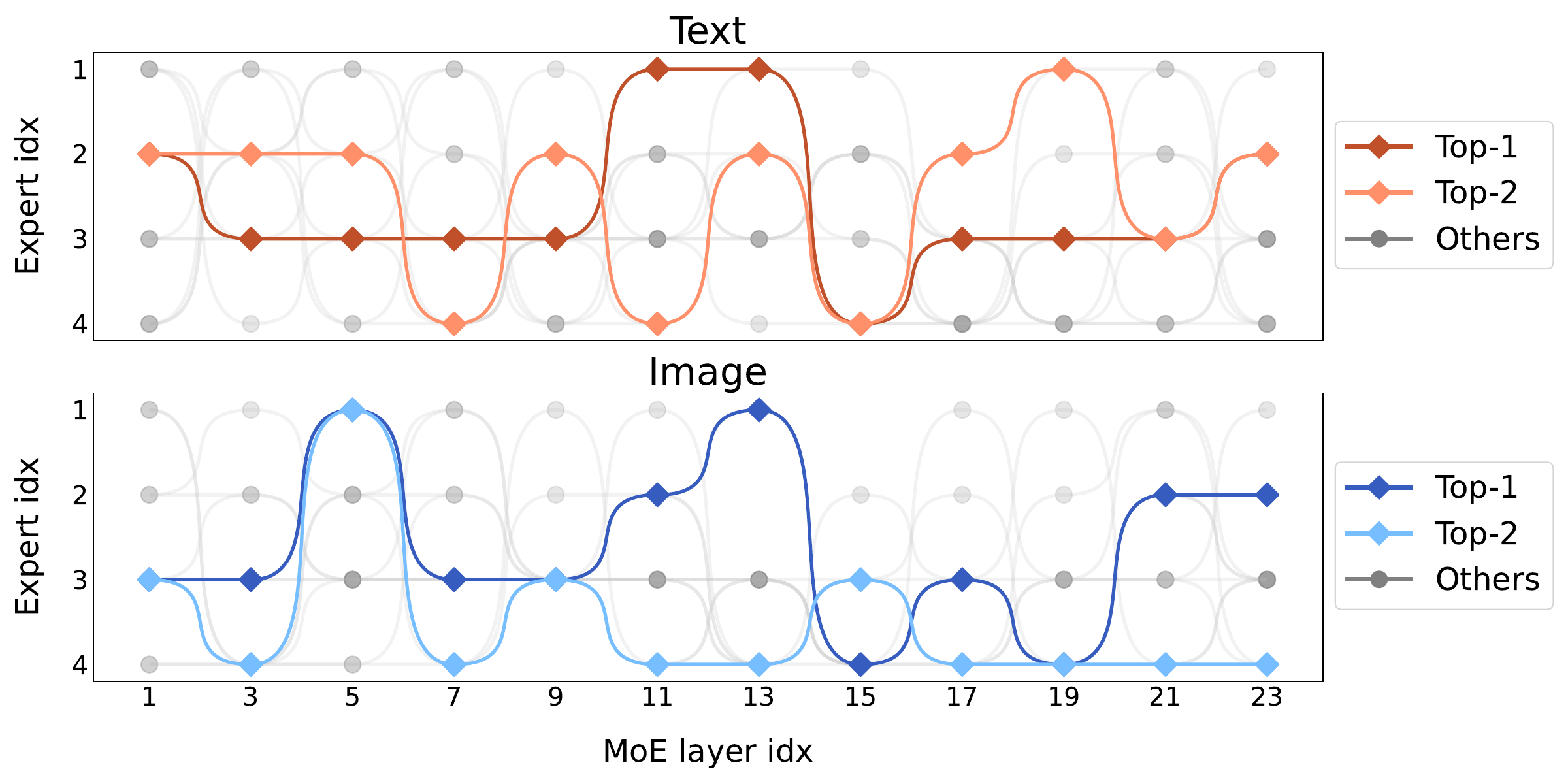}}
\vskip -0.1in
	\caption{Visualization of activated pathways on \textbf{MoE-LLaVA-Qwen-1.8B×4-Top2}.}
        \label{fig:bumpy_text_img_qwen}
\vskip -0.2in
\end{figure}

\begin{figure}[htbp]
	\centering
	\subfloat[ScienceQA-IMG]{\includegraphics[width=.32\columnwidth]{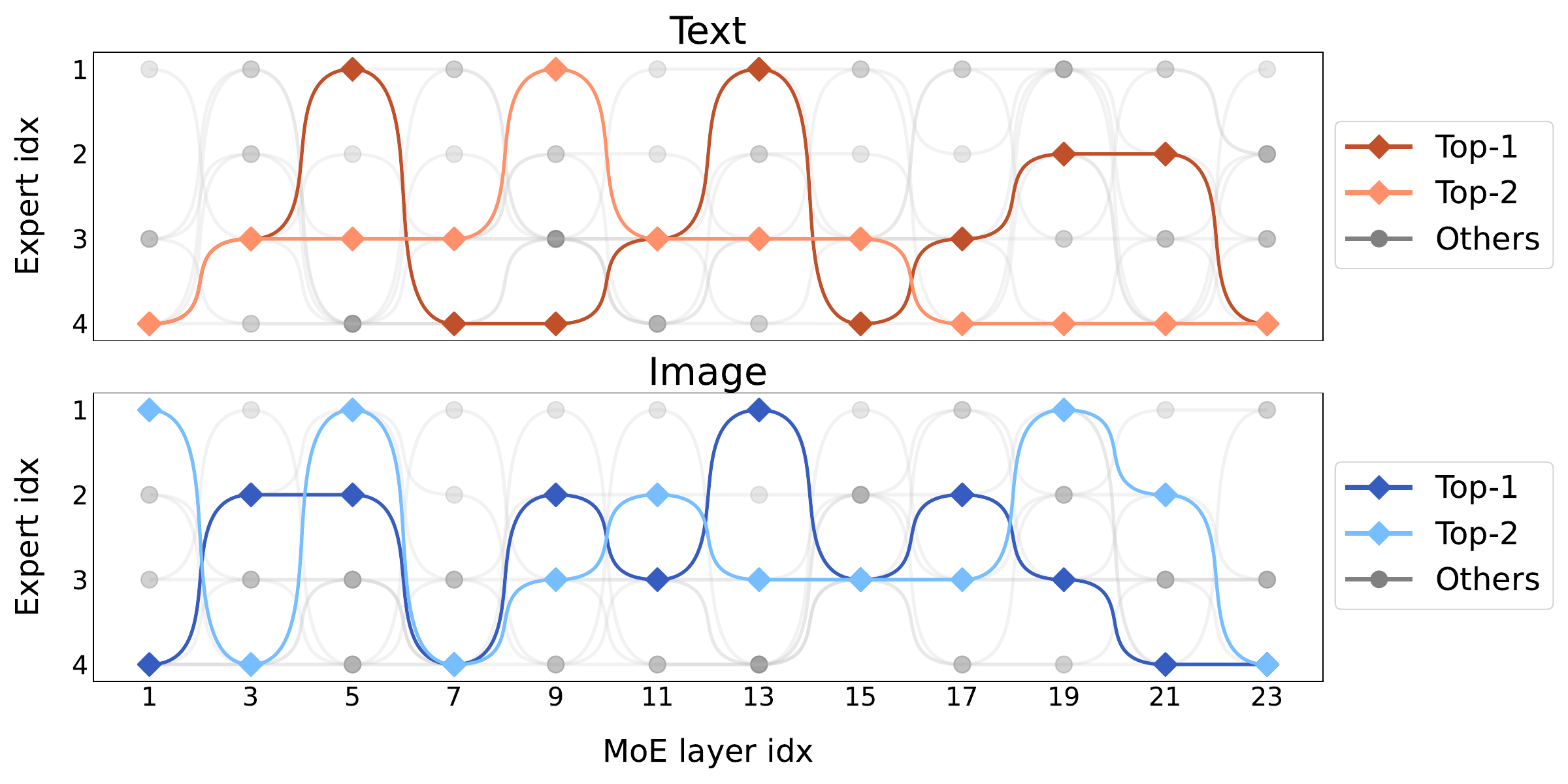}}\hspace{5pt}
	\subfloat[TextQA]{\includegraphics[width=.32\columnwidth]{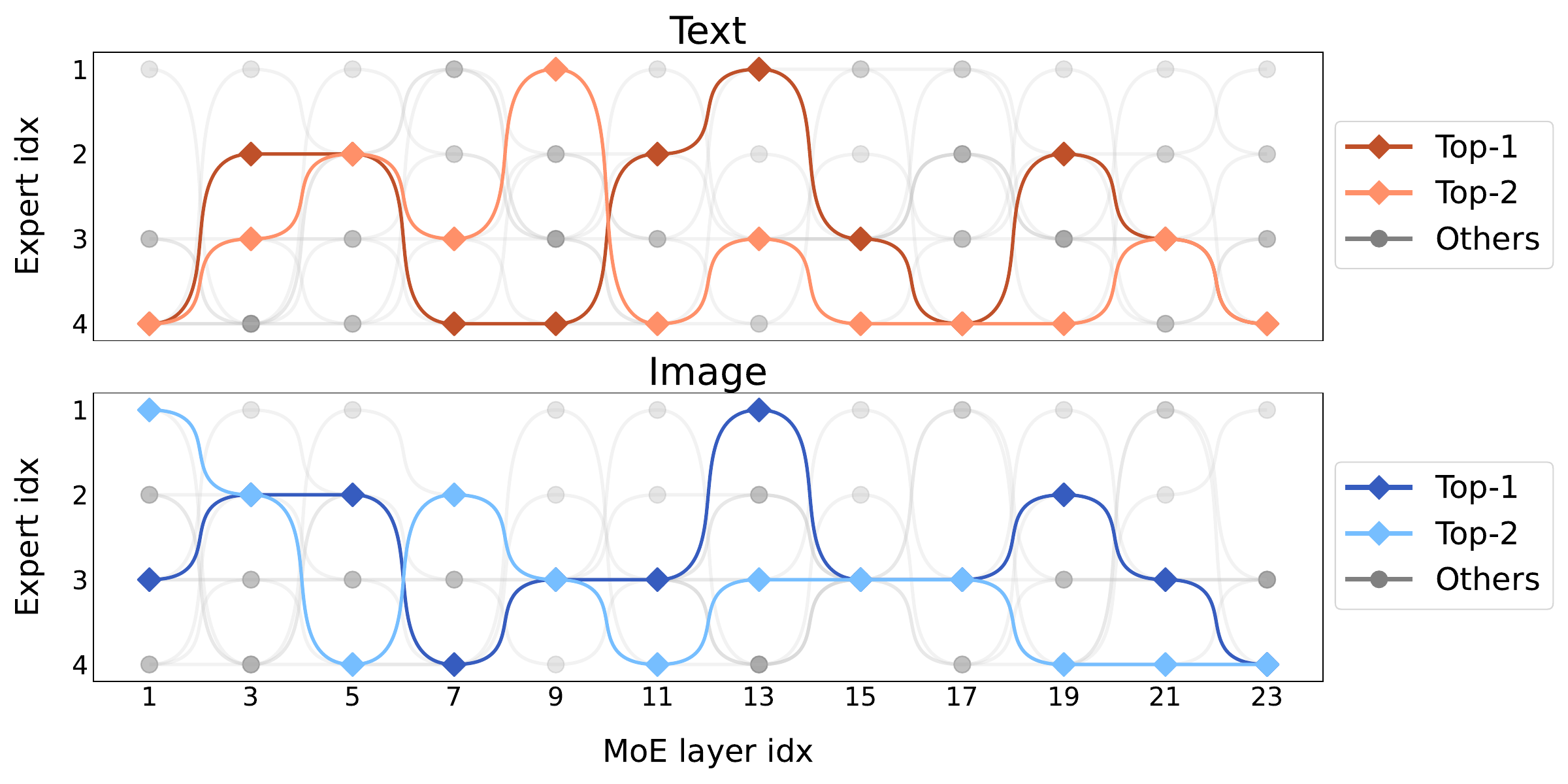}}\hspace{5pt}
	\subfloat[POPE]{\includegraphics[width=.32\columnwidth]{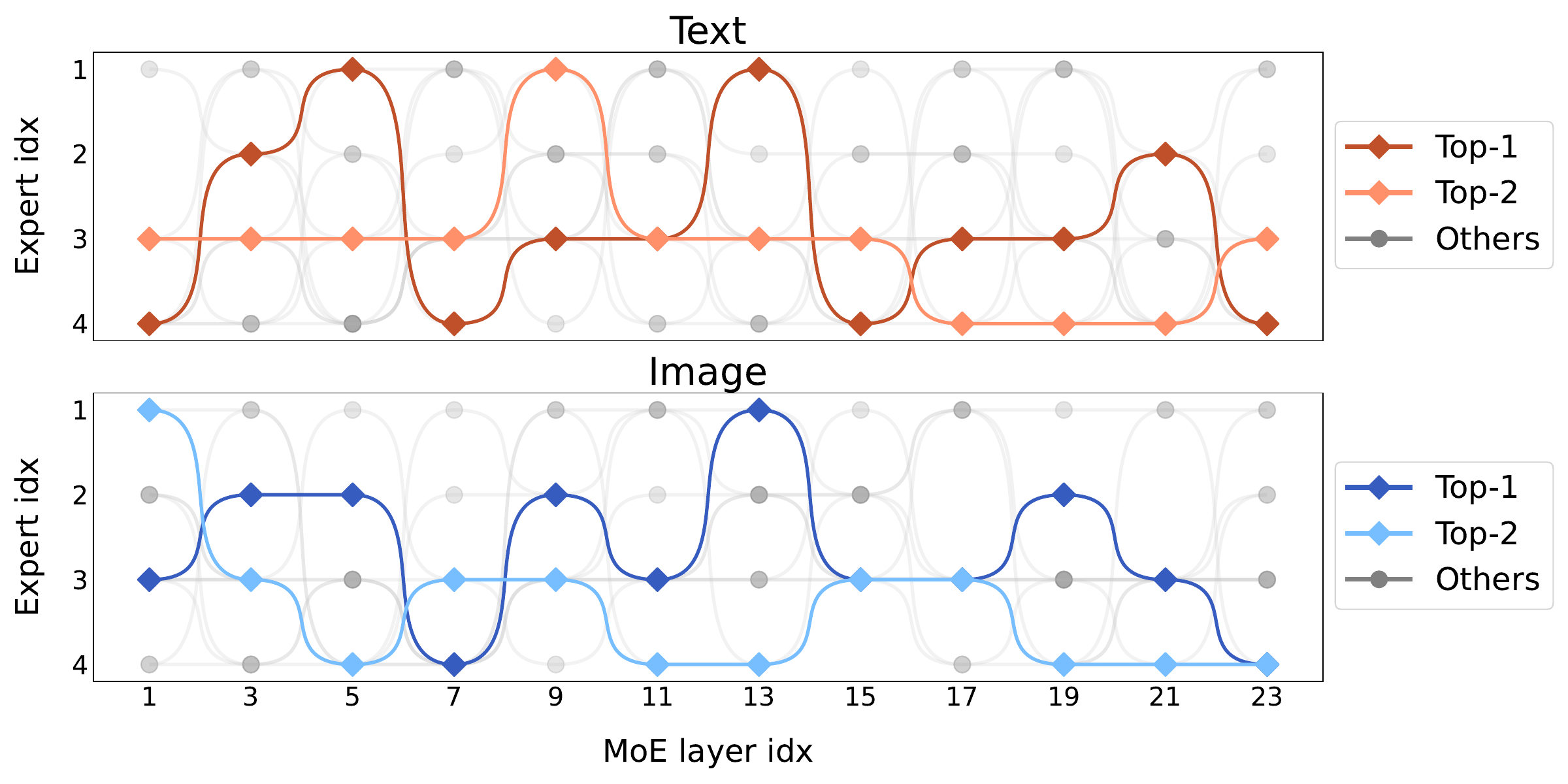}}\\
    \vskip -0.1in
	\subfloat[MMBench]{\includegraphics[width=.32\columnwidth]{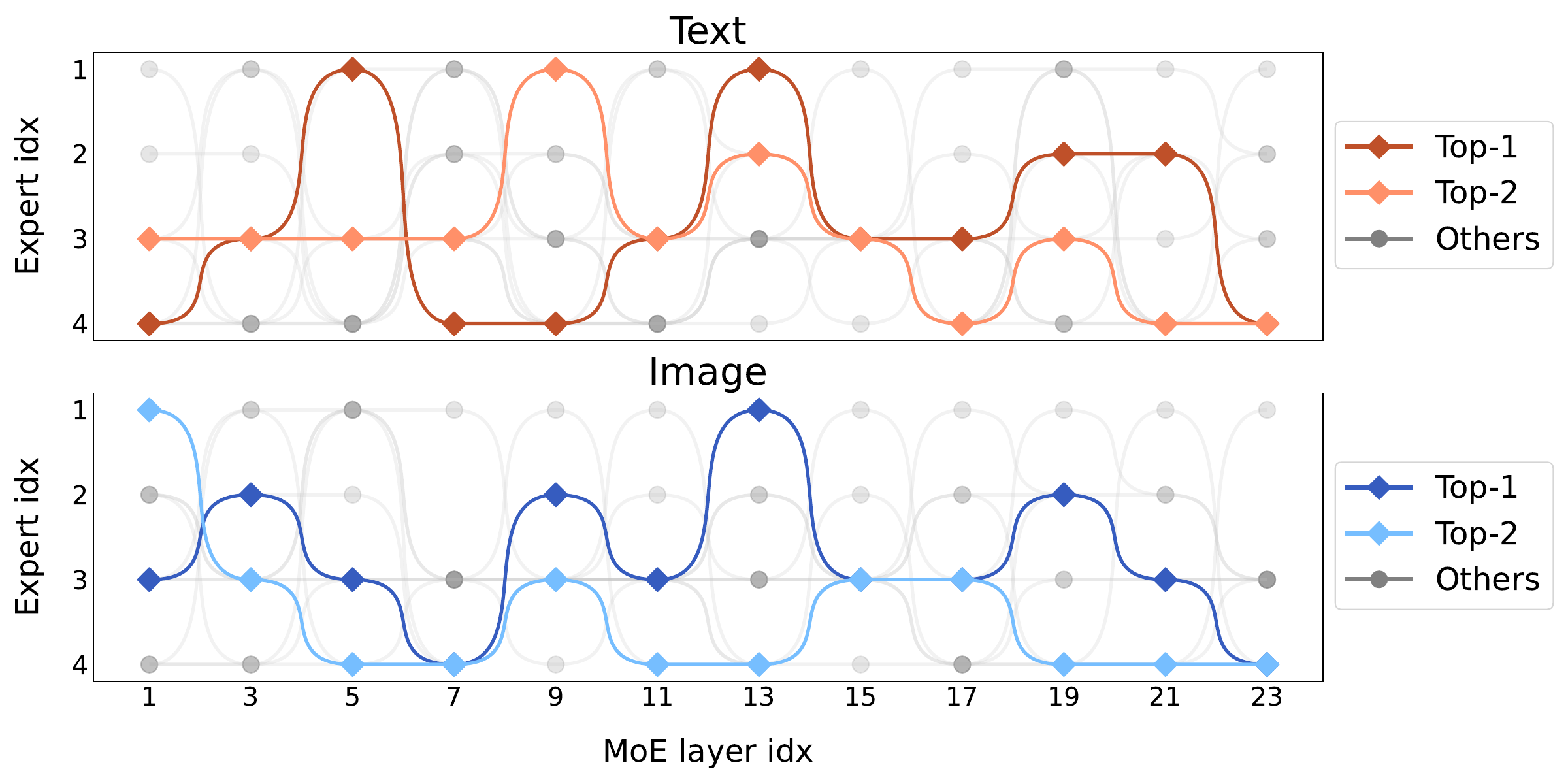}}\hspace{5pt}
	\subfloat[Viswiz]{\includegraphics[width=.32\columnwidth]{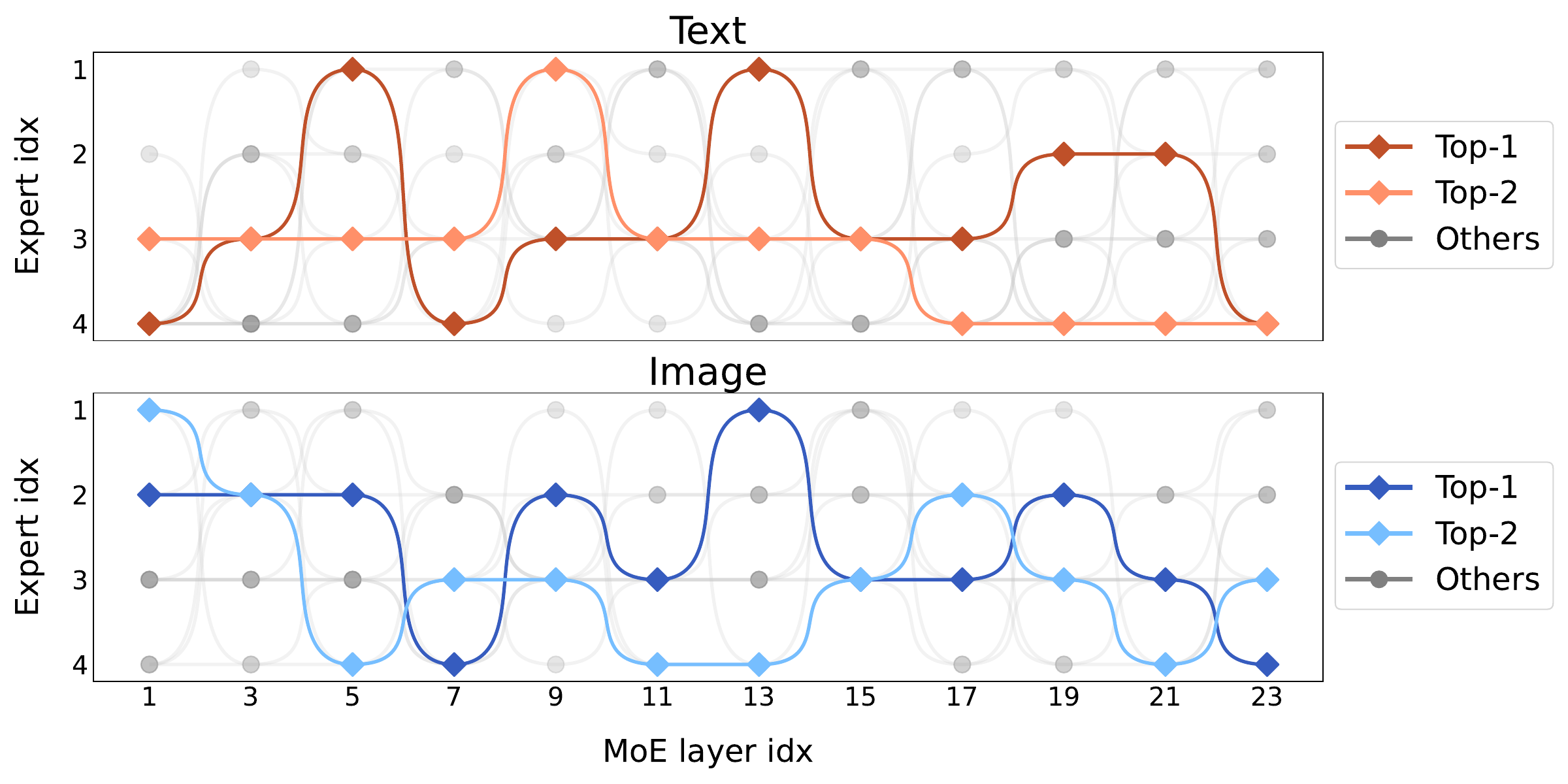}}\hspace{5pt}
	\subfloat[MM-Vet]{\includegraphics[width=.32\columnwidth]{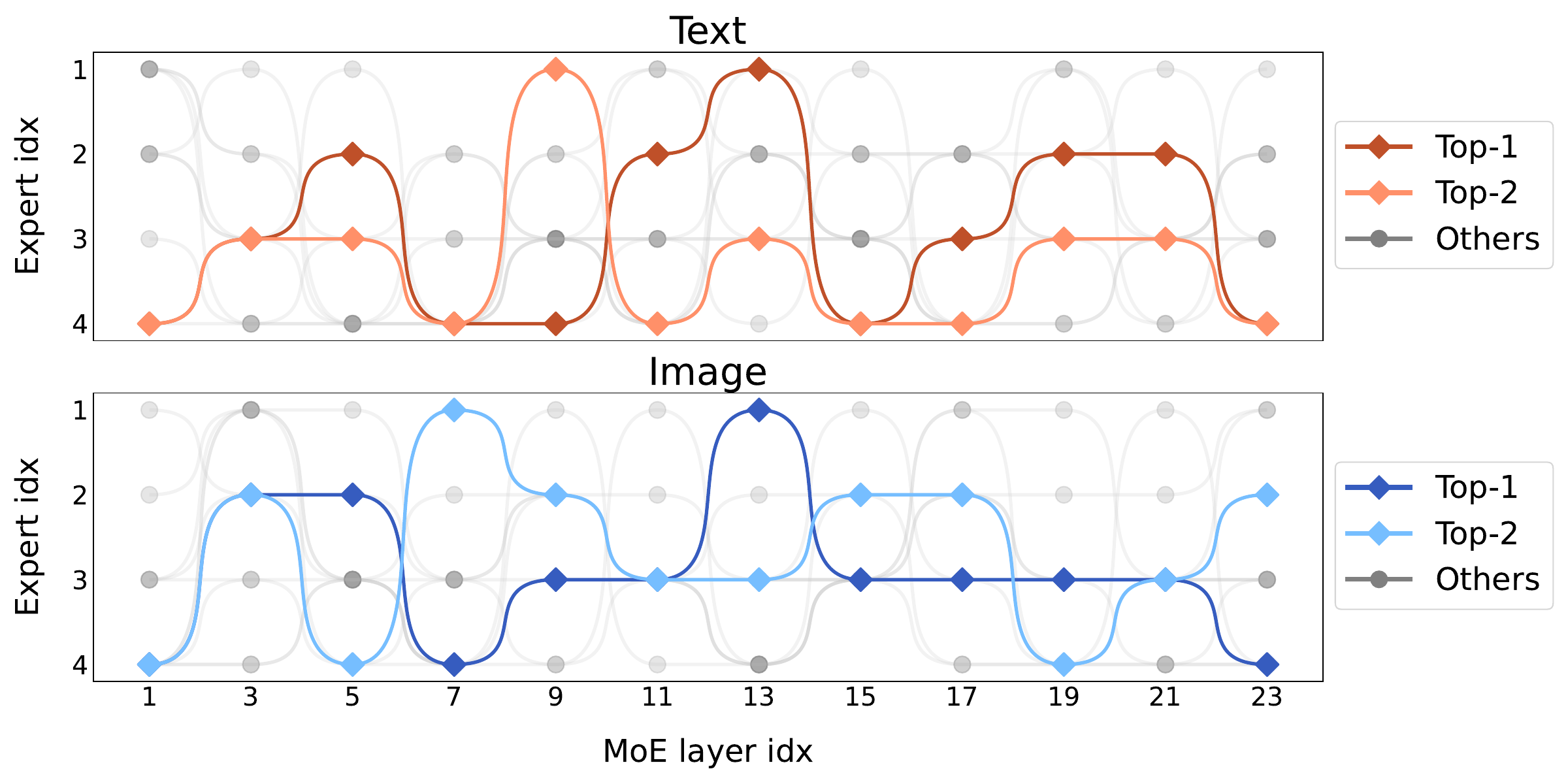}}
\vskip -0.1in
	\caption{Visualization of activated pathways on \textbf{MoE-LLaVA-StableLM-1.6B×4-Top2}.}
        \label{fig:bumpy_text_img_stablelm}
\vskip -0.2in
\end{figure}

\subsection{Exhibition Board}
\label{appendix:board}

In~\cref{tab:board}, we present some classic examples using images from LLaVA~\cite{liu2023visual} and LLaVA-1.5~\cite{liu2023improved}. We observe that MoE-LLaVA performs comparably to them on these classic images, despite using fewer parameters.

\begin{table}
  \caption{\textbf{Exhibition Board of MoE-LLaVA.} MoE-LLaVA demonstrates the ability to detect and answer challenging questions when prompted to verify them.}
  \vskip 0.1in
  \label{tab:board}
    \begin{minipage}{0.5\linewidth}
    \centering
    \scalebox{0.80}{
    \begin{tabular}{l p{7.5cm} }
    \toprule
     \multicolumn{2}{l}{\bf Visual input example, Tricky Question and  Image:}  \\
    \midrule
    &  \includegraphics[height=3.1cm]{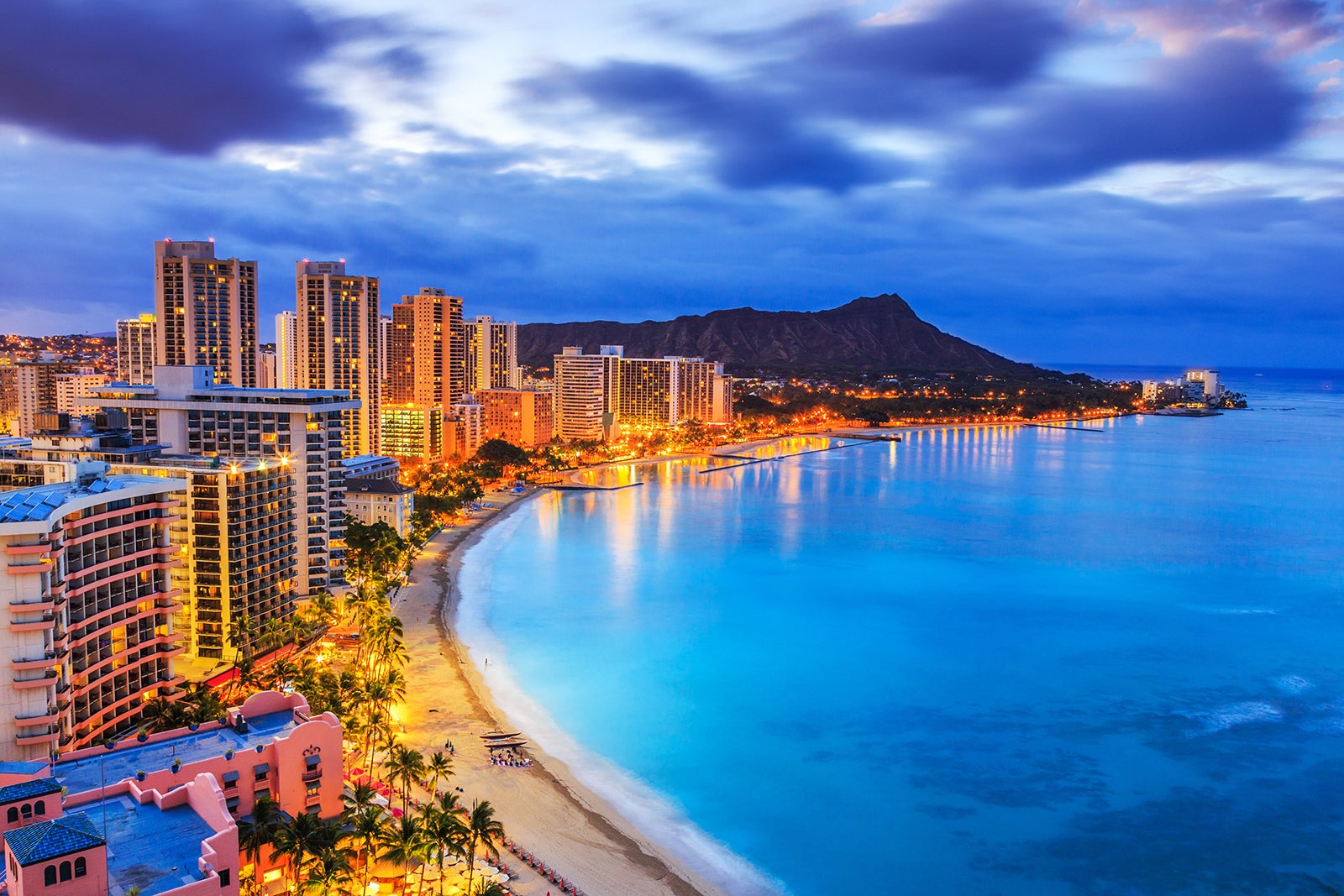} \\
    User & If there are factual errors in the questions, point it out; if not, proceed answering the question. \\
    & What's happening in the desert? \\
    \midrule
    LLaVA-1.5 & There are no deserts in the image. The image features a beach with palm trees, a city skyline, and a large body of water. \\
    \midrule
    MoE-LLaVA & There are no deserts in the image. The scene features a city with a beach and a large body of water, which is more likely to be a coastal city. \\
    \midrule
    &  \includegraphics[height=3.1cm]{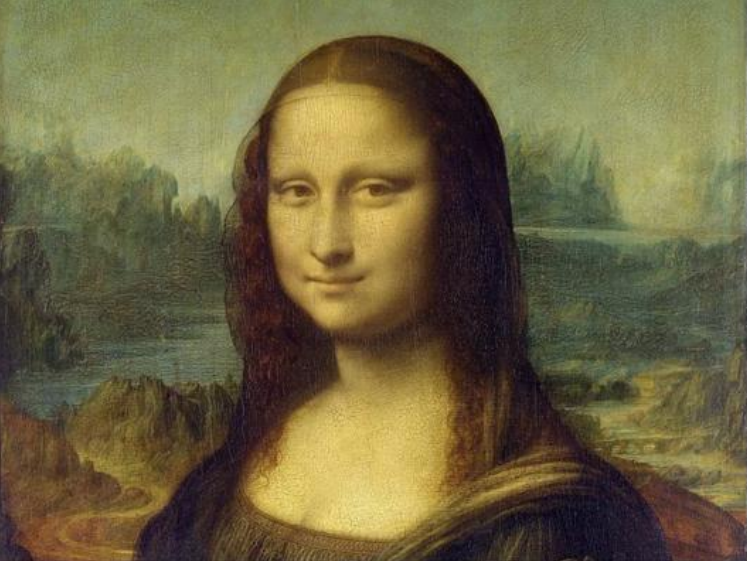} \\
    User & Do you know who drew this painting? \\
    \midrule
    MoE-LLaVA & The painting is a portrait of the famous artist, Leonardo da Vinci. \\
    \midrule
    &  \includegraphics[height=3.1cm]{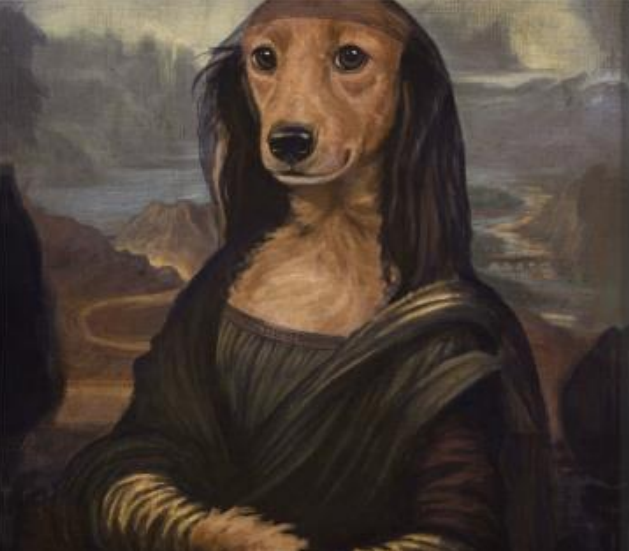} \\
    User & Do you know who drew this painting? \\
    \midrule
    MoE-LLaVA & The painting is a drawing of a dog, and it is not possible to definitively determine the artist who created it based on the image alone. \\
    \midrule
    &  \includegraphics[height=3.1cm]{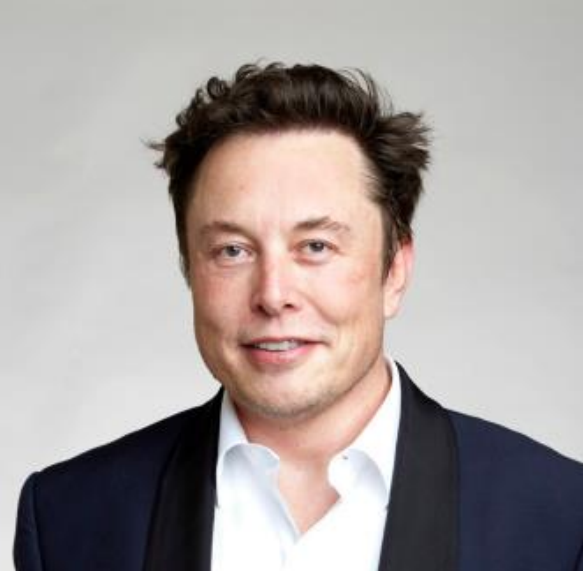} \\
    User & What is the name of the man in the photo? \\
    \midrule
    MoE-LLaVA & The man in the photo is named Elon Musk. \\
    \midrule
    &  \includegraphics[height=3.1cm]{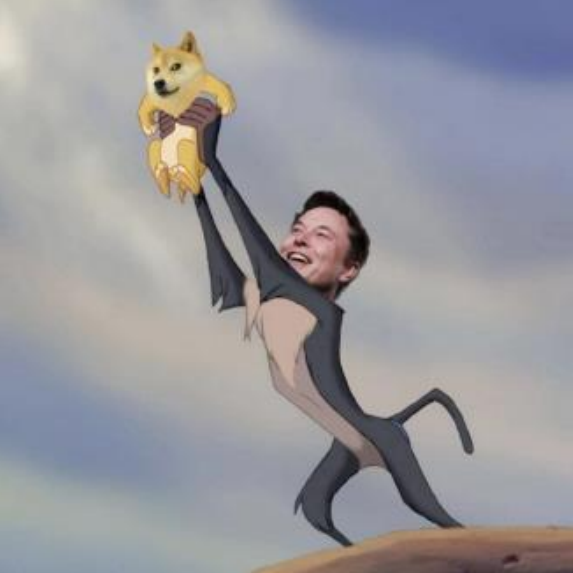} \\
    User & What is the name of the man in the photo? \\
    \midrule
    MoE-LLaVA & The man in the photo is named Elon Musk. \\
    \bottomrule
    \end{tabular}
    }
    \end{minipage}
    \begin{minipage}{0.5\linewidth}
    \centering
    \scalebox{0.80}{
    \begin{tabular}{l p{7.5cm} }
    \toprule
     \multicolumn{2}{l}{\bf Visual input example, Common Question:}  \\
    \midrule
    &  \includegraphics[height=3.2cm]{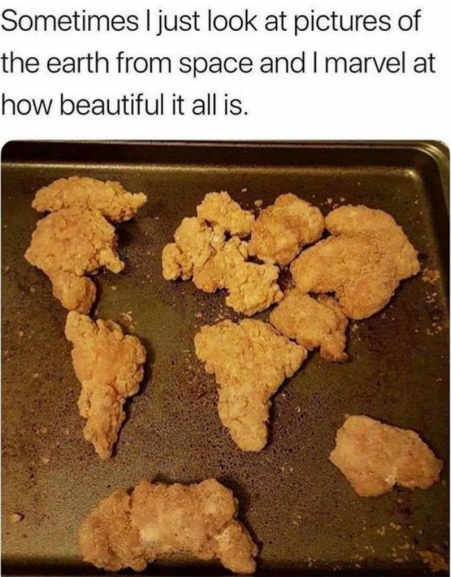} \\
    User &  Can you explain this meme? \\
    \midrule
    GPT-4~\cite{openai2023gpt4} & This meme is a joke that combines two unrelated things: pictures of the earth from space and chicken nuggets. The text of the meme suggests that the image below is a beautiful picture of the earth from space. However, the image is actually of chicken nuggets arranged to vaguely resemble a map of the world. The humor in this meme comes from the unexpected juxtaposition of the text and the image. The text sets up an expectation of a majestic image of the earth, but the image is actually something mundane and silly. \\
    \midrule
    MoE-LLaVA & The image features a pan filled with golden-brown fried chicken pieces arranged in the shape of a world map. There are numerous chicken pieces spread across the pan, with some pieces located near the center, while others are positioned towards the edges. The chicken pieces are placed in a way that resembles the continents and countries, creating a visually appealing and creative presentation. The pan itself is placed on a table, ready to be served and enjoyed. \\
    \midrule
    &  \includegraphics[height=3.2cm]{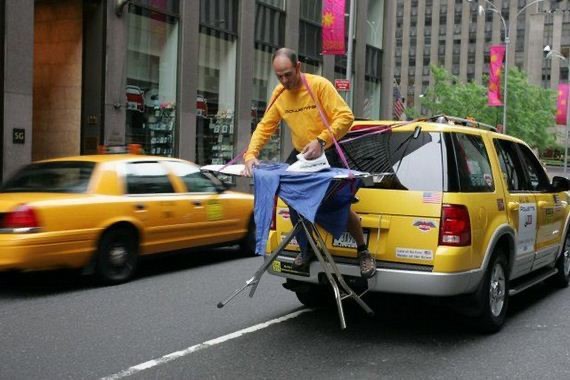} \\
    User & What is unusual about this image? \\
    \midrule
    MoE-LLaVA & The unusual aspect of this image is that a man is standing on a clothesline that is attached to the back of a moving taxi cab. This is not a typical scene, as clotheslines are usually found in residential areas, patios, or laundry rooms, and not on the back of a vehicle. The man is holding a shirt while standing on the clothesline, which adds to the unusualness of the situation. It is likely that the man is performing some task or trying to dry the shirt in an unconventional manner. \\
    \midrule
    &  \includegraphics[height=3.3cm]{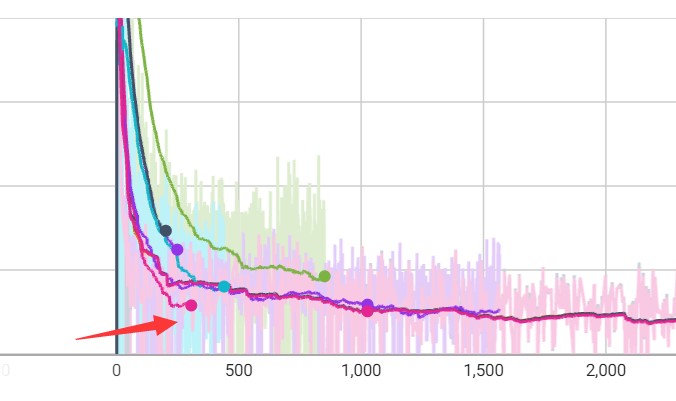} \\
    User & Describe the image and where's the red arrow? \\
    \midrule
    MoE-LLaVA & The image is a graph or chart with a red arrow pointing to the bottom. The graph is color-coded, with pink and green lines, and it appears to be a line graph. The red arrow is located at the bottom of the graph, indicating a specific point or value. \\
    \bottomrule
    \end{tabular}
    }
    \end{minipage}

\end{table}


\end{document}